\pdfoutput=1

\documentclass[11pt]{article}

\usepackage[]{acl}

\usepackage{times}
\usepackage{latexsym}

\usepackage{amsmath}
\usepackage{hyperref}
\usepackage{cleveref}

\usepackage[T1]{fontenc}

\usepackage[utf8]{inputenc}

\usepackage{microtype}

\usepackage{inconsolata}

\usepackage{multirow}
\usepackage{amsmath}
\usepackage{caption}
\usepackage{subcaption}
\usepackage{booktabs}
\usepackage{tabularx}
\usepackage{graphicx}
\usepackage{colortbl}
\usepackage{lipsum}
\usepackage{appendix}
\usepackage{amssymb}
\usepackage{amsmath}

\usepackage{algpseudocode}
\usepackage{subcaption}
\usepackage[ruled,vlined]{algorithm2e}
\usepackage{placeins}

\usepackage{color,soul}
\usepackage{xcolor}
\usepackage{pifont}
\usepackage[colorinlistoftodos,prependcaption,textsize=tiny]{todonotes}

\newcommand{\drflow}{\textsc{DRFlow}}

\definecolor{mygreen}{RGB}{0, 128, 0}
\newcommand{\cmark}{{\color{green!70!black}\ding{51}}}
\newcommand{\xmark}{{\color{red!70!black}\ding{55}}}

\usepackage[most]{tcolorbox}
\usepackage{listings}
\usepackage{multicol}
\usepackage{adjustbox}
\usepackage{fvextra}

\definecolor{PromptCyan}{HTML}{00A6D6}
\definecolor{PromptGray}{HTML}{F2F3F5}

\lstdefinestyle{promptstylecompact}{
    basicstyle=\ttfamily\tiny,
    breaklines=true,
    columns=fullflexible,
    keepspaces=true,
    showstringspaces=false,
    aboveskip=0pt,
    belowskip=0pt,
    lineskip=-1pt
}

\definecolor{gooddown}{RGB}{34,139,34}
\definecolor{baddown}{RGB}{200,30,30}
\newcommand{\dn}[1]{{\scriptsize\textcolor{baddown}{$\downarrow$#1}}}

\title{
\raisebox{-2.1ex}{\protect\includegraphics[height=4.5\fontcharht\font`\B]{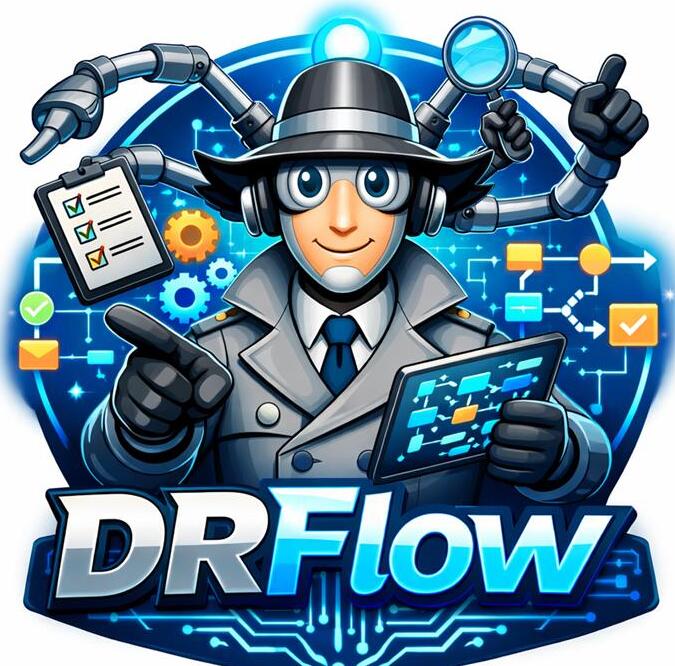}}
\drflow: A Deep Research Benchmark for Personalized Workflow Prediction}

\author{Md Tawkat Islam Khondaker$^{\spadesuit}$$^\diamondsuit$~~ {Raymond Li$^{\spadesuit}$$^\diamondsuit$}\\~~ \textbf{Muhammad Abdul-Mageed$^{\clubsuit}$$^\diamondsuit$}~~ \textbf{Laks V.S. Lakshmanan$^{\diamondsuit}$}~~ \textbf{Issam H. Laradji$^{\spadesuit}$$^\diamondsuit$}
\\\\ 
\normalsize $^{\spadesuit}$ServiceNow AI Research, $^\diamondsuit$The University of British Columbia, $^{\clubsuit}$Canada Research Chair in NLP and ML \\
}

\begin{document}
\maketitle
\begin{abstract}
Deep research (DR) systems are increasingly used for complex information-seeking tasks, but existing works mainly focus on generating reports and summaries. In contrast, many enterprise tasks instead require an agent to identify concrete workflows which is a sequence of action-steps. For example, rather than summarizing budgeting policies, an agent should be able to determine the steps needed to answer a question such as: ``How do I request new headcount given a fixed budget?''. Therefore, we introduce \drflow, a benchmark for evaluating personalized workflows predicted by agents from heterogeneous sources. Each task requires the agent to identify relevant evidence from scattered sources, then use that evidence to predict the correct action-step sequence for the user's task. \drflow~contains $100$ tasks across five domains, with $1,246$ reference workflow steps grounded in more than $3,900$ sources. We define seven diagnostic metrics covering factual grounding, step recovery, structural ordering, condition resolution, and personalization. We further present \drflow-Agent (DRFA), a workflow-oriented reference agent to predict personalized workflow. We show that although DRFA improves over strong baseline agents (upto $10.02$\% average F1 score), there is substantial room for improvement remains across these workflow metrics, indicating that predicting complete and correct personalized workflows remains a challenging frontier for deep research\footnote{Code: \url{https://github.com/ServiceNow/drflow}}.

\end{abstract}

\section{Introduction}
\label{sec:introduction}

\begin{figure}[t]
    \centering
    \includegraphics[width=0.92\linewidth]{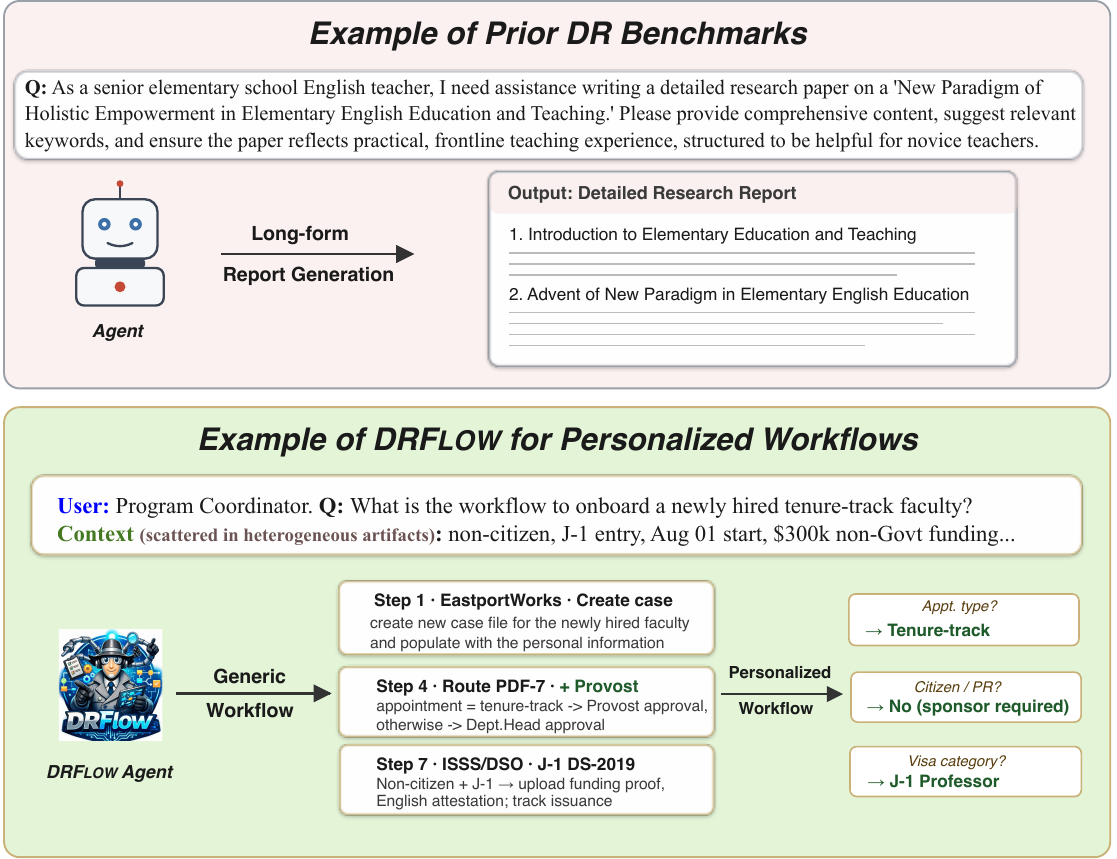}
    \caption{
    Comparison of deep research questions between representative prior work~\cite{du2025deepresearchbench} and \drflow. \drflow~expects actionable procedures rather than free-form output. For other distinctions with the prior works see \S\ref{section:related_work}.
    }
    \label{fig:intro_question_example}
\end{figure}

\begin{figure*}[t]
    \centering
    \includegraphics[width=0.8\textwidth]{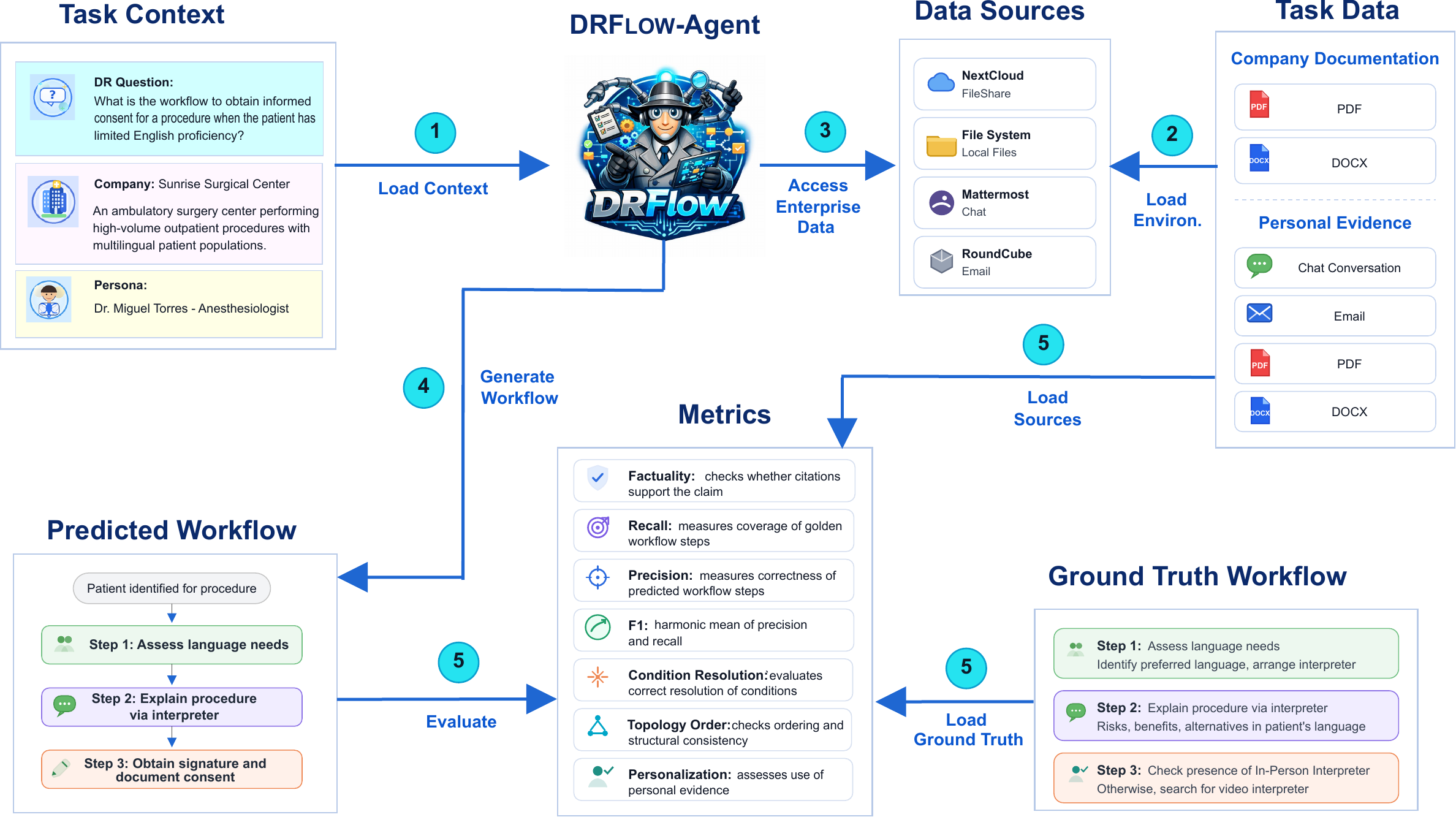}
    \caption{
    Overview of the \drflow~task and evaluation pipeline. Given a task context and heterogeneous task data, the system loads the environment and available sources, generates a predicted personalized workflow, and evaluates it against the ground-truth workflow.
    }
    \label{fig:task_evaluation_pipeline}
\end{figure*}

Deep research has emerged as an important evaluation setting for language-model based research systems~\cite{mialon2024gaia,pmlr-v235-lu24e,yoran-etal-2024-assistantbench,fact-fetch-2025,sharma2025researchrubrics,asai2026synthesizing}. Unlike conventional question answering, it requires a system to search across multiple information sources, identify relevant evidence, reconcile partial findings, and predict a grounded response to an open-ended user request. This shift has pushed evaluation beyond short factual answers toward more realistic tasks that require retrieval, reasoning, and evidence aggregation.

However, existing deep research benchmarks~\cite{du2025deepresearchbench,xu2025researcherbench,zhong2026draco} still primarily expect free-form outputs, such as reports or citation-grounded long-form answers, rather than actionable procedures. This leaves one of the key capabilities underexplored. In many realistic enterprise settings, users do not primarily need narrative summary, rather a workflow detailing what steps to take, in what order, under which conditions, and based on which evidence (See Figure~\ref{fig:intro_question_example} for an example). Other works~\cite{liu-etal-2025-workteam,zhong2026chat2workflow} on workflow generation from natural language specifications examine whether language models can generate executable workflows directly from user instructions. However, these works inherently assume that the intended procedure is largely specified in the input. In a realistic and challenging deep research, by contrast, the workflow is rarely given explicitly. It must be predicted from heterogeneous evidence distributed across artifacts such as documents, emails, and chat logs. Moreover, such workflows also need to be personalized because company artifacts may describe the generic process broadly applicable for various scenarios (e.g., \textit{if appointment is tenure-track, obtain Provost approval; otherwise, obtain Department Head approval} in Figure~\ref{fig:intro_question_example}), while personal artifacts determine which branch of the workflow applies to the user's situation, e.g., which conditions and next actions are appropriate. This distinction matters because many requests, such as policy compliance, eligibility determination, and resolving client issues follow a sequence of personalized steps. This task of personalized workflow prediction is not captured by existing benchmarks (Table~\ref{tab:comparison}).

To address this gap, we introduce \drflow, a benchmark for personalized workflow prediction in deep research settings. Each task combines company-side artifacts, which define a generic process, along with user-side artifacts, which provide the personal evidence required to transform that process to a personalized workflow. Solving a task therefore requires joint reasoning steps: predicting a generic workflow from generic evidence (present in the company artifacts) and grounding that workflow in personal evidence (present in the user-specific artifacts) to produce a personalized workflow. \drflow~contains $100$ tasks: $50$ tasks we call as \textit{original} and the other $50$ we call as \textit{mixed}. The \textit{original} tasks contain artifacts that were specifically synthesized for the corresponding deep research questions. On the other hand, while the \textit{mixed} contains artifacts from multiple deep research questions, making the tasks more challenging.

We synthesized these tasks using an end-to-end data generation pipeline with human verification to ensure quality of the tasks. Note that this data generation pipeline is amenable to any domain. To evaluate personalized deep research workflows in this setting, we introduce seven metrics that measure factual grounding, step prediction, workflow topology, condition resolution, and personalization quality. We further introduce \textbf{\drflow-Agent} (DRFA), a workflow-oriented agent tailored to this problem setup. We provide the overall \drflow~pipeline in Figure~\ref{fig:task_evaluation_pipeline}.

Our contributions are: \textbf{(1)} we introduce \drflow, the first benchmark centered on personalized workflow prediction containing 100 tasks, across 5 domains  with $1{,}246$ reference steps and more than $3,900$ sources/artifacts; \textbf{(2)} we develop a multi-step, extensible pipeline that produces synthesized yet realistic deep research tasks for personalized workflow generation; \textbf{(3)} we propose seven evaluation metrics that explicitly assess structural and personalization-specific properties of predicted workflows; and \textbf{(4)} we introduce DRFA, a workflow-oriented reference agent for personalized workflow prediction, and show that while it improves over baselines (upto 10.02\% in average F1), substantial room for improvement remains, underscoring the challenges of personalized workflow prediction for the frontier models.

\begin{table*}[t]
\centering
\small
\resizebox{0.8\textwidth}{!}{%
\begin{tabular}{lccccccc}
\toprule
\textbf{Benchmark} & \textbf{\# Tasks} & \textbf{Private Data} & \textbf{Het.\ Apps} & \textbf{Personalized} & \textbf{Actionable Steps} & \textbf{Structured Output} & \textbf{Extensible Pipeline} \\
\midrule
\multicolumn{8}{l}{\textit{\textbf{Deep Research Benchmarks}}} \\
\midrule
GAIA~\cite{mialon2024gaia} & 466 & \xmark & \xmark & \xmark & \xmark & \xmark  & \xmark \\
DeepResearch Bench~\cite{du2025deepresearchbench} & 100 & \xmark & \xmark & \xmark & \xmark & \xmark & \xmark \\
ResearcherBench~\cite{xu2025researcherbench} & 65 & \xmark & \xmark & \xmark & \xmark & \xmark & \xmark \\
Mind2Web 2~\cite{gou2025mind2web2} & 130 & \xmark & \xmark & \xmark & \xmark & \xmark & \xmark \\
DRACO~\cite{zhong2026draco} & 100 & \xmark & \xmark & \xmark & \xmark & \xmark & \cmark \\
\textsc{ResearchRubrics}~\cite{sharma2025researchrubrics} & 101 & \xmark & \xmark & \xmark & \xmark & \xmark & \xmark \\
\midrule
\multicolumn{8}{l}{\textit{\textbf{Agent Benchmarks with Multi-Application Environments}}} \\
\midrule
TheAgentCompany~\cite{xu2024theagentcompany} & 175 & \cmark & \cmark & \cmark & \cmark & \xmark & \xmark \\
OfficeBench~\cite{wang2024officebench} & 300 & \cmark & \cmark & \xmark & \cmark & \xmark & \xmark \\
WorkArena~\cite{drouin2024workarena} & 33 & \cmark & \xmark & \xmark & \cmark & \xmark & \xmark \\
FlowBench~\cite{xiao2024flowbench} & 51 & \xmark & \xmark & \xmark & \cmark & \xmark & \xmark \\
DRBench~\cite{abaskohi2025drbench} & 100 & \cmark & \cmark & \cmark & \xmark & \xmark & \cmark \\
\midrule
\multicolumn{8}{l}{\textit{\textbf{Workflow Generation}}} \\
\midrule
AutoFlow~\cite{li2024autoflow} & --- & \xmark & \xmark & \xmark & \cmark & \cmark & \xmark \\
WorkTeam~\cite{liu-etal-2025-workteam} & 315 & \xmark & \xmark & \xmark & \cmark & \cmark & \xmark \\
Chat2Workflow~\cite{zhong2026chat2workflow} & 27 & \xmark & \xmark & \xmark & \cmark & \cmark & \xmark \\
\midrule
\textbf{\drflow~(Ours)} & \textbf{100 ($\mathbf{1,246}$ reference steps)} & \cmark & \cmark & \cmark & \cmark & \cmark & \cmark \\
\bottomrule
\end{tabular}%
}
\caption{Comparison of \drflow~with representative benchmarks.
\textbf{Private Data}: requires access to private or organizational data beyond the public web.
\textbf{Het.\ Apps}: sources span heterogeneous application types (e.g., chat, email, documents).
\textbf{Personalized}: tasks are grounded in user-specific personas and organizational context.
\textbf{Actionable Steps}: output consists of concrete, followable or executable steps.
\textbf{Structured Output}: output is in structured format, not free-form text.
\textbf{Extensible Pipeline}: includes an automated pipeline for scaling up the benchmark construction.
}
\label{tab:comparison}
\end{table*}

\section{Related Work}
\label{section:related_work}

\noindent \textbf{Deep Research Benchmarks.} DR systems automate complex, multi-step information gathering and synthesis, producing citation-grounded outputs in response to open-ended queries~\cite{zheng2025deepresearcher}. Early efforts such as GAIA~\cite{mialon2024gaia} introduced multi-step, tool-augmented question answering tasks, establishing a foundation for measuring agent reasoning and web browsing capabilities.
DeepResearch Bench~\cite{du2025deepresearchbench} offers $100$ PhD-level tasks across $22$ fields and proposes two evaluation methodologies aligned with human judgment.
DeepResearchGym~\cite{coelho2025deepresearchgym} introduces a reproducible sandbox using open-access web corpora to ensure stable evaluation across runs.
ResearcherBench~\cite{xu2025researcherbench} narrows its focus to frontier AI scientific inquiry, compiling $65$ research questions across $35$ AI subjects.
Mind2Web~2~\cite{gou2025mind2web2} targets long-horizon agentic search with fine-grained evaluation nodes spanning $130$ tasks.
More recently, DRACO~\cite{zhong2026draco} sources $100$ tasks from real user queries across ten domains, grading outputs along four rubric dimensions, and \textsc{ResearchRubrics}~\cite{sharma2025researchrubrics} contributes $101$ expert-curated prompts paired with over $2{,}500$ fine-grained evaluation criteria. However, these benchmarks primarily evaluate reports, answers, or rubric-scored summaries, while \drflow~instead evaluates whether agents can predict structured, personalized workflows as sequences of action-steps (Table~\ref{tab:comparison}).


\noindent \textbf{Multi-Application Agent Benchmarks.} 
DRBench~\cite{abaskohi2025drbench} is the first benchmark to combine public web retrieval with private organizational data sourced from enterprise applications, evaluating agents on insight recall and report quality across $100$ tasks.
TheAgentCompany~\cite{xu2024theagentcompany} provides $175$ professional tasks requiring cross-application coordination and simulated colleague communication within a self-hosted environment.
OfficeBench~\cite{wang2024officebench} evaluates multi-application office automation through customized matching and execution-based evaluation.
WorkArena~\cite{drouin2024workarena} tests browser-based knowledge work on the enterprise platform.
FlowBench~\cite{xiao2024flowbench} specifically benchmarks workflow-guided planning for LLM agents, assessing the efficacy of workflow knowledge across multiple domains. Although these works capture important aspects of enterprise interaction, but they do not directly evaluate whether an agent can recover the relevant procedure from scattered artifacts, resolve user-specific conditions, and output the resulting personalized workflow; which is the core aspect of \drflow (Table~\ref{tab:comparison}).


\noindent \textbf{Workflow Generation for LLM Agents.} Prior work has explored workflow generation as a mechanism for improving LLM-based agents. AutoFlow~\cite{li2024autoflow} proposes a framework for automatically generating natural language workflows through iterative optimization. WorkTeam~\cite{liu-etal-2025-workteam} studies multi-agent workflow construction from natural language task specifications, and Chat2Workflow~\cite{zhong2026chat2workflow} generates executable visual workflows from user instructions. These works, however assume that the target workflow, and/or the information needed to construct, is already provided in the input specification. In contrast, \drflow~evaluates whether an agent can predict information from scattered pieces of evidence into a structured actionable workflow that a human user can follow and complete. This distinction positions \drflow{} at the intersection of deep research and workflow understanding, a space that, to our knowledge, no prior benchmark occupies. Further discussion on literature is provided in Appendix~\ref{appendix:extended_related_work}.


\section{\drflow: A Personalized Deep Research Workflow Prediction Benchmark}
\label{section:drflow}

We introduce \drflow, a benchmark for evaluating agents for \emph{personalized workflow prediction} from heterogeneous data sources. \drflow~comprises $100$ tasks: $50$ \textit{original} and $50$ corresponding \textit{mixed} variants, spanning five domains: B2B (Business-to-Business), B2C (Business-to-Consumer), Education, Healthcare, and Legal. Each task is anchored in a realistic deep research question and requires the agent to reason jointly over two complementary sources of evidence, \emph{company-side data}: which specifies the company's generic operating procedure for the question, including dependencies and different conditions based on the generic process and \emph{user-side data}: which provides the personal context needed to determine which of those conditions hold in the current case and how the generic process should therefore be instantiated.


\paragraph{Task Description.}
\drflow~targets a form of deep research that is fundamentally procedural and personalized. The central challenge is not merely to collect relevant evidence, but to predict a \emph{generic workflow} from company-side data and then transform it into a \emph{personalized workflow} using user-side data. For both company- and user-side data: we further create two types of evidence units: insights and distractors. Insights are relevant pieces of information required to predict the workflows, whereas distractors comprise realistic, domain-adjacent information that increases noise in the evidence-space, turning the retrieval of the workflow-bearing insights into a needle-in-the-haystack challenge. Hence, to solve a task, the agent must explore the full collection of company- and user-side artifacts and predict a generic workflow that preserves the underlying organizational logic, while correctly resolve user-specific conditions.

\begin{figure*}[!ht]
    \centering
    \includegraphics[width=\textwidth]{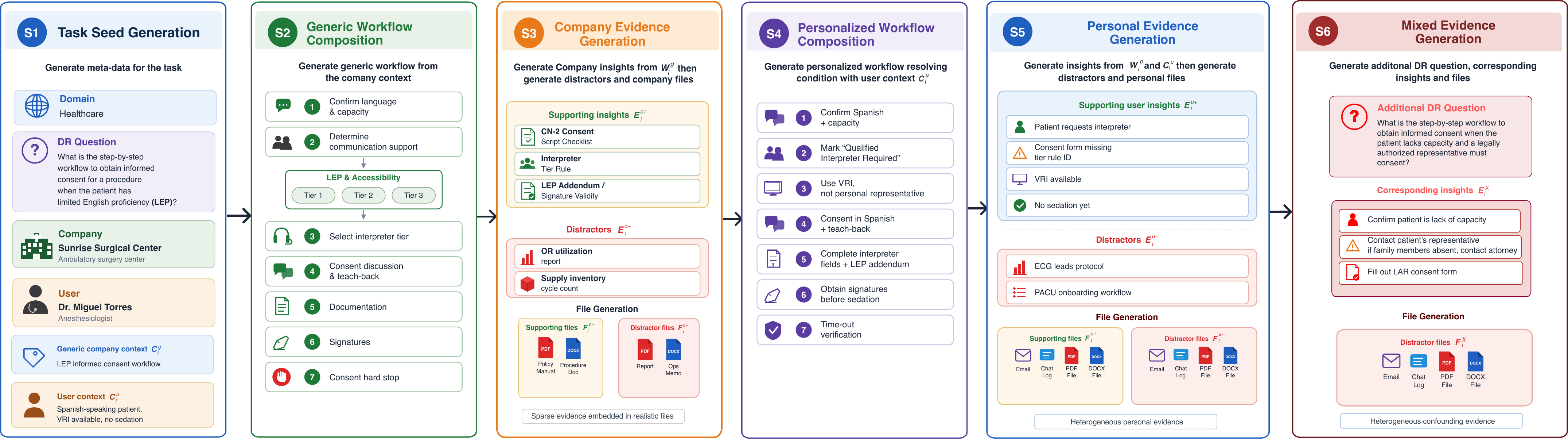}
    \caption{
    Overview of the \drflow~data generation pipeline. Starting from a task seed \textbf{(S1)}, the pipeline first composes a generic company workflow with company and user contexts \textbf{(S2)}. It then generates company-side insights and distractors as heterogeneous artifacts \textbf{(S3)}. The generic workflow is further specialized into a personalized workflow \textbf{(S4)}, after which user-side insights and distractors are generated as personal artifacts \textbf{(S5)}. Finally, similar but distinct questions with corresponding artifacts are generated for the mixed split \textbf{(S6)}.
    }
    \label{fig:data_generation_pipeline}
\end{figure*}

\paragraph{Data Generation.}
\label{section:data_generation}

We construct \drflow~through a staged synthesis pipeline (Figure~\ref{fig:data_generation_pipeline}) that progressively transforms a domain specification into a realistic, personalized deep research environment. 

\paragraph{Stage S1: Task Seed Generation.}
The first stage generates a \emph{task seed} for each benchmark instance. Starting from a target domain and a task $i$, we jointly synthesize the deep research question $Q_i$, structured company metadata, a company persona, and a user persona. Rather than generating the question in isolation, we synthesize it jointly with the company, user, and relationship context that define the scenario. This coupling yields tasks that are grounded simultaneously in a plausible organization, a realistic user role, and a concrete relationship between them.\\
\noindent From this seed, we further instantiate two complementary contextual views. The first is the \emph{generic company context} $C_i^{g}$, which describes the company-level process environment relevant to $Q_i$. The second is the \emph{user context} $C_i^{u}$, which captures the user’s role, circumstances, constraints, and surrounding needs. $C_i^{g}$ provides the process-level foundation from which the generic workflow will be derived, whereas $C_i^{u}$ provides the personalized signals that will later determine how that workflow should be specialized.

\paragraph{Stage S2: Generic Workflow Composition.}
Given the company-side context $C_i^{g}$ and the question $Q_i$, the second stage composes the \emph{generic workflow} $W_i^{g}$. This workflow is a structured sequence of company-level actions and decision points required to address the question. Crucially, $W_i^{g}$ encodes prerequisite relations and different conditions across steps based on the generic context. In other words, it specifies not only what typically needs to be done, but also where the process can branch and under what circumstances alternative actions may become appropriate.


\paragraph{Stage S3: Company Evidence Generation.}

We derive \emph{company supporting insights} $E_i^{c+}$, which encode the substantive content of the generic workflow, including step descriptions and the conditions attached to different parts of the process. In parallel, we generate \emph{company distractors} $E_i^{c-}$, which are realistic and domain-consistent but do not contribute to solving the task. The resulting company insights and distractors are then materialized as company-related files $F_i^{c+}$ and $F_i^{c-}$, respectively. The file generation module uses a three-stage sparse-evidence injection process: it first creates the target artifact type (e.g., PDF or DOCX), then injects the insights or distractors in a contextually appropriate location within the artifact, and finally fills the remaining sections with realistic but task-irrelevant content. This process yields naturalistic company artifacts in which the evidence needed to reconstruct the workflow is embedded within realistic surrounding content rather than explicitly foregrounded, requiring the agent to infer the generic process from distributed sources.

\paragraph{Stage S4: Personalized Workflow Composition.}
Given the generic workflow $W_i^{g}$ and the user context $C_i^{u}$, we compose the \emph{personalized workflow} $W_i^{p}$. Here, $C_i^{u}$ supplies the user-specific facts needed to determine which conditions in $W_i^{g}$ hold for the target user. This stage resolves the branching structure in the generic workflow and instantiates the correct case-specific sequence of actions, which is subsequently verified by human annotators. The resulting personalized workflow $W_i^{p}$ serves as the reference workflow for evaluation.


\paragraph{Stage S5: Personal Evidence Generation.}

Given the personalized workflow $W_i^{p}$ and the user context $C_i^{u}$, we first derive \emph{user supporting insights} $E_i^{u+}$, which encode the personal facts that imply which specific branches are taken and which actions are appropriate in $W_i^{p}$ for the user. We also generate \emph{user distractors} $E_i^{u-}$, which remain realistic but do not help resolve the target task. These user-side insights and distractors are then realized as heterogeneous personal artifacts (e.g., email, chat, local document) $F_i^{u+}$ and $F_i^{u-}$. We follow the same file generation process described in Stage~S3.


\begin{figure*}[t]
    \centering
    \includegraphics[width=0.8\textwidth]{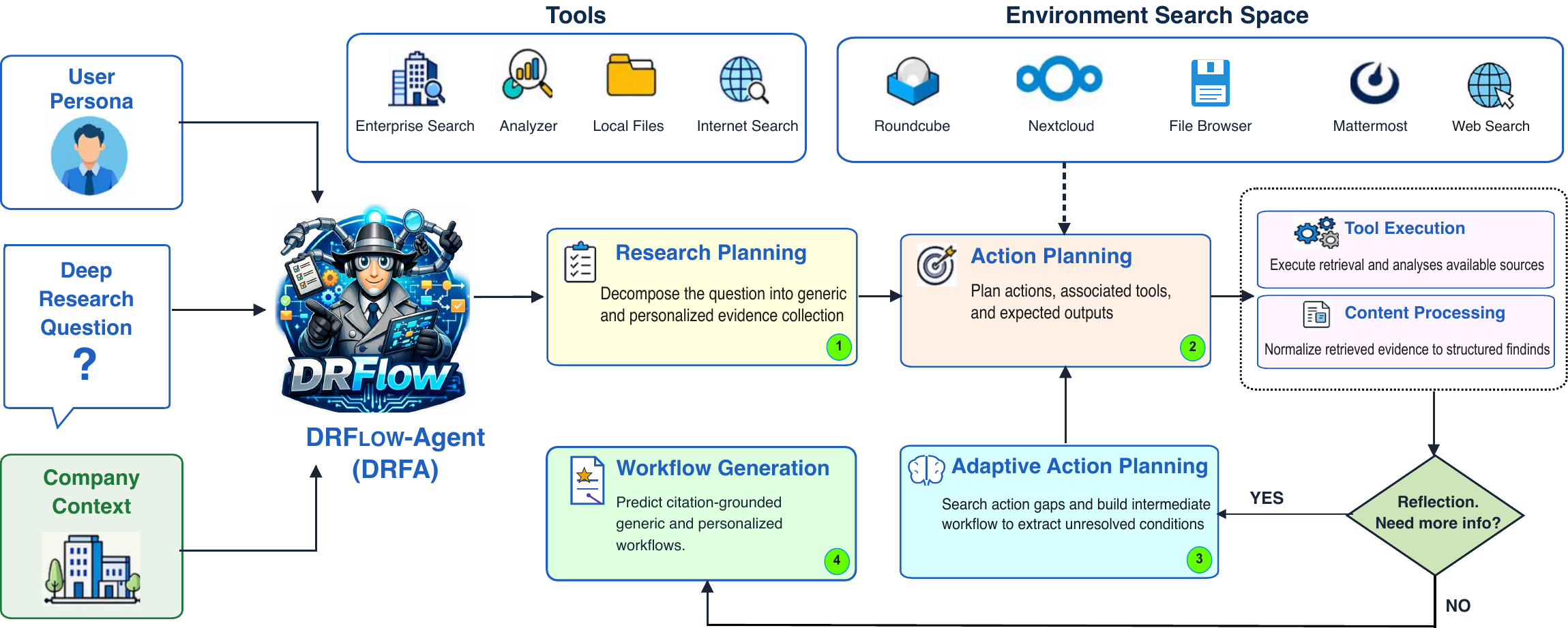}
    \caption{
    Overview of \drflow-Agent~(DRFA). Given the user persona, deep research question, and company information, \textbf{(1)} DRFA first performs research planning. \textbf{(2)} Then it transforms them into corresponding action planning. \textbf{(3)} It uses adaptive action planning to identify gaps and extract unresolved conditions. \textbf{(4)} Finally, it predicts both generic and personalized workflows.
    }
    \label{fig:drfa_agent}
\end{figure*}


\paragraph{Stage S6: Mixed Evidence Generation.}
The five stages above produce a single coherent case per task. To stress-test agents under realistic \emph{case-level disambiguation}, we additionally construct a mixed split in which two additional distinct cases in the same workflow domain are injected into the evidence pool. Given the task seed $(Q_i, C_i^{g}, C_i^{u})$ and the existing user insights, we sample two similar but distinct DR questions that share the workflow domain and stakeholder context with $Q_i$ but pertain to a different case. For each question, we derive a set of coherent insights $E_i^{X}$ that form an internally consistent case profile distinct from $C_i^{u}$. 
Finally, the resulting confounding evidence is realized as heterogeneous files $F_i^{X}$ across the same modalities used in Stages~S3 and~S5. 

\noindent We provide the prompts for the data generation process in Appendix~\ref{appendix:data_generation}, the complete list of tasks in Appendix~\ref{appendix:task_list}, examples of generic and personalized workflow in Appendix~\ref{appendix:workflow_examples}, and screenshots of generated files in Appendix~\ref{appendix:screenshot_task_app}.

\begin{table*}[t]
\centering
\small
\setlength{\tabcolsep}{5pt}
\resizebox{0.6\textwidth}{!}{
\begin{tabular}{llcccc|ccc}
\toprule
\textbf{Backbone} & \textbf{Agent} & \textbf{Factuality}  & \textbf{Topology} & \textbf{Cond.~Res.} & \textbf{Pers.~Comp.} & \textbf{Recall} & \textbf{Precision} & \textbf{F1}\\
\midrule
GPT-5.2 & Basic  & $59.46$ & $47.50$ & $2.86$  & $12.86$ & $24.52$ & $9.79$  & $12.16$ \\
        & DRBA   & $\mathbf{90.01}$  & $89.63$ & $24.85$ & $44.99$ & $60.43$ & $52.11$ & $54.17$\\
        & DRFA   & $85.99$  & $\mathbf{92.22}$ & $\mathbf{42.83}$ & $\mathbf{59.26}$ & $\mathbf{69.43}$ & $\mathbf{62.34}$ & $\mathbf{63.90}$\\
\midrule
Claude-Opus-4.5 & Basic  & $50.10$  & $70.07$ & $1.64$  & $10.29$ & $27.93$ & $10.62$ & $13.54$\\
                & DRBA   & $\mathbf{85.36}$  & $\mathbf{88.47}$ & $25.99$ & $41.89$ & $65.19$ & $44.76$ & $51.09$\\
                & DRFA   & $83.31$  & $88.13$ & $\mathbf{47.18}$ & $\mathbf{62.50}$ & $\mathbf{72.07}$ & $\mathbf{55.87}$ & $\mathbf{60.70}$\\
\midrule
Gemini-3.1-pro & Basic  & $38.59$  & $38.00$ & $0.11$  & $6.49$ & $13.51$ & $4.77$  & $5.05$ \\
               & DRBA   & $\mathbf{90.24}$  & $83.12$ & $14.90$ & $30.38$ & $55.94$ & $37.07$ & $40.99$\\
               & DRFA   & $87.16$ & $\mathbf{91.20}$ & $\mathbf{26.98}$ & $\mathbf{48.08}$  & $\mathbf{67.31}$ & $\mathbf{42.55}$ & $\mathbf{52.05}$\\
\midrule
DeepSeek-v3.2 & Basic  & $43.80$  & $55.33$ & $1.17$  & $9.45$ & $19.04$ & $6.97$  & $8.61$\\
              & DRBA   & $93.57$  & $89.43$ & $15.02$ & $33.51$ & $60.48$ & $38.89$ & $45.37$\\
              & DRFA   & $\mathbf{94.92}$  & $\mathbf{91.74}$ & $\mathbf{35.85}$ & $\mathbf{56.32}$ & $\mathbf{68.40}$ & $\mathbf{48.45}$ & $\mathbf{55.10}$\\
\midrule
Average & Basic  & $47.99$   & $52.72$ & $1.45$  & $9.77$ & $21.25$ & $8.04$  & $9.84$\\
        & DRBA   & $\mathbf{89.80}$  & $87.66$ & $20.19$ & $37.69$ & $60.51$ & $43.21$ & $47.91$\\
        & DRFA   & $87.84$  & $\mathbf{90.82}$ & $\mathbf{38.21}$ & $\mathbf{56.54}$ & $\mathbf{69.30}$ & $\mathbf{52.31}$ & $\mathbf{57.93}$\\
\bottomrule
\end{tabular}
}
\caption{Main results on the original split across backbone models. Within each backbone block, the best score in each column is bolded.}
\label{tab:main_results}
\end{table*}

\begin{table*}[t]
\centering
\small
\setlength{\tabcolsep}{4.2pt}
\resizebox{0.7\textwidth}{!}{
\begin{tabular}{llcccc|ccc}
\toprule
\textbf{Model} & \textbf{Agent} & \textbf{Factuality}  & \textbf{Topology} & \textbf{Cond.~Res.} & \textbf{Pers.~Comp.} & \textbf{Recall} & \textbf{Precision} & \textbf{F1}\\
\midrule
GPT-5.2 & DRBA & $86.41$  & $86.52$ & $24.58$ & $42.61$ & $58.13$ & $37.30$ & $44.36$\\
        & DRFA & $\mathbf{94.38}$  & $\mathbf{90.04}$ & $\mathbf{41.36}$ & $\mathbf{55.67}$ & $\mathbf{65.86}$ & $\mathbf{47.96}$ & $\mathbf{53.67}$\\
\midrule
Claude-Opus-4.5 & DRBA & $\mathbf{86.78}$  & $\mathbf{93.10}$ & $23.59$ & $39.56$ & $63.70$ & $37.86$ & $44.52$\\
                & DRFA & $82.54$  & $89.68$ & $\mathbf{46.72}$ & $\mathbf{58.51}$ & $\mathbf{68.35}$ & $\mathbf{47.43}$ & $\mathbf{53.19}$\\
\midrule
Gemini-3.1-pro & DRBA & $\mathbf{79.10}$ & $\mathbf{86.00}$ & $12.92$ & $33.54$ & $47.75$ & $\mathbf{24.24}$ & $30.18$\\
               & DRFA & $72.36$ & $85.32$ & $\mathbf{18.86}$ & $\mathbf{42.32}$ & $\mathbf{49.26}$ & $24.07$ & $\mathbf{30.42}$ \\
\midrule
DeepSeek-v3.2 & DRBA & $81.32$ & $\mathbf{83.13}$ & $16.12$ & $27.26$ & $48.55$ & $30.36$ & $34.04$ \\
              & DRFA & $\mathbf{82.30}$ & $82.03$ & $\mathbf{28.03}$ & $\mathbf{51.83}$ & $\mathbf{51.35}$ & $\mathbf{32.32}$ & $\mathbf{38.34}$\\
\midrule
Average & DRBA & $\mathbf{83.40}$ & $\mathbf{87.19}$ & $19.30$ & $35.74$ & $54.53$ & $32.44$ & $38.27$  \\
        & DRFA & $82.89$ & $86.77$ & $\mathbf{33.74}$ & $\mathbf{52.08}$ & $\mathbf{58.70}$ & $\mathbf{37.95}$ & $\mathbf{43.91}$\\
\bottomrule
\end{tabular}
}
\caption{Results on the mixed split of DRFA. Within each backbone model, the best score in each column is bolded.}
\label{tab:adv_results}
\end{table*}

\section{\textsc{DRFA}: The Workflow Prediction Agent}

We introduce \drflow-Agent, namely \textsc{DRFA}, a workflow prediction agent for personalized deep research. \textsc{DRFA} (Figure~\ref{fig:drfa_agent}) is designed for the core challenge posed by \drflow: given a DR question, the agent must predict the company's generic procedure from company-side evidence, identify the user-specific conditions, and instantiate the final personalized workflow using user-specific evidence. DRFA operates through four components:

\paragraph{Research Planning.}
The first component decomposes the deep research question into a set of investigation areas that specify what evidence should be collected before workflow prediction. In \textsc{DRFA}, this stage is organized around two complementary objectives: \emph{Generic Requirement Collection}, which targets company-level policy documents, procedures that define the generic process; and \emph{Personal Evidence Collection}, which targets personal artifacts to resolve user-dependent conditions.

\paragraph{Action Planning.}
The second component translates each investigation area into concrete actions. Each action specifies a retrieval or analysis objective, along with the expected outputs. Action execution proceeds through the appropriate tool-calling until a fixed number of iterations or all actions are executed.

\paragraph{Adaptive Action Planning.}
At the end of every iteration, \textsc{DRFA} analyzes the newly collected findings and determines whether additional actions are needed. The module serves two functions. First, it performs \emph{gap finding}, assessing whether the current evidence over-represents either generic workflow or user-specific evidences, and proposing additional actions to cover the missing side. Second, it performs \emph{conditional action planning}: the agent derives an intermediate generic workflow from the currently available policy and workflow evidence, identifies the unresolved conditional branches, and formulates targeted actions to retrieve the personalized facts to resolve them.\\

\paragraph{Workflow Generation.}
The fourth component predicts the final personalized workflow from the collected evidence. \textsc{DRFA} first 
constructs a complete generic workflow, preserving prerequisite structure and alternative branches. It then uses the collected user-specific evidence to resolve conditional branches and produce the final personalized recommended workflow. Further details of DRFA are provided in Appendix~\ref{appendix:drfa_details}.

\section{Experimental Settings}
\label{section:experimental_settings}

\paragraph{Baselines.}
We compare DRFA~against two baseline agents: \textbf{Basic} is a minimal agent without an explicit planning stage. It ingests all available documents and directly predicts a workflow for the deep research question. 
\textbf{DRBench Agent (DRBA)}~\cite{abaskohi2025drbench} is a stronger planning-based baseline adapted from DRBench. 
 Using DRBA as a baseline allows us to evaluate whether DRFA~improves workflow generation beyond prior enterprise deep research pipelines built around staged planning and adaptive execution. 
\noindent We instantiate the agents with multiple frontier models, including GPT-5.2, Claude-Opus-4.5, Gemini-3.1-pro-preview, and DeepSeek-V3.2. We provide further details on implementation in Appendix~\ref{appendix:implementation_details} and approximate cost estimation in Appendix~\ref{appendix:cost_estimation}. To simulate realistic setting, we run the applications inside a containerized environment (Appendix~\ref{appendix:application_environment}).

\begin{table*}[t]
\centering
\small
\setlength{\tabcolsep}{4.5pt}
\resizebox{0.7\textwidth}{!}{
\begin{tabular}{llcccc|ccc}
\toprule
\textbf{Backbone} & \textbf{Variant} & \textbf{Factuality}  & \textbf{Topology} & \textbf{Cond.~Res.} & \textbf{Pers.~Comp.} & \textbf{Recall} & \textbf{Precision} & \textbf{F1}\\
\midrule
GPT-5.2 & DRFA & $89.82$  & $\mathbf{92.82}$ & $41.43$ & $\mathbf{59.36}$ & $64.82$ & $\mathbf{54.58}$ & $\mathbf{58.35}$\\
& w/o GF & $89.44$  & $85.21$ & $\mathbf{45.84}$ & $58.51$ & $70.69$ & $48.24$ & $56.53$\\
& w/o CAP & $\mathbf{90.22}$ & $85.16$ & $41.34$ & $55.73$ & $\mathbf{71.94}$ & $49.31$ & $57.21$\\
& w/o Both & $88.42$  & $86.64$ & $35.07$ & $57.29$ & $69.23$ & $50.94$ & $56.11$\\
\midrule
Claude-Opus-4.5 & DRFA & $84.57$  & $89.32$ & $\mathbf{47.09}$ & $\mathbf{62.95}$ & $\mathbf{71.40}$ & $\mathbf{48.07}$ & $\mathbf{56.35}$\\
& w/o GF & $\mathbf{85.36}$  & $89.98$ & $45.34$ & $58.44$ & $70.76$ & $46.90$ & $55.48$\\
& w/o CAP & $85.33$  & $\mathbf{90.57}$ & $42.53$ & $61.00$ & $68.85$ & $45.56$ & $52.12$\\
& w/o Both & $81.67$  & $82.81$ & $39.10$ & $54.99$ & $64.56$ & $44.78$ & $50.33$\\
\midrule
Gemini-3.1-pro & DRFA & $83.27$  & $\mathbf{89.61}$ & $\mathbf{31.45}$ & $\mathbf{45.35}$ & $\mathbf{68.58}$ & $\mathbf{45.31}$ & $\mathbf{53.88}$\\
& w/o GF & $85.09$  & $85.82$ & $27.83$ & $42.75$ & $64.40$ & $29.70$ & $39.26$\\
& w/o CAP & $\mathbf{86.44}$  & $86.75$ & $21.69$ & $42.50$ & $67.42$ & $33.79$ & $44.37$\\
& w/o Both & $79.46$  & $54.42$ & $15.20$ & $28.32$ & $46.40$ & $23.67$ & $30.98$\\
\midrule
DeepSeek-v3.2 & DRFA & $\mathbf{84.28}$  & $85.74$ & $\mathbf{37.85}$ & $\mathbf{46.32}$ & $\mathbf{65.32}$ & $\mathbf{49.73}$ & $\mathbf{55.88}$\\
& w/o GF & $83.28$  & $84.22$ & $28.31$ & $42.37$ & $54.34$ & $47.91$ & $50.14$\\
& w/o CAP & $83.45$  & $\mathbf{87.33}$ & $34.13$ & $29.08$ & $63.57$ & $48.41$ & $53.38$\\
& w/o Both & $76.34$  & $80.14$ & $14.74$ & $22.82$ & $45.78$ & $34.56$ & $37.96$\\
\midrule
Average & DRFA & $85.48$  & $\mathbf{89.37}$ & $\mathbf{39.45}$ & $\mathbf{53.50}$ & $67.53$ & $\mathbf{49.42}$ & $\mathbf{56.12}$\\
& w/o GF & $85.79$  & $86.31$ & $36.83$ & $50.52$ & $65.05$ & $43.19$ & $50.35$\\
& w/o CAP & $\mathbf{86.36}$  & $87.45$ & $34.92$ & $47.08$ & $\mathbf{67.94}$ & $44.27$ & $51.77$\\
& w/o Both & $81.47$  & $76.00$ & $26.03$ & $40.86$ & $56.49$ & $38.49$ & $43.85$\\
\bottomrule
\end{tabular}
}
\caption{Ablation study on planning modules. Within each backbone model, the best score in each column is bolded.}
\label{tab:ablation_modules}
\end{table*}

\paragraph{Evaluation Metrics.}
We report seven metrics:\\
\noindent \textbf{Factuality}~\cite{abaskohi2025drbench} verifies whether a predicted step is supported by its cited evidence.\\
\noindent \textit{Recall} quantifies the proportion of reference workflow steps that are successfully recovered by the predicted workflow, whereas \textit{Precision} measures the proportion of predicted steps that can be validly aligned to reference steps. \textbf{F1} is computed as the task-wise harmonic mean of Recall and Precision.\\
\noindent \textit{Condition Resolution} measures whether the predicted step resolves the condition of a step using the appropriate personalized evidence.\\ \noindent \textit{Topology} evaluates pairwise order consistency between aligned predicted and reference steps.\\
\noindent \textit{Personalized Comprehensiveness} measures whether the predicted workflow preserves the personalized and operational details present in the reference workflow, such as specific names, forms, dates, thresholds, and other user-specific constraints. 
We provide additional details for the evaluation metrics in Appendix~\ref{appendix:evaluation_metrics}.

\section{Results}
\label{section:results}
\paragraph{Performance on Original Split.}
Table~\ref{tab:main_results} reports the main results on the original split of $50$ tasks across four backbone models. On aggregate, DRFA~achieves the best performance on all the metrics, except Factuality, where DRFA~is slightly worse due to the instability caused by heterogeneous planning~(see  \cite{abaskohi2025drbench}) and synthesis of multiple documents, making it harder for the judge to verify (See Appendix~\ref{appendix:drfa_factuality}).\\
\noindent DRFA~has a stronger average F1 over DRBA  ($57.93$ vs $47.91$), Condition Resolution  ($38.21$ vs $20.19$), and Personalized Comprehensiveness  ($56.54$ vs $37.69$). This indicates that DRFA~produces workflows that are more accurate,  structurally better coordinated, and more personalized.\\
 Finally, we stress that the overall performance also underscores the importance of the \drflow~\textit{benchmark}: despite the strength of existing frontier models, absolute performance still leaves considerable room for improvement, indicating that personalized workflow prediction is a challenging problem and that \drflow~provides a meaningful testbed for measuring further progress. We provide the domain-wise performance in Appendix~\ref{appendix:domain_results} and results  on mean with deviation in Appendix~\ref{appendix:mean_std_results}.

\paragraph{Performance on the Mixed Split.}
Table~\ref{tab:adv_results} reports  the results averaged across the four backbones.
Both DRFA and DRBA degrade across nearly all metrics, confirming the challenging nature of the split. However, DRFA retains its lead on all the metrics for all the backbones, under this stress test, except for Factuality and Topology. Compared to Table~\ref{tab:main_results}, average DRFA's F1 falls from $57.93$ to $43.91$, Condition Resolution from $38.21$ to $33.74$, and Personalized Comprehensiveness from $56.54$ to $52.08$.  \emph{The persistence of sizeable gaps between the original and mixed splits on all the metrics underscores the challenging nature of robust personalized workflow prediction.} Performance comparison between the original and mixed splits is shown in Appendix~\ref{appendix:performance_original_mixed}.

\section{Ablation Study on DRFA Design}

We select a subset of $15$ tasks ($3$ from each domain) as ablation set to keep the cost manageable. We study the contribution of the two planning components in DRFA: \emph{gap finding} (GF) and \emph{conditional action planning} (CAP) in Table~\ref{tab:ablation_modules}. Removing GF decreases the average F1  from $56.12$ to $50.35$, while removing CAP decreases it to $51.77$. The effect of CAP is especially clear on \emph{Condition Resolution}, where removing it  lowers the average score from $39.45$ to $34.92$. This behavior is expected since CAP explicitly constructs intermediate workflows to identify unresolved conditions and then generates targeted action plans to resolve them. Finally, with both components removed, F1 drops to $43.85$, indicating both are jointly important, supporting the full DRFA design. We provide further ablation studies including recent frontier models' performance in Appendix~\ref{appendix:ablation_studies}, qualitative analyses in Appendix~\ref{appendix:extended-qualitative-analysis}, and error analyses in Appendix~\ref{appendix:error_analyses}. 

\section{Human Evaluation}
\label{section:human_evaluation}

We conduct human evaluation to validate our benchmark  and the reliability of our LLM-based evaluator. For DR questions selection, three annotators independently reviewed $50$ candidate tasks for realism and applicability to the associated company, persona, and workflow setting. Annotators unanimously agreed on $33$ tasks in the initial round, and the remaining tasks were revised and reviewed in a second round before inclusion.

\noindent Then we conduct human evaluation on $10$ tasks across the outputs of DRFA and DRBA from GPT-5.2, yielding $250$ predicted-step decisions. All three annotators agree on $57.2\%$ decisions, and at least two  agree on $97.6\%$ decisions. The resulting Fleiss' $\kappa$ is $0.663$, implying substantial agreement~\cite{landis1977measurement}. We further compare human evaluation with the LLM judge. The Pearson correlation between human-average F1 and LLM-judge F1 is $0.883$, indicating strong agreement between expert and automatic evaluation (Please see Appendix~\ref{appendix:human_evaluation} for details).

\section{Conclusion}
\label{sec:conclusion}

We introduce \drflow, a benchmark for personalized workflow prediction for deep research. Unlike prior works that primarily emphasize free-form report generation, \drflow~targets a procedural setting in which agents must predict a generic workflow from heterogeneous company artifacts and then ground it in user-specific evidence to produce a personalized workflow. 

\noindent We further present \drflow-Agent (DRFA) as a workflow-oriented baseline and evaluate it across multiple frontier backbone models. The results show that workflow-aware agent design yields noticeable gains. However, substantial headroom remains across core metrics, indicating that personalized workflow prediction continues to be a challenging capability for current systems. 

\section{Limitations and Ethics Statement}

\subsection{Limitations}

\paragraph{Benchmark Scope.}
\drflow~focuses on personalized workflow extraction from heterogeneous enterprise-style sources. Although the benchmark spans multiple domains and covers realistic procedural settings, it remains limited to a finite set of task families, domains, application types, and workflow structures. Consequently, the benchmark should not be interpreted as exhaustive coverage of all enterprise deep research workflows. Future work may extend the setting to longer-running workflows, collaborative workflows involving multiple users, temporally evolving evidence, and scenarios where policies or user contexts change during task execution.

\paragraph{Synthetic Data Generation.}
\drflow~uses an automated data generation pipeline to construct task contexts, company- and user-side workflow insights, distractors, and heterogeneous artifacts. This design enables extensible benchmark construction, but it also introduces the usual limitations of synthetic data. The generated companies, personas, documents, emails, and chat messages may not fully capture the ambiguity and institutional variation found in real deployments.

\paragraph{Evaluation Granularity.}
Our evaluation decomposes model outputs into workflow steps and measures factuality, recall, precision, F1, topology order, condition resolution, and personalized comprehensiveness. This step-level design provides a more diagnostic view than holistic report scoring, but it does not capture every aspect of workflow usefulness. For example, two workflows may receive similar scores while differing in clarity, communication, or downstream executability.

\paragraph{LLM-as-a-Judge Dependence.}
Several metrics rely on LLM-based judgments for semantic alignment, factual support, condition resolution, and personalization quality. Although the prompts are designed to be strict and structured, LLM judges can still introduce variance, domain-specific terminology, or partially correct condition resolutions.

\paragraph{Model Performance.}
The empirical results show that current frontier models still struggle with personalized workflow prediction. Even the strongest \drflow~agent configurations leave substantial gaps in F1, condition resolution, and personalized comprehensiveness, indicating that models often fail to recover complete workflows and predict user-specific branches. The mixed split further compresses core workflow metrics, suggesting that robust personalized workflow prediction remains difficult under stress test. These findings highlight substantial room for improvement and pave the way for future research on structured, grounded, and personalized workflow prediction.

\paragraph{Ablation Cost.}
We conduct the module ablation study on a representative subset of $15$ tasks to keep computational and API costs manageable. This subset is sufficient to perform the controlled experiments, but it may not capture the full variance of module behavior across all domains and task difficulties. The ablation results should therefore be interpreted as diagnostic evidence rather than a full benchmark-wide estimate.

\paragraph{Reproducibility.}
\drflow~is evaluated in a controlled, containerized multi-application environment. This setup improves experimental reproducibility by fixing the application stack, preserving source boundaries, and allowing tasks to be re-instantiated consistently across runs. However, reproducibility may still be affected by stochastic model behavior, API-level changes in closed-source backbones, and variance introduced by LLM-based evaluation. We therefore report task and implementation details, prompts, model choices, and repeated-run statistics to support reproducibility. Furthermore, we will release the benchmark and the code implementation upon acceptance.

\subsection{Ethics Statement}

\paragraph{Data Privacy.}
\drflow~is designed to simulate enterprise deep research over private-like data while avoiding the use of real personal or confidential enterprise records. The benchmark artifacts, personas, companies, documents, emails, and chats are synthetically generated for research purposes. This design reduces the risk of exposing private information or enabling re-identification while preserving the structural challenges of searching across heterogeneous organizational and personal evidence.

\paragraph{Responsible Use of Models Output.}
Several \drflow~domains, including healthcare, legal, education, and compliance-oriented business workflows, involve sensitive decision contexts. Although the benchmark data is synthetic, the task format resembles scenarios in which real users may rely on generated workflows for consequential decisions. Systems evaluated on \drflow~should therefore be used with caution in such settings. Their outputs should be treated as decision-support artifacts that require review by qualified professionals, especially when workflows concern legal obligations, clinical procedures, privacy rights, or compliance actions.

\paragraph{Potential Misuse.}
A workflow extraction agent that can search across enterprise sources and synthesize personalized procedures could be misused to expose private organizational information, infer sensitive user attributes, or automate actions beyond a user's authorization. \drflow~is intended to advance evaluation of grounded, structured, and personalized workflow prediction, not to encourage unrestricted access to enterprise data. Practical deployments should enforce access control, source-level permissions, audit logging, citation transparency, and clear separation between retrieval, recommendation, and execution.

\paragraph{Bias and Representational Coverage.}
Because \drflow~relies on synthetic personas, organizations, and domain scenarios, it may encode distributional assumptions from the generation process. These assumptions can affect which roles, industries, user needs, and procedural constraints are represented. Future benchmark construction should therefore consider domain and persona diversity, representational balance, and performance differences across task types and user contexts.

\paragraph{Use of Generative AI.}
Generative AI has been used to polish the writing of the manuscript.

\clearpage
\bibliography{custom}

\appendix

\clearpage
\appendixpage
\addappheadtotoc
\counterwithin{figure}{section}
\counterwithin{table}{section}

\section{Extended Related Work}
\label{appendix:extended_related_work}

\noindent \textbf{Evidence-grounded reasoning and attribution.}
A substantial body of work studies whether models can combine evidence across documents and expose the basis for their predictions. HotpotQA~\cite{yang2018hotpotqa} evaluates explainable multi-hop question answering with supporting facts, while MuSiQue~\cite{trivedi2022musique} constructs multi-hop questions through compositional single-hop dependencies to reduce disconnected reasoning. FEVER~\cite{thorne2018fever} frames evidence use as claim verification against textual sources, requiring systems to classify claims and recover supporting evidence, while ALCE~\cite{gao2023enabling} evaluates long-form generation with citations, measuring not only answer quality but also whether generated statements are adequately supported by cited passages. 

\noindent \textbf{Long-form information-seeking question answering.}
Long-form QA benchmarks further move evaluation beyond short extractive answers. ELI5~\cite{fan2019eli5} studies open-ended explanatory questions requiring multi-sentence responses, while Qasper~\cite{dasigi2021qasper} anchors information-seeking questions in full scientific papers and asks systems to recover evidence from long documents. ASQA~\cite{stelmakh2022asqa} focuses on ambiguous factoid questions whose answers must synthesize multiple valid interpretations, and QAMPARI~\cite{amouyal2023qampari} evaluates questions with many answers distributed across multiple paragraphs. These benchmarks are closely related to deep research because they reward synthesis over dispersed evidence. Nevertheless, they evaluate the quality of textual answers, not whether agents can transform heterogeneous evidence into a followable workflow with prerequisites, branch conditions, and user-specific applicability.

\noindent \textbf{Interactive web and tool-use agents.}
Agent benchmarks have also examined long-horizon interaction with external environments and tools. WebShop~\cite{yao2022webshop} introduced a simulated e-commerce environment for grounded web interaction, and WebArena~\cite{zhou2024webarena} extended this direction to realistic self-hosted websites spanning multiple web domains. AgentBench~\cite{liu2024agentbench} evaluates LLM agents across interactive environments, while API-Bank~\cite{li2023apibank} and ToolLLM~\cite{qin2024toolllm} study tool selection, API invocation, and tool-use training at scale. These works test whether agents can plan and act in environments. In contrast, \drflow~evaluates the workflow prediction capability that precedes or guides such action: the agent must recover the applicable procedure from evidence and express it as a structured user-facing action sequence.

\noindent \textbf{Procedural text understanding.}
A related line of work evaluates whether models understand processes described in text or multimodal instructions. ProPara~\cite{dalvi2018tracking} requires tracking entity state changes across procedural paragraphs, WIQA~\cite{tandon2019wiqa} evaluates counterfactual reasoning over procedural influence graphs, and RecipeQA~\cite{yagcioglu2018recipeqa} studies multimodal comprehension of cooking recipes with temporal and procedural structure. These tasks capture important aspects of temporal ordering, state transitions, and procedural semantics. However, the relevant process is generally provided within a relatively bounded input. \drflow~instead requires agents to infer the process itself from heterogeneous artifacts, decide which conditional branch applies to a particular user scenario, and produce a personalized workflow as the evaluated output.

\noindent \textbf{Positioning of \drflow.}
To summarize, prior works have made substantial progress on evidence attribution, long-form synthesis, interactive tool use, and procedural reasoning. However, these works do not test whether agents can predict personalized, evidence-conditioned workflows scattered in heterogeneous artifacts simulating the real-world setting. \drflow~targets this missing capability by requiring systems to recover procedural structure, preserve action ordering, resolve conditional branches using user-specific evidence, and predict the actionable personalized workflow rather than a free-form narrative.

\section{Implementation Details}
\label{appendix:implementation_details}

We access all backbone models, including both open-source and closed-source models, through the OpenRouter API~\cite{openrouter2026}. For benchmark construction, we use GPT-$5.2$ as the generation model. For evaluation, we use GPT-$4$o as the judge model for prompt-based metrics. For citation-grounded factuality verification, we use \texttt{text-embedding-3-small} to retrieve the top five relevant chunks from cited sources before passing the evidence to the judge model. For comparing against \textbf{DRBench Agent (DRBA)}~\cite{abaskohi2025drbench},   we update the report generation module of the agent, since the original work focuses on long-form report generation instead of workflow prediction. We use the same workflow prediction prompting and output format as DRFA to make the comparison fair.\\
\noindent We run the application layer of the benchmark environment inside Docker containers. In particular, the enterprise-style services used by the agent, including \textbf{Nextcloud}, \textbf{Mattermost}, \textbf{IMAP email}, and the \textbf{file sharing system}, are deployed as containerized services. This setup provides an isolated and reproducible multi-application environment for document retrieval, messaging, email access, and shared-file interaction.

\section{Ablation Studies}
\label{appendix:ablation_studies}

\begin{table*}[t]
\centering
\resizebox{0.8\textwidth}{!}{
\begin{tabular}{ccccc|ccc}
\toprule
\textbf{Iter.} & \textbf{Fact.}  & \textbf{Topology} & \textbf{Cond.~Res.} & \textbf{Pers.~Comp.} & \textbf{Recall} & \textbf{Precision} & \textbf{F1}\\
\midrule
$5$  & $\mathbf{88.89}$  & $88.72$ & $42.36$ & $57.22$ & $58.01$ & $45.26$ & $50.85$\\
$10$ & $85.99$  & $\mathbf{92.22}$ & $42.83$ & $\mathbf{59.26}$ & $\mathbf{64.97}$ & $54.58$ & $58.15$\\
$15$ & $86.83$  & $88.93$ & $\mathbf{46.31}$ & $58.24$ & $63.75$ & $\mathbf{59.32}$ & $\mathbf{61.86}$\\
\bottomrule
\end{tabular}
}
\caption{Ablation on the number of iterations with GPT-5.2 as the backbone. Higher is better for all metrics. The best score in each column is bolded.
\label{tab:ablation_iterations}}
\end{table*}

\subsection{Ablation on DRFA Iteration}
\label{appendix:iteration_budget}

We further study the effect of the iteration budget of DRFA using GPT-5.2 as the backbone. Table~\ref{tab:ablation_iterations} shows that increasing the number of iterations does not yield a monotonic improvement across metrics. Iteration $5$ exhibits the best factuality. This is potentially because it tends to cover fewer action planning steps, resulting in fewer document search and facts referencing. Moving from $5$ to $10$ iterations improves all other metrics, whereas increasing the budget further to $15$ does not preserve these gains uniformly. Although $15$ iterations gives the best Precision, F1, and condition resolution, it is slightly worse than $10$ iterations on other metrics.

Overall, this result shows that increasing the iteration does not guarantee consistent improvements. In particular, the comparison between $10$ and $15$ iterations suggests that additional iterations may slightly improve alignment quality among predicted steps, but do not necessarily improve recovery of the full target workflow or its ordering consistency. Based on this result and also to keep the budget manageable, we use the iteration of $10$ in all experiments throughout this work.

\subsection{Performance Comparison of Recent Frontier Models}

Recently introduced state-of-the-art frontier models are compared on the ablation subset to assess progress in personalized workflow prediction. As shown in Table~\ref{tab:frontier_model_comparison}, DRFA consistently improves over DRBA across the main workflow metrics. Averaged over five backbones, DRFA raises F1 from $62.97$ to $68.06$, with gains in both Recall ($63.78$ to $70.84$) and Precision ($64.94$ to $67.69$). The improvement is also reflected in condition resolving and personalization metrics, where Condition Resolution increases from $33.96$ to $48.16$ and Personalization Comprehensiveness increases from $44.21$ to $60.60$. Among the frontier models, Claude-Opus-4.7 achieves the strongest overall performance, obtaining the highest F1 ($69.14$), Topology Order ($91.24$), and Condition Resolution ($51.83$). GPT-5.5 and Kimi-K2.6 are close in F1, reaching $68.62$ and $68.54$, respectively. The performance of these frontier models further show that the gaps across workflow structure, condition handling, and personalization in personalized workflow prediction remain far from saturated.

\begin{table*}[t]
\centering
\small
\setlength{\tabcolsep}{4.2pt}
\resizebox{\textwidth}{!}{
\begin{tabular}{llcccc|ccc}
\toprule
\textbf{Model} & \textbf{Agent} & \textbf{Factuality}  & \textbf{Topology} & \textbf{Cond.~Res.} & \textbf{Pers.~Comp.} & \textbf{Recall} & \textbf{Precision} & \textbf{F1}\\
\midrule
GPT-5.5 & DRBA & $96.66$  & $88.50$ & $35.46$ & $46.22$ & $66.29$ & $63.75$ & $64.78$\\
        & DRFA & $85.92$  & $90.37$ & $48.94$ & $65.55$ & $71.34$ & $67.22$ & $68.62$\\
\midrule
Claude-Opus-4.7 & DRBA & $83.55$  & $89.44$ & $38.59$ & $50.98$ & $69.06$ & $64.76$ & $65.19$\\
                 & DRFA & $82.63$  & $91.24$ & $51.83$ & $65.67$ & $73.11$ & $68.60$ & $69.14$\\
\midrule
GLM-5.1 & DRBA & $88.55$  & $86.96$ & $33.50$ & $40.16$ & $62.02$ & $64.98$ & $62.30$\\
        & DRFA & $86.19$  & $86.18$ & $46.91$ & $59.59$ & $70.73$ & $65.36$ & $66.57$\\
\midrule
Kimi-K2.6 & DRBA & $92.26$  & $87.55$ & $34.84$ & $45.25$ & $62.30$ & $65.48$ & $62.16$\\
                    & DRFA & $83.51$  & $86.41$ & $49.71$ & $59.38$ & $70.87$ & $68.74$ & $68.54$\\
\midrule
Qwen3.6-plus & DRBA & $77.65$  & $88.45$ & $27.40$ & $38.42$ & $59.23$ & $65.75$ & $61.40$\\
              & DRFA & $82.30$  & $88.51$ & $43.39$ & $52.83$ & $68.14$ & $68.54$ & $67.42$\\
\midrule
Average & DRBA & $87.73$  & $88.18$ & $33.96$ & $44.21$ & $63.78$ & $64.94$ & $63.16$\\
        & DRFA & $84.11$  & $88.54$ & $48.16$ & $60.60$ & $70.84$ & $67.69$ & $68.06$\\
\bottomrule
\end{tabular}
}
\caption{Performance comparison of recent frontier models on the ablation subset of $15$ tasks.}
\label{tab:frontier_model_comparison}
\end{table*}

\section{Results by Domain}
\label{appendix:domain_results}

Table~\ref{tab:domain_results} reports the domain-wise performance across all the backbone models. The same pattern holds across all five domains: DRFA consistently achieves the strongest performance on recall, precision, F1, topology order, condition resolution, and personalized comprehensiveness, while DRBA attains the highest factuality in every domain. This result is consistent with Section~\ref{section:results}: the advantage of DRFA lies in the construction of personalized, condition-aware, and well-ordered workflows.

\begin{table*}[h]
\centering
\small
\setlength{\tabcolsep}{4pt}
\renewcommand{\arraystretch}{1.1}
\resizebox{\textwidth}{!}{
\begin{tabular}{llcccc|ccc}
\toprule
\textbf{Domain} & \textbf{Agent} & \textbf{Factuality} & \textbf{Topology} & \textbf{Cond.~Res.} & \textbf{Pers.~Comp.}  & \textbf{Recall} & \textbf{Precision} & \textbf{F1}\\
\midrule
\multirow{3}{*}{B2B}
& Basic & $44.79$  & $60.00$ & $1.29$ & $8.30$ & $17.92$ & $8.86$ & $9.79$\\
& DRBA  & $\mathbf{85.27}$  & $78.88$ & $16.30$ & $32.06$ & $53.36$ & $34.73$ & $41.28$\\
& DRFA  & $84.24$  & $\mathbf{87.93}$ & $\mathbf{35.38}$ & $\mathbf{48.45}$ & $\mathbf{69.51}$ & $\mathbf{40.81}$ & $\mathbf{50.30}$\\
\midrule
\multirow{3}{*}{B2C}
& Basic & $46.76$  & $52.50$ & $1.64$ & $10.47$ & $20.44$ & $6.58$ & $8.44$\\
& DRBA  & $\mathbf{91.31}$  & $90.15$ & $19.73$ & $36.13$ & $56.34$ & $34.46$ & $42.54$\\
& DRFA  & $88.64$  & $\mathbf{90.95}$ & $\mathbf{37.56}$ & $\mathbf{54.82}$ & $\mathbf{69.53}$ & $\mathbf{45.72}$ & $\mathbf{51.84}$\\
\midrule
\multirow{3}{*}{Education}
& Basic & $44.23$  & $35.00$ & $0.56$ & $5.76$ & $18.62$ & $4.04$ & $5.37$\\
& DRBA  & $\mathbf{92.78}$  & $88.29$ & $22.44$ & $41.53$ & $58.68$ & $33.21$ & $41.07$\\
& DRFA  & $88.06$  & $\mathbf{91.12}$ & $\mathbf{40.15}$ & $\mathbf{60.43}$ & $\mathbf{69.71}$ & $\mathbf{48.66}$ & $\mathbf{56.05}$\\
\midrule
\multirow{3}{*}{Healthcare}
& Basic & $51.12$  & $64.25$ & $2.32$ & $16.59$ & $30.26$ & $12.91$ & $16.22$\\
& DRBA  & $\mathbf{94.57}$  & $88.07$ & $27.54$ & $44.82$ & $65.80$ & $46.34$ & $53.09$\\
& DRFA  & $93.28$  & $\mathbf{90.60}$ & $\mathbf{47.43}$ & $\mathbf{62.00}$ & $\mathbf{73.40}$ & $\mathbf{58.04}$ & $\mathbf{63.35}$ \\
\midrule
\multirow{3}{*}{Legal}
& Basic & $50.90$  & $53.33$ & $0.93$ & $6.03$ & $17.17$ & $7.15$ & $7.97$\\
& DRBA  & $\mathbf{85.02}$  & $92.89$ & $15.12$ & $34.13$ & $51.69$ & $34.49$ & $39.66$\\
& DRFA  & $84.96$  & $\mathbf{93.48}$ & $\mathbf{30.80}$ & $\mathbf{57.15}$ & $\mathbf{64.52}$ & $\mathbf{48.05}$ & $\mathbf{55.24}$\\
\bottomrule
\end{tabular}
}
\caption{Domain-wise results across all domains.}
\label{tab:domain_results}
\end{table*}

Overall, the domain-level analysis further supports the central claim of the paper: DRFA improves workflow quality in a systematic manner across diverse domains, with especially clear gains on metrics that require personalization, conditional reasoning, and coherent step organization.
\section{Mean and Standard Deviation Results}
\label{appendix:mean_std_results}
We report the mean and the standard deviation (across $3$ runs) of DRBA and DRFA on a subset of $15$ tasks in Table~\ref{tab:mean_std_results}.

\begin{table*}[t]
\centering
\small
\setlength{\tabcolsep}{5pt}
\resizebox{\textwidth}{!}{
\begin{tabular}{llcccc|ccc}
\toprule
\textbf{Backbone} & \textbf{Agent} & \textbf{Factuality}  & \textbf{Topology} & \textbf{Cond.~Res.} & \textbf{Pers.~Comp.} & \textbf{Recall} & \textbf{Precision} & \textbf{F1}\\
\midrule
GPT-5.2 & DRBA & $90.63{\scriptstyle \pm 1.93}$  & $89.80{\scriptstyle \pm 1.95}$ & $27.10{\scriptstyle \pm 4.68}$ & $39.20{\scriptstyle \pm 3.25}$ & $57.20{\scriptstyle \pm 6.78}$ & $52.23{\scriptstyle \pm 13.07}$ & $54.23{\scriptstyle \pm 10.27}$\\
        & DRFA& $89.37{\scriptstyle \pm 2.78}$  & $90.71{\scriptstyle \pm 2.04}$ & $43.89{\scriptstyle \pm 3.74}$ & $52.60{\scriptstyle \pm 6.26}$ & $66.08{\scriptstyle \pm 4.28}$ & $58.33{\scriptstyle \pm 7.96}$ & $61.87{\scriptstyle \pm 6.19}$\\
\midrule
Claude-Opus-4.5 & DRBA & $81.87{\scriptstyle \pm 4.44}$  & $91.80{\scriptstyle \pm 0.19}$ & $28.09{\scriptstyle \pm 6.03}$ & $38.34{\scriptstyle \pm 1.50}$ & $62.76{\scriptstyle \pm 2.91}$ & $47.31{\scriptstyle \pm 7.70}$ & $53.75{\scriptstyle \pm 5.80}$\\
                & DRFA & $83.51{\scriptstyle \pm 1.40}$  & $89.24{\scriptstyle \pm 0.07}$ & $50.00{\scriptstyle \pm 2.84}$ & $56.59{\scriptstyle \pm 4.80}$ & $69.57{\scriptstyle \pm 2.23}$ & $59.97{\scriptstyle \pm 9.58}$ & $64.11{\scriptstyle \pm 6.62}$\\
\midrule
Gemini-3.1-pro & DRBA & $87.53{\scriptstyle \pm 4.66}$  & $88.27{\scriptstyle \pm 0.58}$ & $32.40{\scriptstyle \pm 2.08}$ & $32.81{\scriptstyle \pm 8.81}$ & $63.56{\scriptstyle \pm 4.24}$ & $51.47{\scriptstyle \pm 8.00}$ & $56.77{\scriptstyle \pm 6.56}$\\
               & DRFA & $81.65{\scriptstyle \pm 6.81}$  & $87.76{\scriptstyle \pm 2.08}$ & $35.91{\scriptstyle \pm 6.48}$ & $45.41{\scriptstyle \pm 7.98}$ & $67.33{\scriptstyle \pm 2.12}$ & $56.10{\scriptstyle \pm 8.59}$ & $60.98{\scriptstyle \pm 6.04}$\\
\midrule
DeepSeek-v3.2 & DRBA & $86.44{\scriptstyle \pm 2.30}$  & $88.41{\scriptstyle \pm 0.86}$ & $31.13{\scriptstyle \pm 2.42}$ & $27.90{\scriptstyle \pm 5.43}$ & $61.94{\scriptstyle \pm 3.08}$ & $52.46{\scriptstyle \pm 2.58}$ & $56.80{\scriptstyle \pm 2.77}$\\
              & DRFA & $84.11{\scriptstyle \pm 8.34}$  & $78.87{\scriptstyle \pm 5.86}$ & $38.56{\scriptstyle \pm 0.57}$ & $51.65{\scriptstyle \pm 4.83}$ & $64.42{\scriptstyle \pm 0.71}$ & $56.04{\scriptstyle \pm 3.29}$ & $59.91{\scriptstyle \pm 2.18}$\\
              
\bottomrule
\end{tabular}
}
\caption{Mean and standard deviation results on the subset across backbone models. Higher is better for all metrics.}
\label{tab:mean_std_results}
\end{table*}

\section{Human Evaluation Details}
\label{appendix:human_evaluation}

\begin{figure*}[t]
    \centering
    \includegraphics[width=0.7\textwidth]{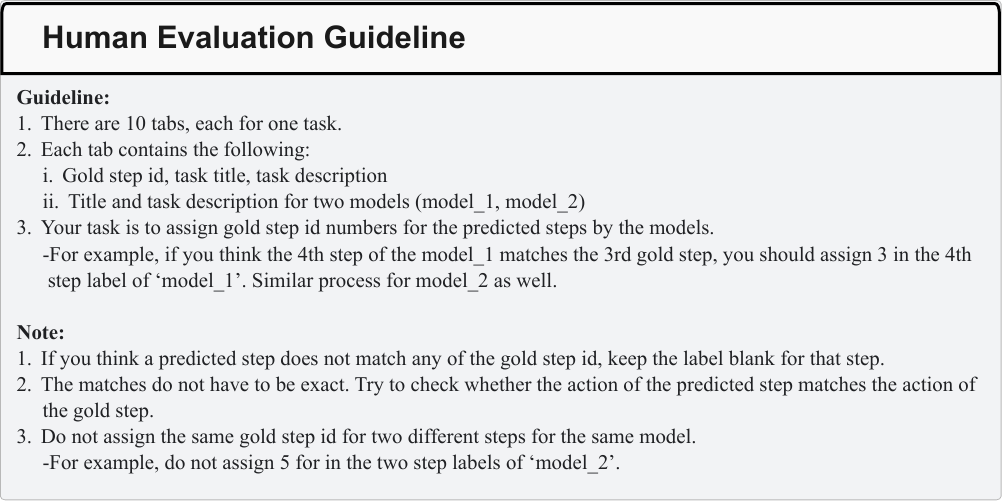}
    \caption{Human evaluation guideline.}
    \label{fig:human-eval-guideline}
\end{figure*}

\subsection{Step-Matching Annotation Protocol}
\label{appendix:human-evaluation-protocol}

We conduct a human study on $10$ benchmark tasks across the outputs of \drflow~and DRBA, producing $250$ predicted-step decisions in total. First, we provide a guideline to the annotators detailing the annotation protocol (Figure~\ref{fig:human-eval-guideline}). For each task, annotators are shown the gold workflow steps and the predicted workflow steps produced by each agent. Three annotators independently align predicted steps to gold steps using a one-to-one semantic matching protocol. A predicted step is considered matched only if it expresses the same underlying procedural action as a gold step. The annotation does not require lexical overlap: differences in phrasing are ignored when the predicted step satisfies the same workflow obligation. If a predicted step does not correspond to any gold step, it is left unmatched. Annotators are also instructed not to assign the same gold step to multiple predicted steps within the same workflow.

\paragraph{Results.}
All three annotators agree on $143$ of $250$ decisions ($57.2\%$), while at least two of three annotators agree on $244$ of $250$ decisions ($97.6\%$). The inter-annotator agreement measured by Fleiss' $\kappa$ is $0.663$, implying substantial agreement range under the Landis and Koch interpretation~\cite{landis1977measurement}.

\begin{figure*}[t]
    \centering
    \includegraphics[width=\textwidth]{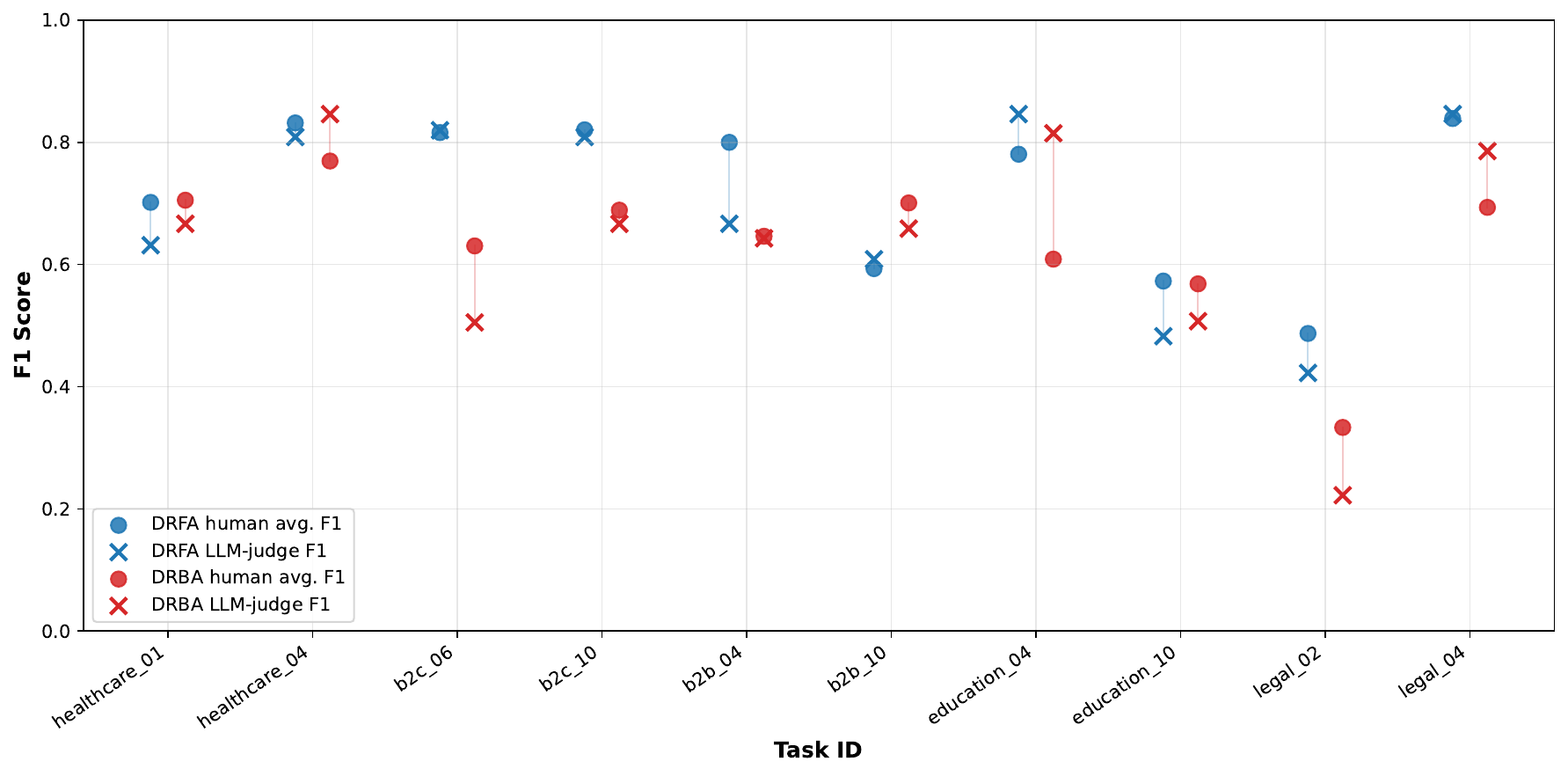}
    \caption{Task-wise agreement between human and LLM-judge F1. The x-axis shows the evaluated tasks, and the y-axis shows F1. For each task, we plot the human-average F1 and LLM-judge F1 for \drflow~and DRBA. Smaller vertical gaps between the human and LLM markers indicate closer agreement.}
    \label{fig:human-llm-f1}
\end{figure*}

\subsection{Agreement with LLM-as-a-Judge}
\label{app:human-llm-agreement}

We compare the LLM-as-a-judge evaluator against the aggregated human annotations on the same $10$-task subset. For each task and agent, we compute the average human F1 score across the three annotators and compare it with the corresponding LLM-judge F1 score. The Pearson correlation between human-average F1 and LLM-judge F1 is $0.883$, and the mean absolute F1 difference is $0.063$.

Figure~\ref{fig:human-llm-f1} reports task-wise F1 scores for human evaluation and LLM-judge evaluation. The x-axis denotes the task, and the y-axis denotes F1. For both DRFA and DRBA, the human and LLM-judge scores remain close across most tasks, indicating that the LLM evaluator preserves the same task-level performance trends as expert annotation. For our task of personalized workflow prediction, this finding on the positive correlation between human and LLM judge contradicts some prior works~\cite{shen-etal-2023-large,chen-etal-2024-humans}, but corroborates other works~\cite{chiang-lee-2023-large,zheng2023judging} that support the similar claim.

\section{Qualitative Analysis}
\label{appendix:extended-qualitative-analysis}

\subsection{Condition-aware personalization.}
\label{appendix:condition_aware_personalization}

\begin{table}[h!]
\centering
\small
\begin{tabular}{p{1.9cm}p{3.2cm}}
\toprule
\textbf{Agent} & \textbf{Example Steps} \\
\midrule
\textbf{DRBA} & ``Plan and request a qualified interpreter (choose tier/mode)''; ``Connect interpreter before any sedating premedication''; ``Use the standardized consent content checklist/script'' \\
\textbf{DRFA} & ``Confirm interpreter plan: qualified VRI for consents''; ``Run the consent conversation through the interpreter (do not use family)''; ``Perform the pre-signature sedation/capacity gate check and escalate if needed'' \\
\bottomrule
\end{tabular}
\caption{Qualitative comparison on \texttt{healthcare\_01}. DRFA improves patient-specific branch instantiation while preserving the same overall clinical scaffold.}
\label{table:qual-healthcare-main}
\end{table}

With GPT-$5.2$ fixed we conduct qualitative analysis on the condition-aware personalization quality of agents. The comparison between the DRBA and DRFA on \texttt{healthcare\_01} task shows that DRFA improves on both Condition Resolution ($58.33$ to $75.00$) and Personalized Comprehensiveness ($55.17$ to $62.00$) compared to DRBA. 
We manually inspect the predicted steps for both agents to understand the possible reason. As Table~\ref{table:qual-healthcare-main} shows,  DRBA agent predicts the pre-operative scaffold, but several branch-sensitive decisions remain distributed across adjacent operational steps. By contrast, DRFA packages the same evidence into more decision-oriented obligations, e.g., \textit{directly committing to the VRI path}, \textit{binding the no-family rule to the actual consent encounter}, and \textit{coupling the sedation-capacity check to signature execution}. The gain in metric is therefore comes not from broader retrieval, but condition-aware workflow personalization.

\subsection{Factuality versus workflow coverage}
\label{app:qual-b2c08-factuality}


We find that sometimes a non-factual predicted step can also carry a matched reference workflow step. For example, on \texttt{b2c\_08} task, the Claude-Opus-$4.5$ workflow step (Table~\ref{tab:qual-b2c08-appendix}) ``Deletion Will Be Sent to MindMint's Service Providers and Vendors'' reaches the correct part of the workflow, namely processor-side deletion, but hallucinates what the evidence warrants. This makes the step  problematic from a factuality perspective. However, simply removing it does not cleanly solve the workflow. Because the step is also one of the matched obligations, pruning it would reduce the number of matched steps, weakening procedural coverage even as factuality improves.

\begin{table}[h]
\centering
\small
\begin{tabular}{p{1cm}p{3cm}c}
\toprule
\textbf{Workflow} & \textbf{Example step} & \textbf{\# matched steps} \\
\midrule
\textbf{Original} & \textcolor{red}{``Deletion Will Be Sent to MindMint's Service Providers and Vendors''}
\textit{Justification:} The sources do not explicitly mention that MindMint will send formal deletion requests to each vendor or track vendor confirmations. While the sources discuss the internal processes for handling deletion requests and maintaining compliance, they do not provide specific details about coordination with third-party vendors for data deletion, making the claim unsupported. & $15$\\
\midrule
\textbf{Pruned} & After removing ``Deletion Will Be Sent to MindMint's Service Providers and Vendors'' & $14$\dn{1} \\
\bottomrule
\end{tabular}
\caption{Pruning an \textcolor{red}{non-factual step} on \texttt{b2c\_08} improves factuality, but also removes a matched processor-deletion obligation. This illustrates that unsupported steps are not always disposable over-generation.}
\label{tab:qual-b2c08-appendix}
\end{table}

This example showcases why workflow cannot be optimized by simply guardrailing against non-factual steps alone. If an unsupported step is unmatched, pruning is mostly beneficial. If it is unsupported but also matched, pruning improves local reliability at the expense of workflow completeness.

\subsection{Cost–Performance Analysis}

\begin{figure*}[t]
    \centering
    \begin{subfigure}[t]{0.49\textwidth}
        \centering
        \includegraphics[width=\linewidth]{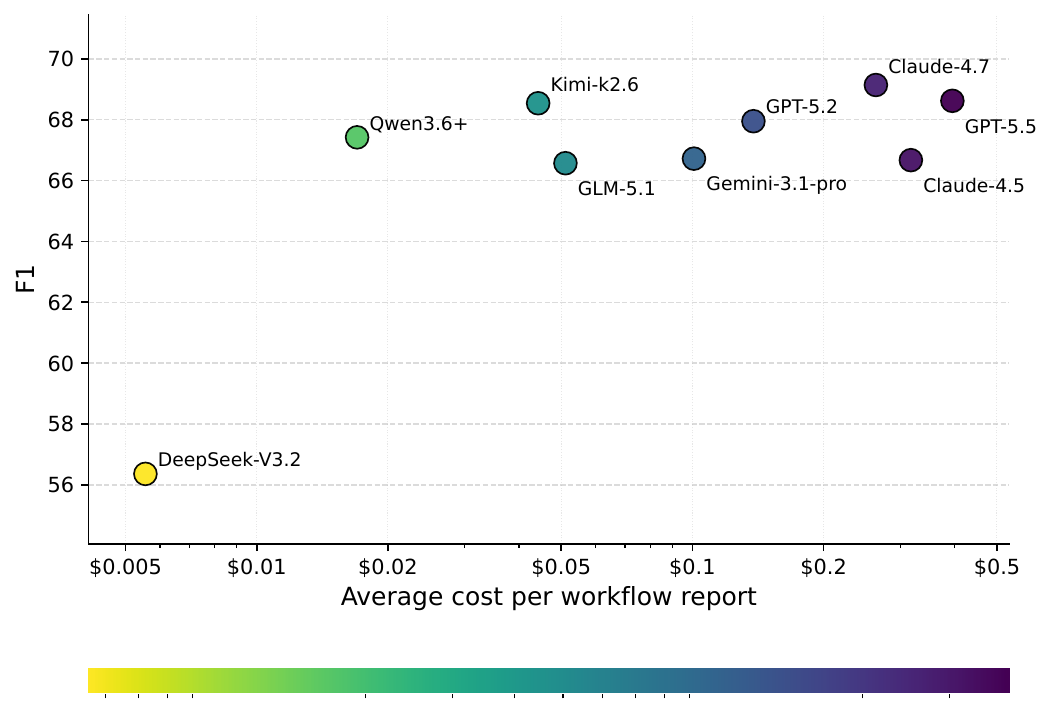}
        \caption{Performance comparison on F1.}
        \label{fig:pareto_f1}
    \end{subfigure}
    \hfill
    \begin{subfigure}[t]{0.49\textwidth}
        \centering
        \includegraphics[width=\linewidth]{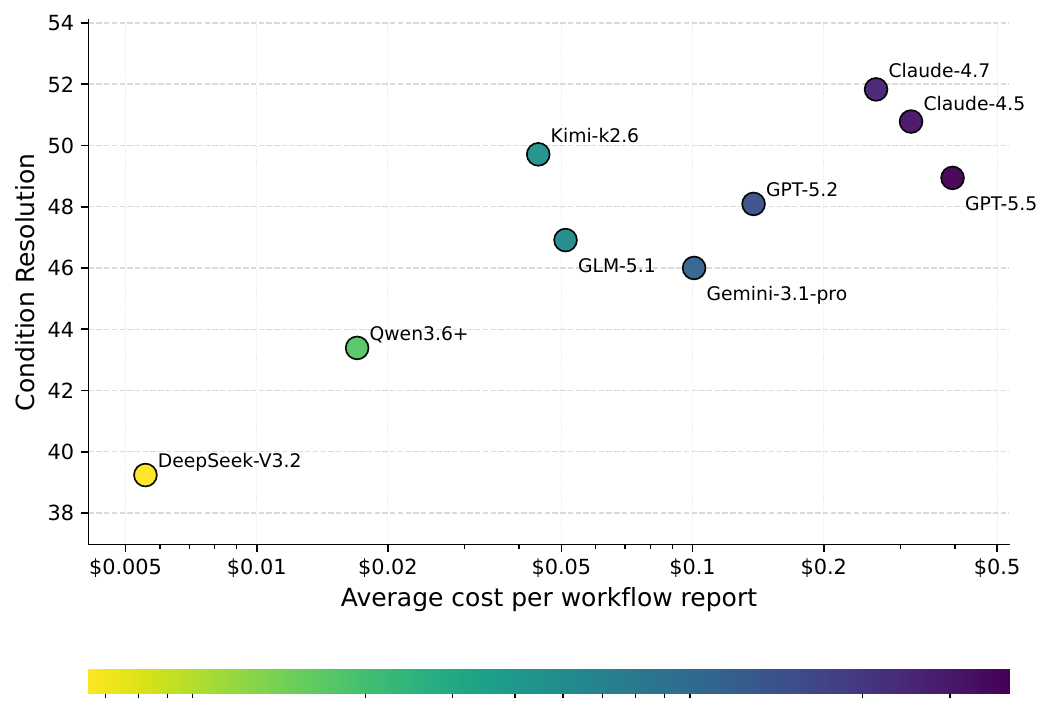}
        \caption{Performance comparison on Condition Resolution.}
        \label{fig:pareto_cond_res}
    \end{subfigure}
    \caption{Cost-performance analysis of the models on the ablation subset. The x-axis shows cost per report, with prices (USD/$1$M tokens) collected from OpenRouter~\cite{openrouter2026}, and the y-axis shows performance on F1 or Condition Resolution.}
    \label{fig:performance_cost}
\end{figure*}

We further analyze the cost-performance trade-off of DRFA agents using backbone prices collected from OpenRouter~\cite{openrouter2026}. We use GPT-OSS-120b~\cite{openaigptoss} tokenizer to tokenize the reports then calculate the average report cost. Figure~\ref{fig:performance_cost} shows the comparison among the frontier models on F1 and Condition Resolution. Both comparisons exhibit the same Pareto pattern: DeepSeek-V3.2 anchors the lowest-cost regime, Qwen3.6-plus and Kimi-k2.6 provide strong mid-cost trade-offs, and Claude-Opus-4.7 achieves the best absolute performance. On F1, Kimi-k2.6 reaches $68.54$ at a cost of $0.044$, closely approaching Claude-Opus-4.7 at $69.14$ despite being substantially cheaper. On Condition Resolution, Kimi-k2.6 again offers the strongest mid-cost operating point at $49.71$, while Claude-Opus-4.7 remains the top performer with the cost of $0.395$. Overall, the Pareto frontier indicates that DRFA can construct most of its workflow quality with mid-cost backbones, while the most expensive model yields additional gains.

\subsection{Closed-source versus open-source behavior}
\label{app:qual-healthcare-open-closed}

We manually inspect the predicited workflow of closed-source (GPT-$5.2$) vs open-source (DeepSeek-V$3.2$) models on \texttt{healthcare\_01} task (Table~\ref{tab:qual-healthcare-appendix}).

\begin{table}[h]
\centering
\small
\begin{tabular}{p{1cm}ccp{2cm}}
\toprule
\textbf{Model} & \textbf{Precision} & \textbf{Cond. Res.} & \textbf{Example steps} \\
\midrule
GPT-$5.2$ & $50.00$ & $75.00$ & ``Confirm interpreter plan: qualified VRI for consents''; ``Run the consent conversation through the interpreter (do not use family)''; ``Perform the pre-signature sedation/capacity gate check and escalate if needed'' \\
\midrule
DeepSeek-V$3.2$ & $33.33$ & $41.67$ & ``Schedule the Contracted VRI Service for the Procedure Day''; ``Conduct Consent Discussion Entirely Through VRI Interpreter''; ``Re-assess Capacity if Sedating Medication Administered'' \\
\bottomrule
\end{tabular}
\caption{Closed-source versus open-source comparison on \texttt{healthcare\_01}. Both models predict the broad clinical scaffold, but GPT-$5.2$ is more effective at binding evidence to branch-specific clinical obligations.}
\label{tab:qual-healthcare-appendix}
\end{table}

We find that GPT-$5.2$ describes the relevant evidence into steps that are closer to the atomic decision structure of the reference workflow. For example, it ties \textit{the VRI choice}, \textit{the no-family constraint}, and \textit{the sedation-capacity gate} directly to the actual consent path. On the other hand, DeepSeek-V$3.2$ often distributes the same evidence across a more generic operational sequence. The resulting workflow remains plausible and factual, but is less decisive at the point where the personalized branch should actually be chosen. This explains why the models diverge substantially on precision and condition resolution.

\section{Error Analyses}
\label{appendix:error_analyses}

\subsection{Manual Analysis of DRFA Factuality}
\label{appendix:drfa_factuality}

DRFA, unlike DRBA, combines adaptive gap finding with conditional action planning. While this design improves coverage and condition resolution, the interaction between these modules may also produce overlapping strategies, which can create redundant or unstable planning behavior, as observed in~\cite{abaskohi2025drbench}, hurting the factuality of the predicted workflows.

Furthermore, we manually analyze the workflows generated by the agents, to better understand why DRFA obtains a lower factuality score than DRBA. We find that DRFA's comparatively lower factuality score can potentially be due to the specificity of the claim cited by the agent. A side-by-side comparison between DRFA vs DRBA on the opening step of the personalized workflow on \texttt{healthcare\_01} (Claude-Opus-4.5) in Table~\ref{tab:fact_qualitative} illustrates this pattern. Both steps target the same obligation (\emph{confirm the patient's preferred language during pre-op intake}), and both cite valid policy sources. DRBA's task description stays close to the language of the cited policy: a generic obligation that the judge can verify directly against the source. DRFA's task description, in contrast, binds the obligation to case-specific operational detail (\emph{``navigate to the designated Language \& Communication area,''} \emph{``requested interpreter support for all medical decisions''}) that fuses the generic policy with personal evidence. The judge, which retrieves the top relevant chunks from cited sources before evaluating each claim (Appendix~\ref{appendix:implementation_details}), still has a reasonable chance to verify such fused claims. But since the verification process is inherently stricter: the cited chunks must semantically match the predicted claim, fusing the policy with personalized evidence can reduce the semantic alignment, resulting in lower factuality score. Adding further citations partially compensates, since each additional source raises the chance that the predicted claim is locally supported, which is why the observed gap is small ($\sim$2 points). The residual gap is therefore not a deficiency of grounding but a measurable cost of DRFA's design choice to produce more personalized claims; a cost that is repaid by its gains on Recall, F1, Condition Resolution, and Personalized Comprehensiveness.  Overall, these findings suggest that DRFA's lower factuality score is partly a consequence of its more aggressive evidence-seeking and planning behavior, which improves coverage but increases the likelihood of citation misalignment.

\begin{table*}[t]
\centering
\small
\begin{tabular}{p{0.08\linewidth}p{0.42\linewidth}p{0.42\linewidth}}
\toprule
\textbf{Agent} & \textbf{Step 1 Task Description} (\texttt{healthcare\_01}, Claude-Opus-4.5) & \textbf{Verification surface} \\
\midrule
\textbf{DRBA}  
& ``During pre-operative intake, ask the patient directly, not accompanying family members, what their preferred language is for medical communication. Document this language preference in the EHR immediately.''  
& Generic, policy-aligned obligation; phrasing closely tracks the cited policy text, so a single chunk retrieved from the cited source supports the claim. \\
\addlinespace
\textbf{DRFA}  
& ``During pre-op intake at Sunrise Surgical Center, access the patient's EHR chart and navigate to the \emph{designated Language \& Communication area}. For this case, pre-op intake has identified Spanish as the patient's preferred language and the patient has \emph{requested interpreter support for all medical decisions}.'' 
& Synthesized obligation that fuses generic policy with personal-evidence facts. The judge must verify the operational detail (the named EHR area, the patient-specific request) against retrieved chunks; each citation supports a sub-claim, and the step is factual only if those sub-claims jointly hold. \\
\bottomrule
\end{tabular}
\caption{Paired DRBA and DRFA steps on the same task, illustrating the claim-specificity difference behind the small Factuality gap.}
\label{tab:fact_qualitative}
\end{table*}

\subsection{Role of Distractors in Low Precision}
We analyze whether low precision in Table~\ref{tab:main_results} is associated with models incorporating distracting information into the generated workflow reports. For each model, we compute the distractor-reference rate in the DRFA report as the percentage of references that point to distractor information. We then compare this rate against the corresponding model-level precision score.

\begin{figure}[t]
    \centering
    \includegraphics[width=0.92\linewidth]{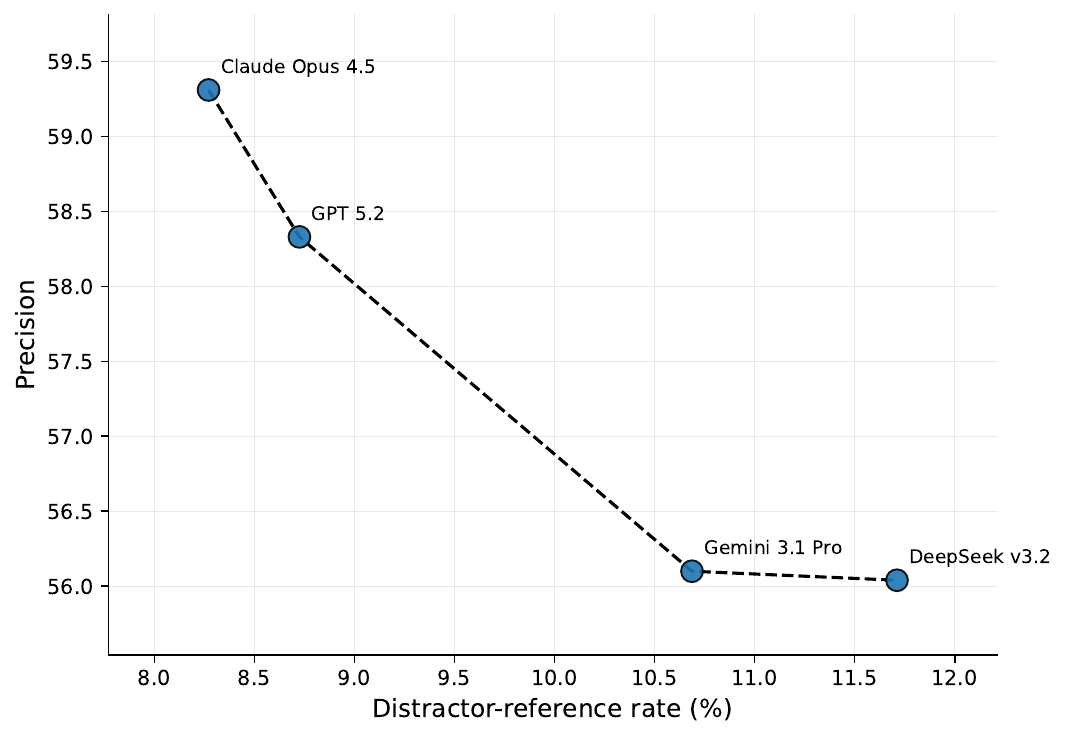}
    \caption{
    Relationship between distractor-reference rate and precision. The downward trend indicates that models with higher proportions of distractor references tend to achieve lower precision.
    }
    \label{fig:drfa-distractor-precision}
\end{figure}

Figure~\ref{fig:drfa-distractor-precision} shows a negative association between distractor-reference rate and precision. Models that include a larger fraction of distractor references tend to obtain lower precision, suggesting that precision errors are not only caused by missing relevant evidence, but also by selecting and propagating spurious evidence into the workflow.

\subsection{Impact of the Mixed Split}
\label{sec:adversarial_error_analysis}

We inspect a random subset of $20$ tasks to characterize how the mixed split degrades performance in Table~\ref{tab:adv_results}. For each predicted workflow, we count citations to supporting ($\texttt{*\_IN\_*}$), distractor ($\texttt{*\_DI\_*}$), and mixed ($\texttt{*\_AD\_*}$) sources, and we manually inspect predicted personalized workflows to determine whether contamination propagates into the steps.

\paragraph{Citation composition.} Table~\ref{tab:adv_citation_composition} reports the share of citations by source type. Both agents refer a substantial amount of citations to mixed sources, while classical distractor citations remain low ($2$ - $8$\%). This confirms that the mixed artifacts, which are internally consistent and topically aligned with the target question, attract retrieval far more than off-topic distractors, degrading the overall performance of the agents.

\begin{table}[t]
\centering
\small
\begin{tabular}{lccc}
\toprule
Agent & $\texttt{*\_IN\_*}$ (\%) & $\texttt{*\_DI\_*}$ (\%) & $\texttt{*\_AD\_*}$ (\%) \\
\midrule
DRFA & 70.7 & 2.2 & 27.2 \\
DRBA & 65.0 & 8.2 & 26.8 \\
\bottomrule
\end{tabular}
\caption{Citation composition on the mixed split.}
\label{tab:adv_citation_composition}
\end{table}

\paragraph{Manual inspection.} Table~\ref{tab:adv_qualitative} shows two representative cases in which the agent grounds the personalized workflow on the mixed case rather than the user's case. In \texttt{b2c\_02}, the predicted workflow opens with the proration case identifiers (\texttt{Marcus Whitfield}, \texttt{SG-48219}) sourced from \texttt{dispute-review-sg48219.docx}, even though the original user's task concerns a duplicate charge. In \texttt{education\_10}, $5$ of $15$ predicted steps cite mixed artifacts (\texttt{chair-provost-approval.docx}, \texttt{aaf-split-checklist.docx}) describing a conversion with prior-service-credit case, while the user's task concerns onboarding a newly hired tenure-track faculty.

\begin{table*}[t]
\centering
\small
\begin{tabular}{p{0.08\linewidth}p{0.22\linewidth}p{0.22\linewidth}p{0.36\linewidth}}
\toprule
\textbf{Task} & \textbf{Original DR Question} & \textbf{Mixed DR Question} & \textbf{Contaminated Step (excerpt)} \\
\midrule
\texttt{b2c\_02} & What is the workflow to dispute a duplicate charge on a customer's subscription? & What is the workflow to handle a mid-cycle upgrade proration dispute? & ``Log Marcus Whitfield's SG-48219 \emph{Proration} Dispute \dots\ attach screenshot 11/08/2025 20:14 UTC \dots\ target resolution Nov 21, 2025 [\^{}6]'' : \textcolor{red}{\textit{[\^{}6] resolves to \texttt{dispute-review-sg48219.docx}}} \\
\addlinespace
\texttt{edu\_10} & What is the workflow to onboard a newly hired tenure-track faculty? & What is the workflow to convert a non-tenure-track faculty with prior service credit? & ``Submit Department Chair Endorsement and Dossier \dots\ \emph{prior service credit recommendations} [\^{}5]'' : \textcolor{red}{\textit{[\^{}5] resolves to \texttt{chair-provost-approval.docx}}} \\
\bottomrule
\end{tabular}
\caption{Qualitative examples of mixed contamination. The predicted workflow binds steps to the mixed case.}
\label{tab:adv_qualitative}
\end{table*}

\paragraph{Reconciliation with Table~\ref{tab:adv_results}.} When comparing against the performance on the original variant (Table~\ref{tab:main_results}), two patterns explain the metric movements observed. (i) \emph{Precision drops more than Recall} ($-14.37$ vs.\ $-10.60$ averaged for DRFA; $-10.77$ vs.\ $-5.98$ for DRBA) because the agents recover the generic procedural scaffold but inject step-level details that inlcude the mixed case. (ii) \emph{Factuality remain over $80$\%}, because mixed artifacts are internally consistent documents that locally support the (wrong) claims they are cited for. This shows that an mixed-grounded workflow can be locally factual against its citations, yet globally wrong with respect to the user. This failure mode is invisible to citation-only evaluation but is captured by Recall, Precision, F1, Condition Resolution, and Personalized Comprehensiveness, validating the effectiveness of the mixed split to stress test the agents.

\begin{figure*}[t]
    \centering
    \begin{subfigure}[t]{0.45\textwidth}
        \centering
        \includegraphics[width=\linewidth]{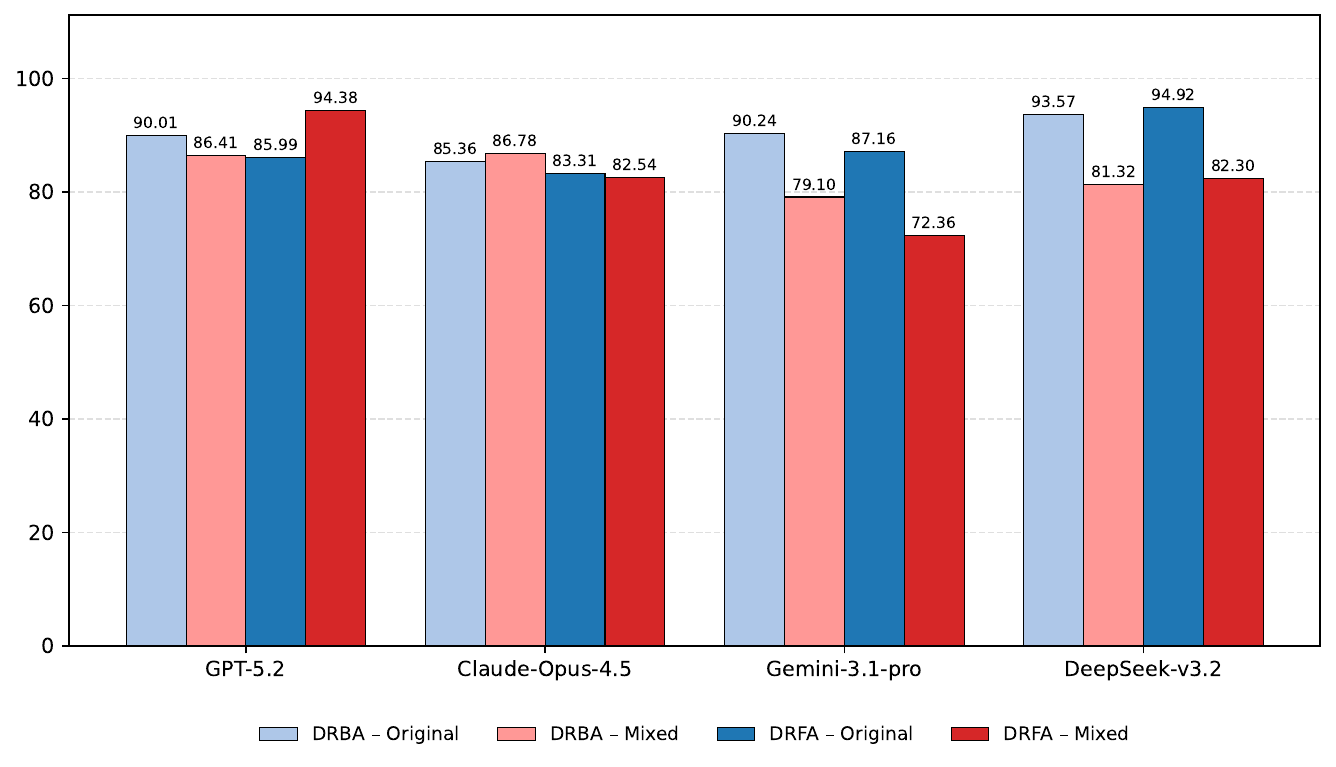}
        \caption{Factuality}
        \label{fig:cmp_factuality}
    \end{subfigure}
    \hfill
    \begin{subfigure}[t]{0.45\textwidth}
        \centering
        \includegraphics[width=\linewidth]{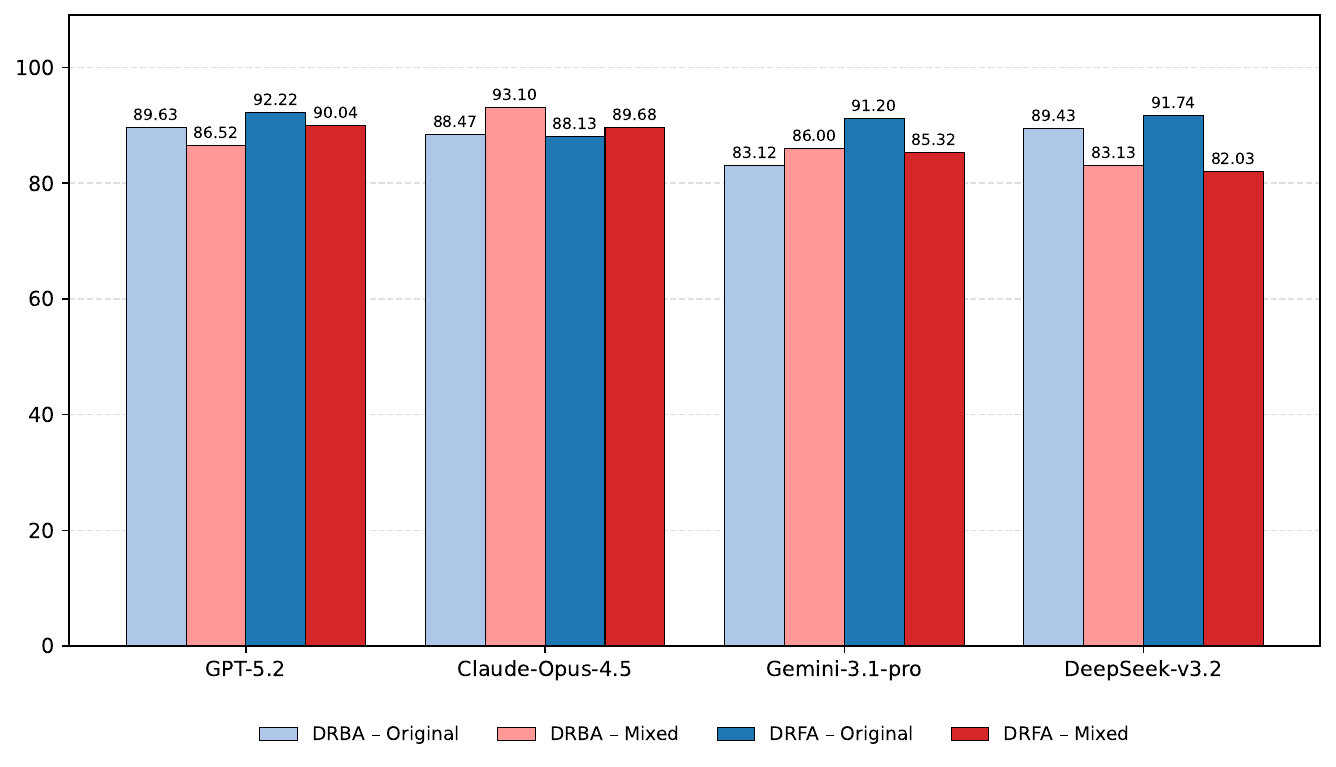}
        \caption{Topology}
        \label{fig:cmp_topology}
    \end{subfigure}
    \hfill
    \begin{subfigure}[t]{0.45\textwidth}
        \centering
        \includegraphics[width=\linewidth]{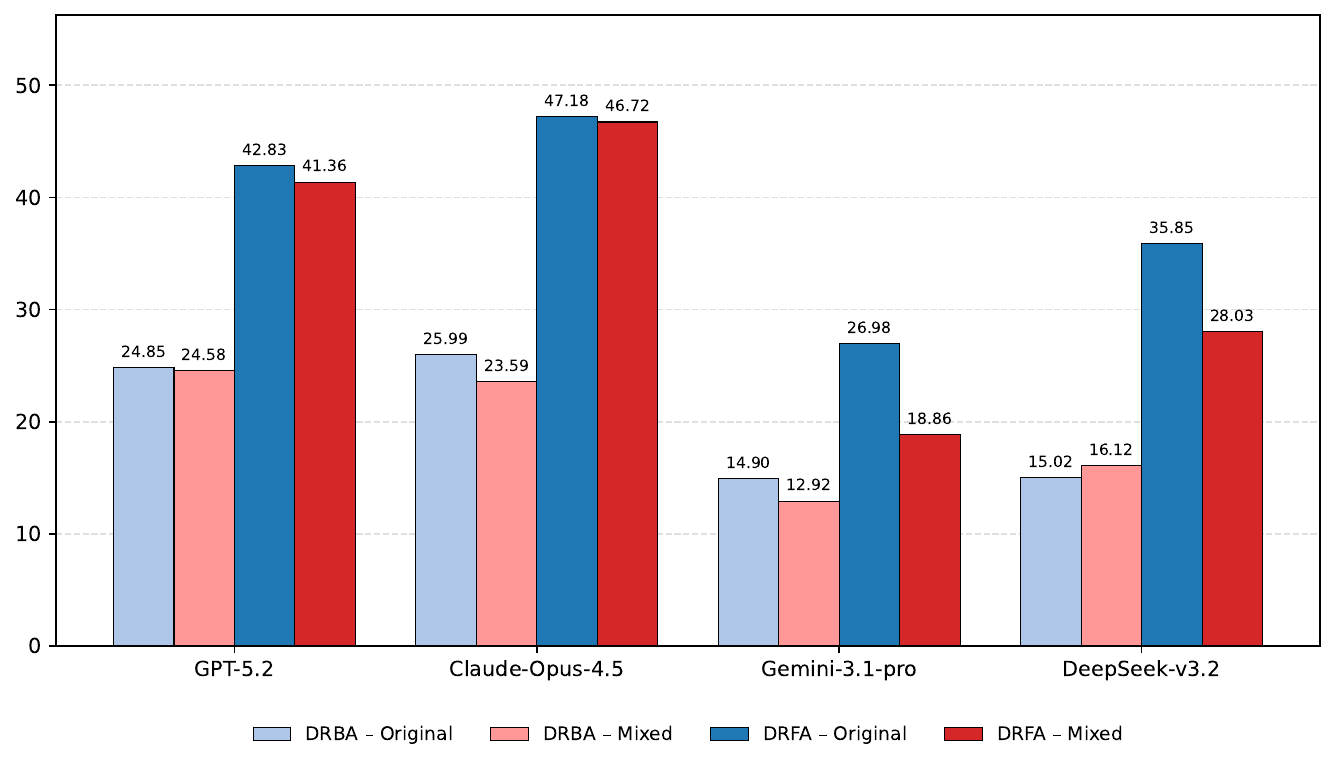}
        \caption{Cond.~Res.}
        \label{fig:cmp_condres}
    \end{subfigure}
    \hfill
    \begin{subfigure}[t]{0.45\textwidth}
        \centering
        \includegraphics[width=\linewidth]{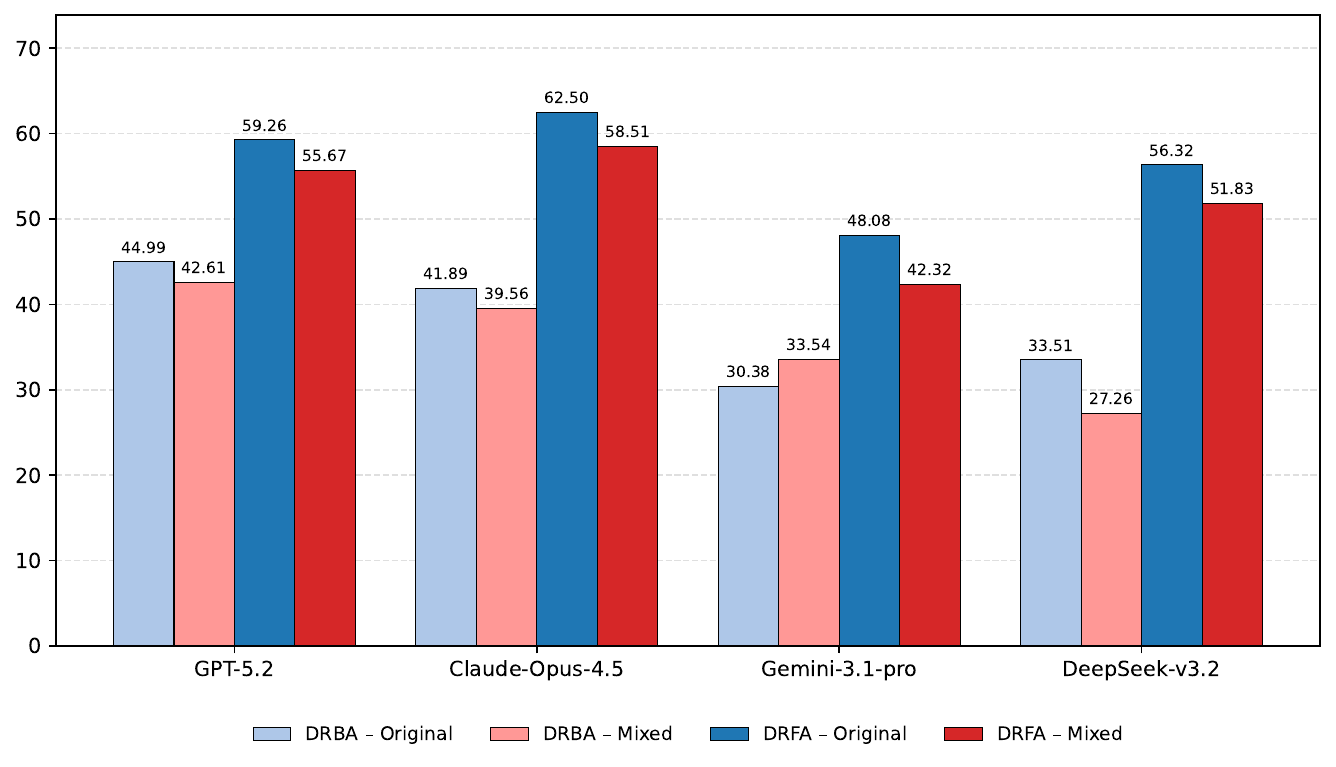}
        \caption{Pers.~Comp.}
        \label{fig:cmp_perscomp}
    \end{subfigure}

    \hfill

    \begin{subfigure}[t]{0.45\textwidth}
        \centering
        \includegraphics[width=\linewidth]{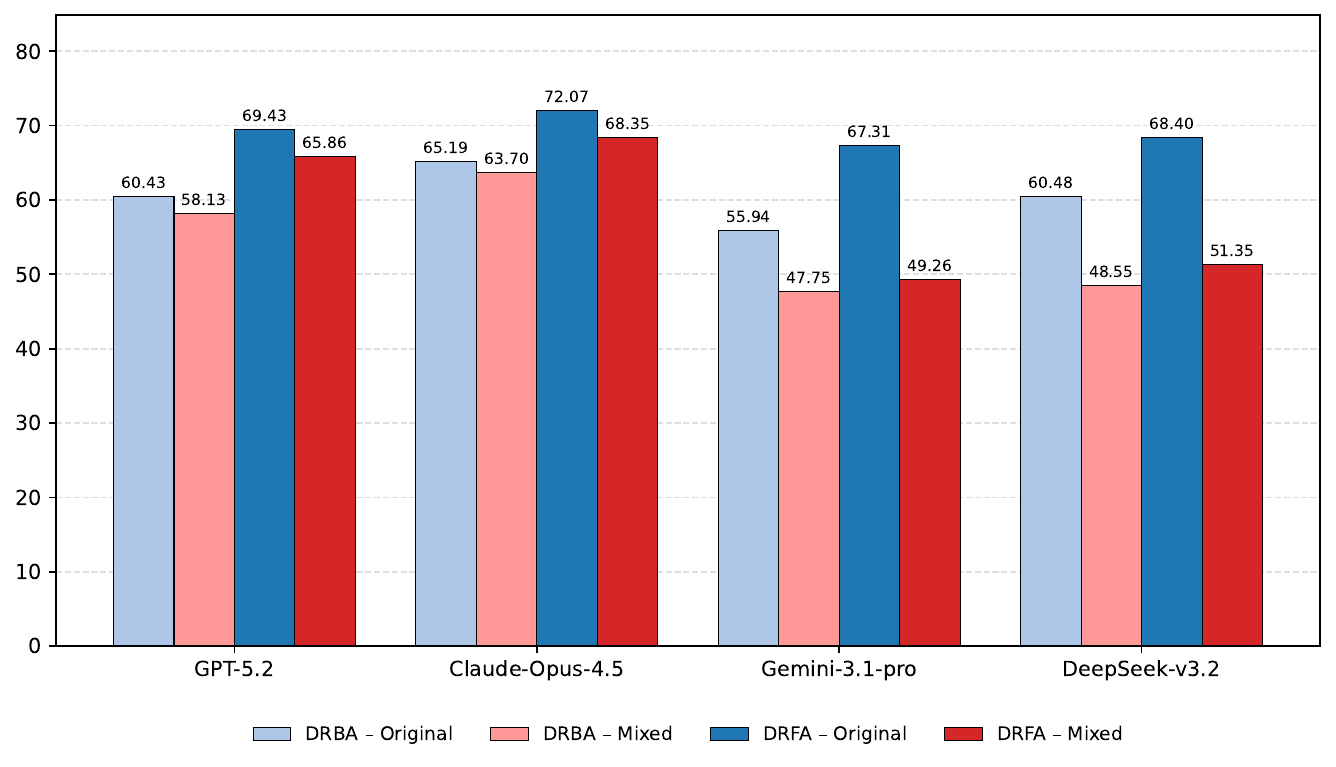}
        \caption{Recall}
        \label{fig:cmp_recall}
    \end{subfigure}
    \hfill
    \begin{subfigure}[t]{0.45\textwidth}
        \centering
        \includegraphics[width=\linewidth]{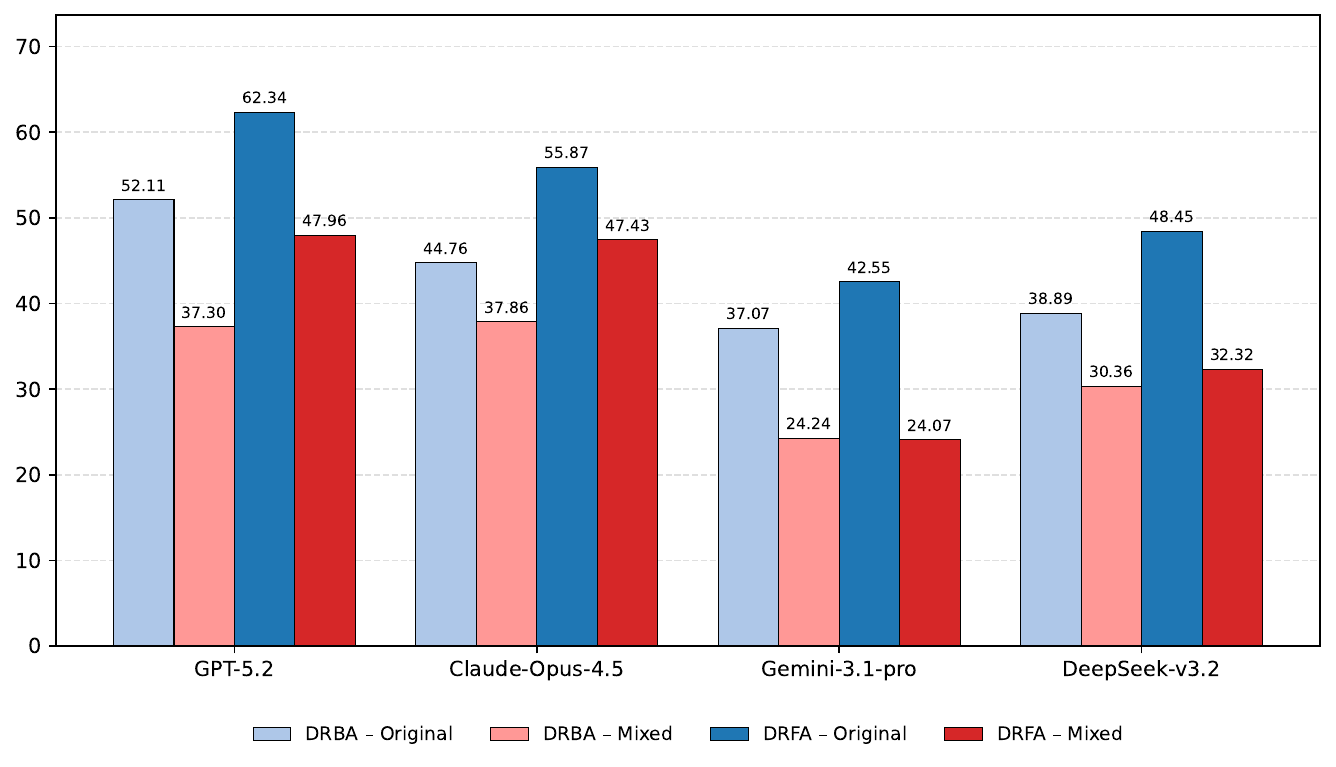}
        \caption{Precision}
        \label{fig:cmp_precision}
    \end{subfigure}
    \hfill
    \begin{subfigure}[t]{0.45\textwidth}
        \centering
        \includegraphics[width=\linewidth]{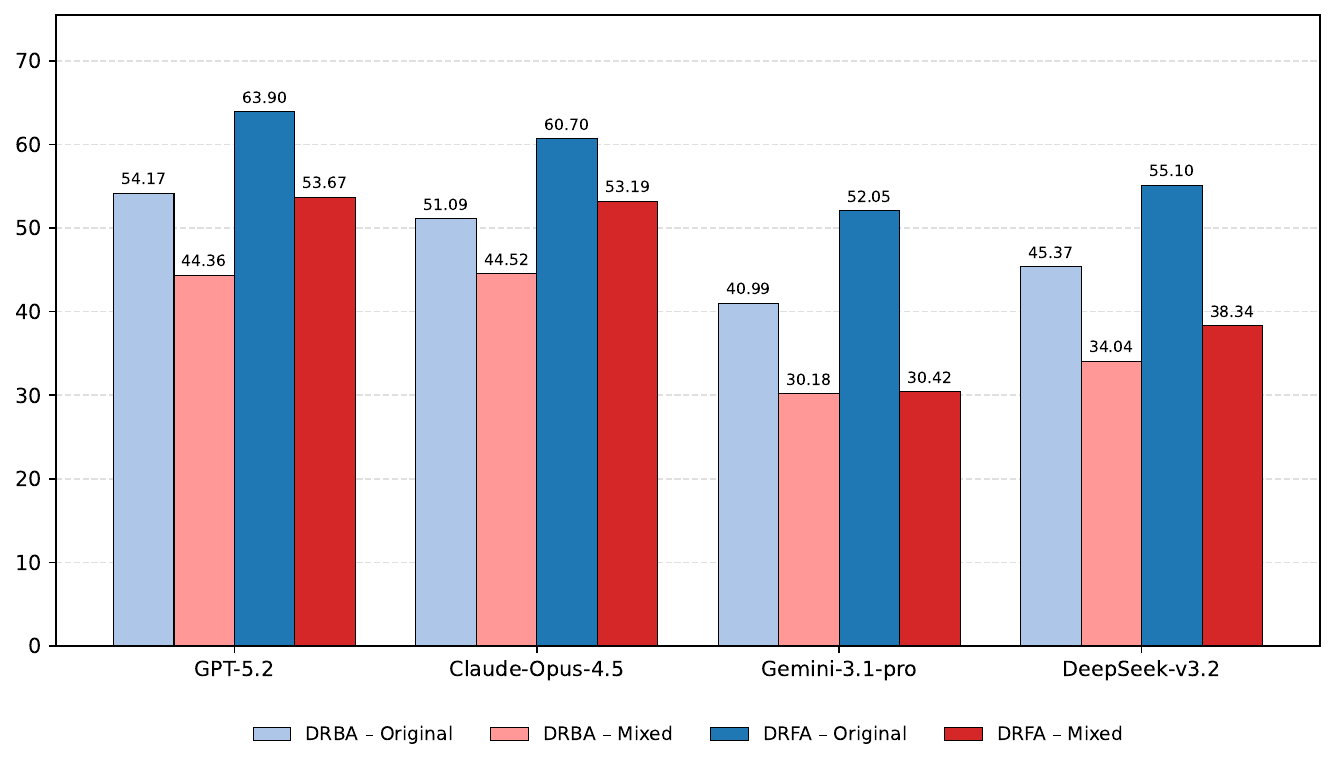}
        \caption{F1}
        \label{fig:cmp_f1}
    \end{subfigure}

    \caption{Per-metric comparison of DRBA and DRFA on the original split (blue) versus the mixed split (red) across four backbone models. Lighter shades denote DRBA; darker shades denote DRFA.}
    \label{fig:original_vs_mixed}
\end{figure*}

\section{Performance Comparison Between Original \& Mixed Variants}
\label{appendix:performance_original_mixed}

We show the performance comparison between the original and the corresponding mixed variants across all the models in Figure~\ref{fig:original_vs_mixed}.

\section{DRFA Agent Details}
\label{appendix:drfa_details}

In this section, we provide implementation details for \textsc{DRFA}. The design goal of \textsc{DRFA} is to predict a generic workflow from company-side evidence, identify the decision points that remain unresolved at the generic level, and then instantiate a personalized workflow using user-specific evidence.

\subsection{Detailed Workflow}
\label{app:drfa_detailed_workflow}

\textsc{DRFA} operates as an iterative tool-calling agent over heterogeneous private information sources. At the beginning of a session, the agent initializes a task-specific vector store, registers its tool inventory, and constructs a research context tied to the original deep research question. The execution pipeline then proceeds through the following stages.

\paragraph{Stage $1$: Research Planning.}
The agent decomposes the original deep research question into a small set of research investigation areas. These areas are designed to cover two complementary evidence objectives: constructing the generic organizational procedure and collecting user-specific evidence for personalization.

\paragraph{Stage $2$: Action Planning.}
The research plan is translated into executable actions. Each action is parameterized by an action type, a natural-language description, tool arguments, expected outputs, dependencies, preferred tool choices, and a workflow stage label. The planner prioritizes internal sources over external ones, favoring local documents and enterprise applications, while reserving web search as a fallback when internal evidence is insufficient. The resulting action plan forms the initial retrieval program for the session.

\paragraph{Stage $3$: Iterative Research Loop.}
The agent executes the action plan under a fixed iteration budget. At each iteration, it selects the next executable actions, invokes the appropriate tools, processes the returned content, and stores the resulting findings in the task context and vector store. This loop continues until the plan converges or the iteration budget is exhausted.

\paragraph{Stage $4$: Adaptive Action Planning.}
At the end of each action-planning round, \textsc{DRFA} analyzes the newly collected findings and decides whether the plan should be expanded. This adaptive step first prioritizes conditional action planning, which extracts an intermediate generic workflow and proposes focused actions for resolving its conditional branches, and then performs complementary gap finding when evidence remains unbalanced across the generic and personal sides.

\paragraph{Stage $5$: Workflow Generation.}
After the research loop converges, the agent predicts the final workflow output. This stage reconstructs the full generic workflow, resolves it using user-specific evidence, and produces a cited personalized workflow in which each step is justified by retrieved facts.

\subsection{Research Planning Implementation}
\label{app:drfa_research_planning}

The research planning module organizes the problem around the structure of the target output rather than around generic topical decomposition alone. In particular, it separates the search problem into two evidence objectives.

The first objective is \textbf{Generic Requirement Collection}, which targets company-side materials such as policies, procedures, playbooks, templates, and operational documents. These materials define the organization-level procedure that the agent must predict as the generic workflow.

The second objective is \textbf{Personal Evidence Collection}, which targets user-side materials such as emails, chats, notes, records, and other personal artifacts. These materials determine which conditions in the generic workflow hold for the current case and therefore how the workflow should be instantiated for the user.

\begin{figure*}[t]
    \centering
    \includegraphics[width=\textwidth]{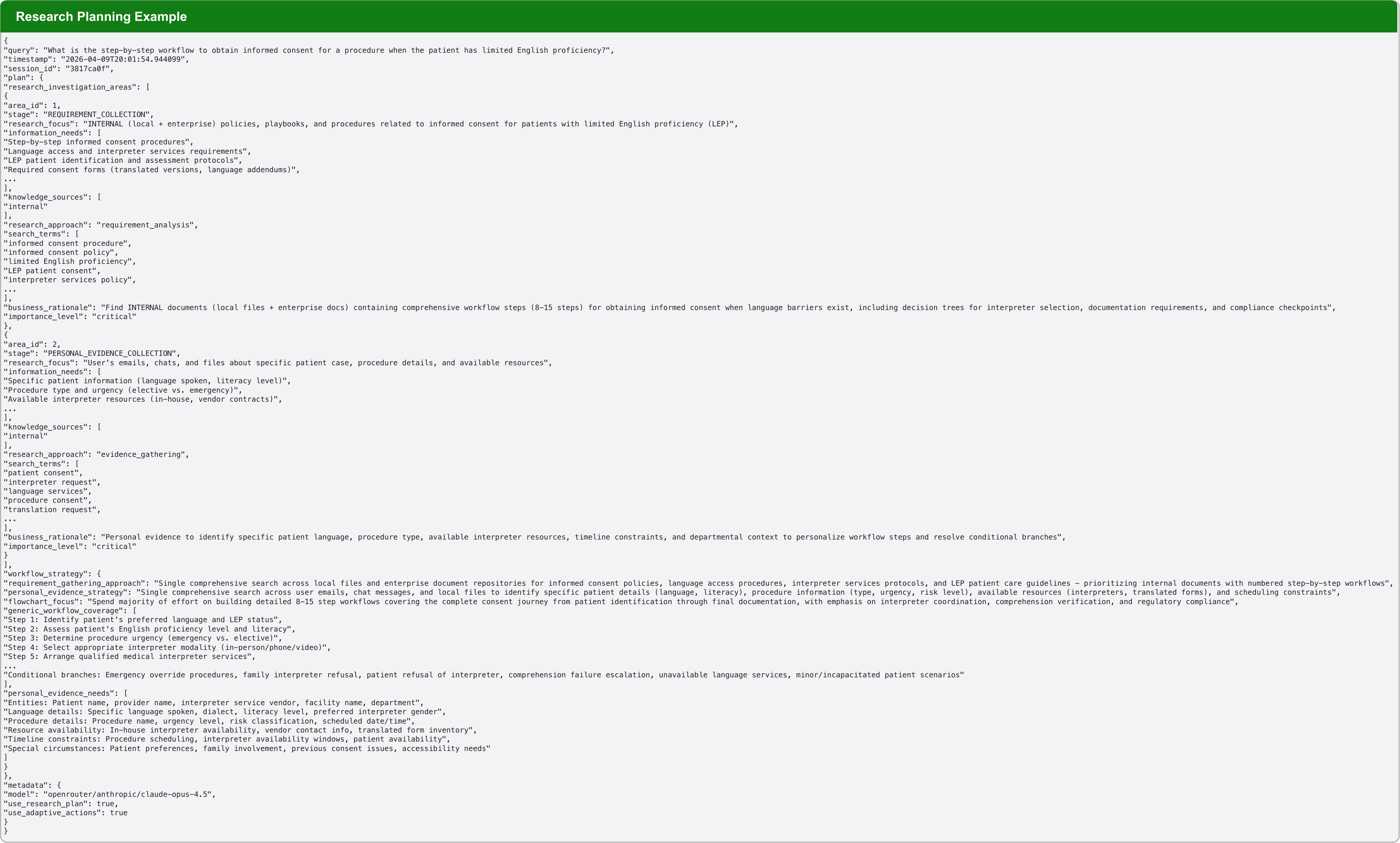}
    \caption{Example of a research plan produced by DRFA.}
    \label{fig:drfa_research_plan_example}
\end{figure*}

Each research investigation area is represented as a structured object with fields such as the research focus, information needs, expected knowledge sources, and workflow stage. This representation is intentionally execution-facing, so that it can be passed directly to the action planner without requiring an additional translation step. The research planning module therefore produces a decomposition aligned with workflow prediction and workflow personalization. A representative example of a research plan is shown in Figure~\ref{fig:drfa_research_plan_example}.

\subsection{Action Planning System}
\label{app:drfa_action_planning}

The action planning system converts research objectives into executable actions while managing tool choice, prioritization, and dependencies. Each action is represented as a typed record containing an action identifier, action type, description, argument dictionary, priority score, dependency list, expected output, preferred tool, and workflow stage. This representation makes the plan explicit enough to support iterative scheduling, tool-aware execution, and post hoc traceability.

The current action space includes retrieval actions over the web, enterprise systems, and local documents, as well as analysis-oriented actions for workflow prediction. The workflow stage labels used by the planner distinguish between requirement collection, personal evidence collection, generic workflow design, and workflow personalization. This stage-aware action representation is important because \textsc{DRFA} does not treat retrieval as a uniform process: different stages require different search targets and different criteria for success.

Tool priority follows a simple internal-first policy:
\begin{itemize}
    \item \textbf{Local document search} is preferred for uploaded documents, policies, playbooks, and other authoritative files.
    \item \textbf{Enterprise API search} is preferred for emails, chats, enterprise files, and other application-level records.
    \item \textbf{Web search} is used as a fallback when internal sources are insufficient or when external validation is required.
\end{itemize}

\begin{figure*}[t]
    \centering
    \includegraphics[width=\textwidth]{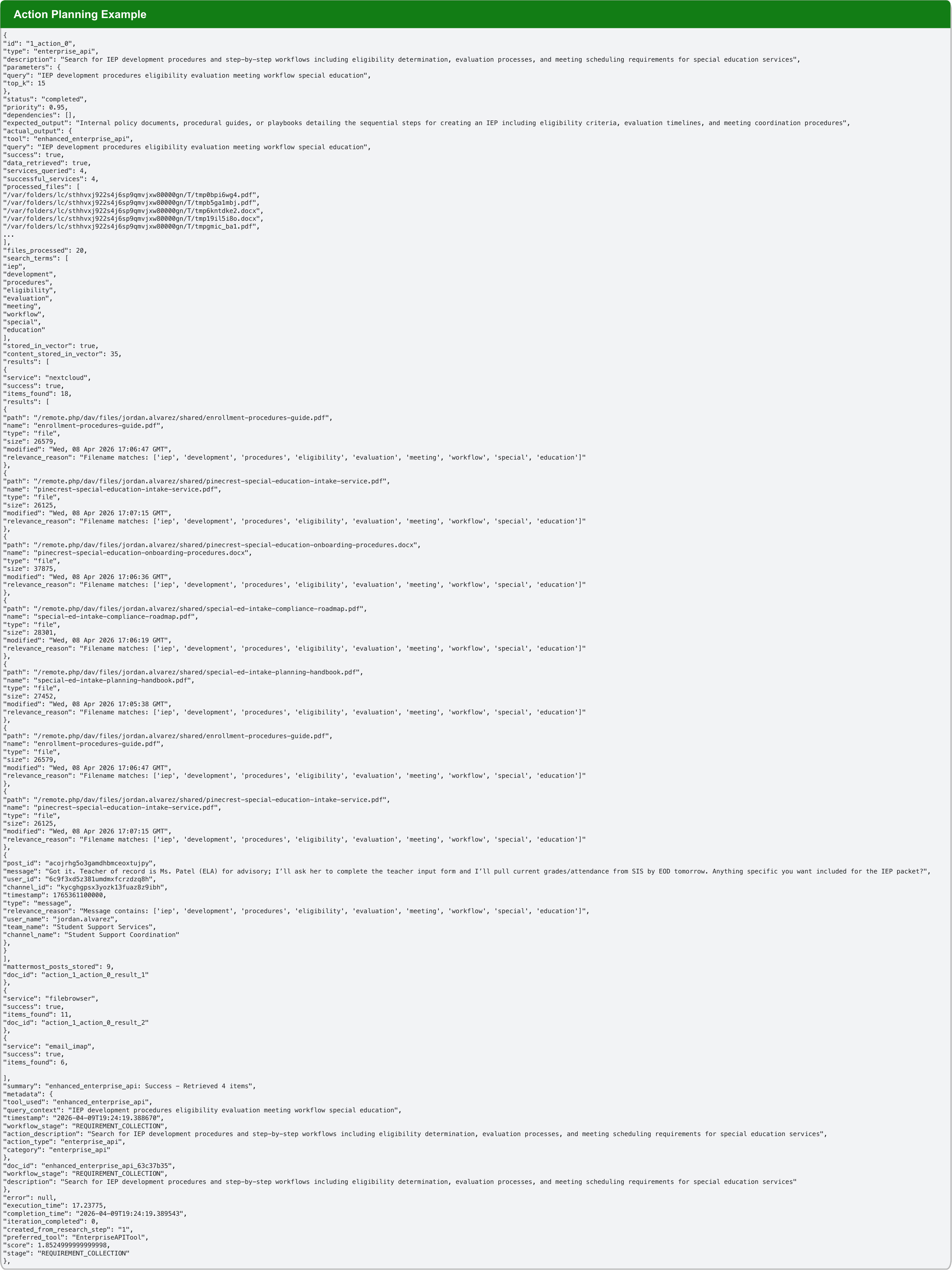}
    \caption{Example of an action plan produced by DRFA. The full planning is not shown for brevity.}
    \label{fig:drfa_action_plan_example}
\end{figure*}

Action priorities are assigned on a continuous scale and are further refined during execution using novelty and source-diversity signals. The scheduler favors actions that explore new evidence dimensions while balancing internal and external retrieval when appropriate. Dependency detection is handled separately, allowing the planner to serialize actions only when later steps genuinely depend on earlier outputs. A representative example of an action plan is shown in Figure~\ref{fig:drfa_action_plan_example}.

\subsection{Enterprise Integration Architecture}
\label{app:drfa_enterprise_integration}

\textsc{DRFA} supports retrieval over both enterprise applications and locally supplied files through a unified tool interface. Enterprise connectivity is provided by an enterprise API search tool, while local files are handled through a local document ingestion tool and a local semantic search tool. This makes it possible to run the agent in both enterprise-backed settings and file-backed settings without changing the high-level execution logic.

The integration layer preserves source metadata throughout retrieval and processing. As a result, downstream prediction stages can still distinguish among enterprise emails, enterprise chats, enterprise files, internal documents, and external sources. This source-type preservation is critical for workflow grounding, because the role of a piece of evidence depends not only on its content but also on whether it encodes organizational policy, personal context, or external background information.

\subsection{Tool Selection and Execution}
\label{app:drfa_tool_execution}

Tool execution is iterative and priority-driven. At each iteration, the scheduler retrieves the subset of pending actions whose dependencies have already been satisfied, computes their execution scores, and selects the next actions subject to the current concurrency limit. The default setting executes one action at a time, which simplifies traceability and makes it easier to attribute downstream workflow decisions to specific retrieval events, although the infrastructure supports bounded concurrency.

Tool selection itself follows a two-level policy. If the action planner specifies a preferred tool, the executor attempts to honor that choice directly. Otherwise, \textsc{DRFA} falls back to type-aware selection based on the action class and workflow stage. Internal and enterprise retrieval actions are routed toward local-document or enterprise API search tools, web-oriented actions are routed toward search or URL-fetching tools, and prediction-heavy actions are routed toward the analysis module.

Once an action is executed, its outputs are integrated back into the research context together with stage annotations, action-type metadata, execution timing, and any processed files created during execution. Significant results are also written into the vector store so that they remain available to later planning and workflow-generation stages.

\subsection{Adaptive Action Planning}
\label{app:drfa_adaptive_planning}

Adaptive action planning is the key mechanism that makes \textsc{DRFA} workflow-aware. At the end of each execution round, the agent re-examines the current evidence and decides whether the existing plan should be extended. This adaptation has two components: \textbf{gap finding} and \textbf{conditional action planning}.

The first component, gap finding, analyzes the current source composition and coverage of the action plan. If the evidence disproportionately covers only the generic side or only the personal side, the planner proposes a small number of complementary actions to reduce that imbalance. These actions are constrained by explicit diversity rules so that the agent does not repeatedly issue near-duplicate searches.

The second component, conditional action planning, extracts an intermediate generic workflow from the currently available policy and workflow evidence. It then identifies the subset of workflow steps that contain unresolved decisions. For each such step, the model returns a structured condition object containing a condition identifier, the corresponding step number, a natural-language description of the decision, the alternative branches, the type of personal evidence needed to resolve it, and a set of diverse search queries that are likely to surface that evidence. These condition objects are then converted into focused personal-evidence actions over emails, chats, records, and files.

\begin{figure*}[t]
    \centering
    \includegraphics[width=\textwidth]{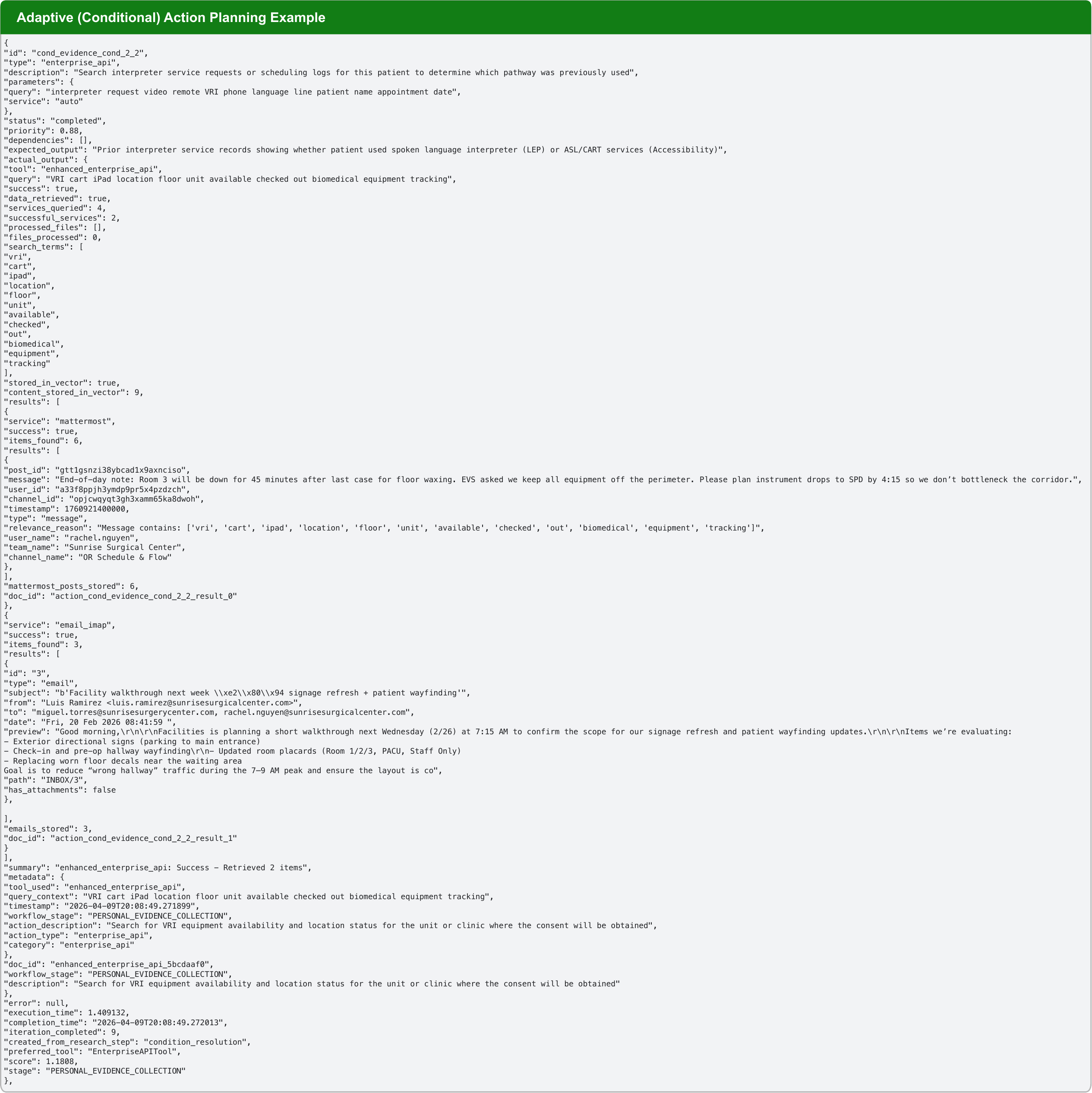}
    \caption{Example of an adaptive conditional action plan produced by DRFA. The full planning is not shown for brevity.}
    \label{fig:drfa_adaptive_plan_example}
\end{figure*}

Importantly, \textsc{DRFA} prioritizes conditional action planning when selecting the next action items. Newly proposed condition-resolution actions are inserted before more general gap-filling actions, and they typically receive high priority because they are the most directly useful for transforming a generic workflow into a personalized one. In effect, once the agent has constructed candidate workflow structure, it shifts from broad evidence gathering toward targeted branch disambiguation. A representative adaptive-planning example is shown in Figure~\ref{fig:drfa_adaptive_plan_example}.

\subsection{Content Processing and Vector Store}
\label{app:drfa_vector_store}

Similar to DRBench~\cite{abaskohi2025drbench}, the content-processing subsystem implements a unified ingestion pipeline for heterogeneous documents and retrieved results. Documents in different formats are normalized into text, cleaned, segmented, and embedded before being stored in the task-specific vector store. The vector store therefore acts as the main long-horizon memory of the agent.

This design is particularly important for workflow extraction. Evidence relevant to different workflow steps may be acquired several iterations apart, and the final workflow-generation stage must still be able to retrieve and jointly reason over these dispersed fragments. By storing both raw retrieved content and synthesized intermediate findings, the vector store enables semantic retrieval not only over original files but also over the agent’s own accumulated research state.

The current implementation creates an isolated vector store for each session, which prevents interference across tasks and preserves reproducibility of the retrieval context. Session-level metadata is also retained so that execution traces, saved plans, and workflow outputs can be inspected after generation.

\subsection{Workflow Generation and Citation Grounding}
\label{app:drfa_workflow_generation}

The workflow-generation module predicts the final output from the accumulated evidence. Its prediction procedure has three phases.

In the first phase, the agent gathers candidate evidence either from the full vector store or from the executed action plan alone. This stage collects both company-side workflow evidence and user-side personal evidence, while preserving their source metadata.

In the second phase, the system constructs a source registry for grounding. It filters invalid or unusable sources, normalizes source identities, registers valid documents in a citation registry, and maps them into a simplified citation namespace used during prompting. This registry is also responsible for preserving the source distinctions needed for later references, including internal documents, enterprise files, emails, chats, and external materials.

In the third phase, the model predicts the final workflow output. The generation prompt asks for a complete generic workflow, a user-situation analysis, a condition-resolution table, a personalized workflow, and step-by-step personalized guidance. The generic workflow must preserve prerequisite structure and explicitly record branch logic through a dedicated \texttt{condition\_description} field. The personalized workflow must preserve the same step inventory, but replace generic descriptions with user-specific facts wherever possible and clear the \texttt{condition\_description} field when the relevant branch has been resolved by evidence. When evidence remains insufficient, the unresolved condition is retained explicitly.

The agent ensures that each workflow step is grounded in citation-supported facts. The generator requires citations in step-level task descriptions, personalized justifications, and action guidance. After initial generation, the system performs targeted citation repair, validates that each personalized step contains at least one citation and a non-empty justification, resolves simplified citation identifiers back to real source identifiers, and finally produces a filtered references section. The citation registry is therefore not only a post-processing convenience, but part of the workflow-grounding mechanism itself.

\subsection{Prompts for DRFA}
\label{app:drfa_prompts}



\begin{figure*}[t]
    \centering
    \includegraphics[width=\textwidth]{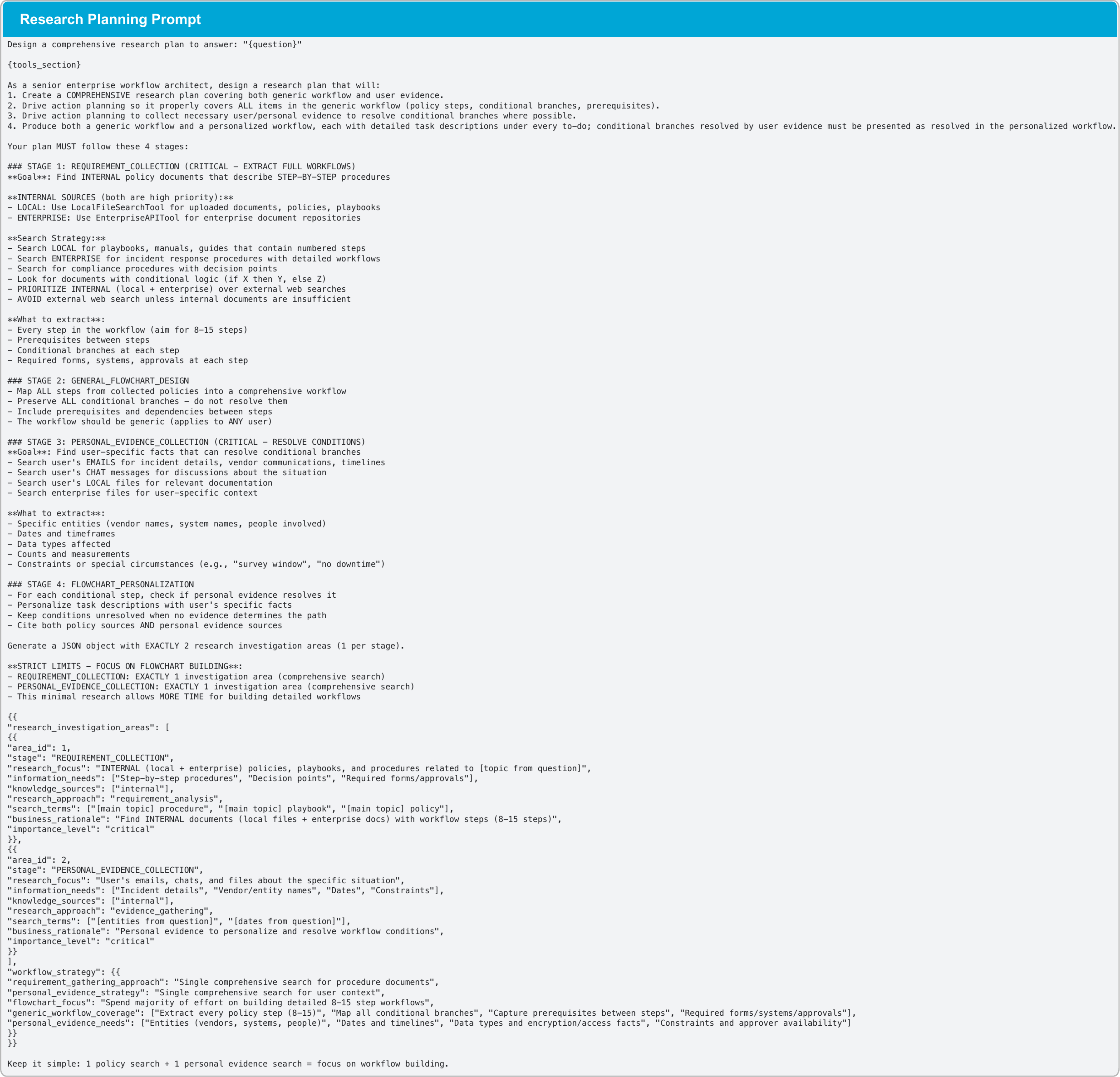}
    \caption{Research planning prompt used by the DRFA agent to construct generic and personalized workflows.}
    \label{fig:drfa_prompt_research_planning}
\end{figure*}

\paragraph{Research Planning Prompt.}
The research planning prompt decomposes the task into workflow-oriented investigation areas, with explicit distinction between generic requirement collection and personal evidence collection. We show the prompt in Figure~\ref{fig:drfa_prompt_research_planning}.


\begin{figure}[h]
    \centering
    \includegraphics[width=\columnwidth]{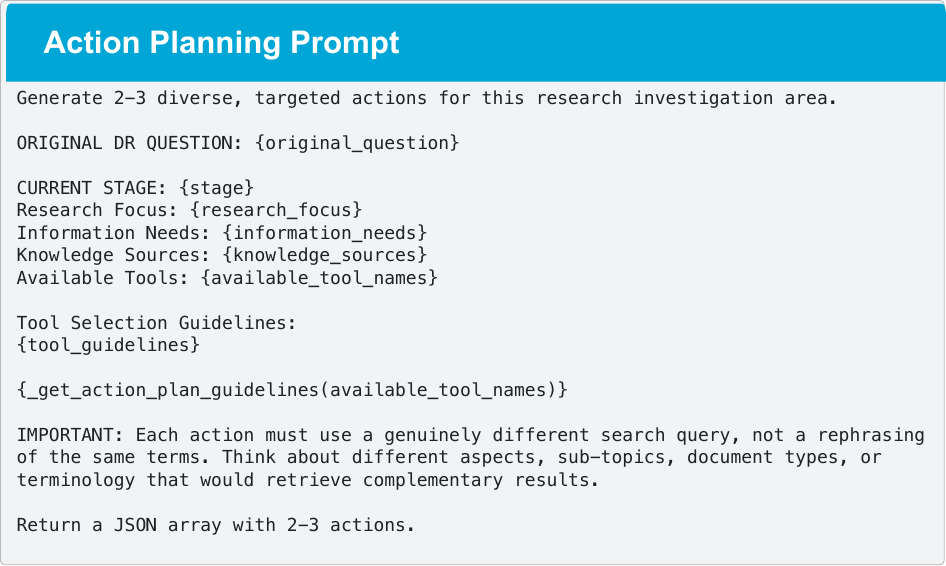}
    \caption{Action planning prompt used by DRFA agent.}
    \label{fig:drfa_prompt_action_generation}
\end{figure}

\paragraph{Action Generation Prompt.}
The action-generation prompt converts each investigation area into a small set of diverse executable actions, conditioning on the workflow stage, available tools, and internal-first source priority. Prompt is shown in Figure~\ref{fig:drfa_prompt_action_generation}.

\begin{figure}[h]
    \centering
    \includegraphics[width=\columnwidth]{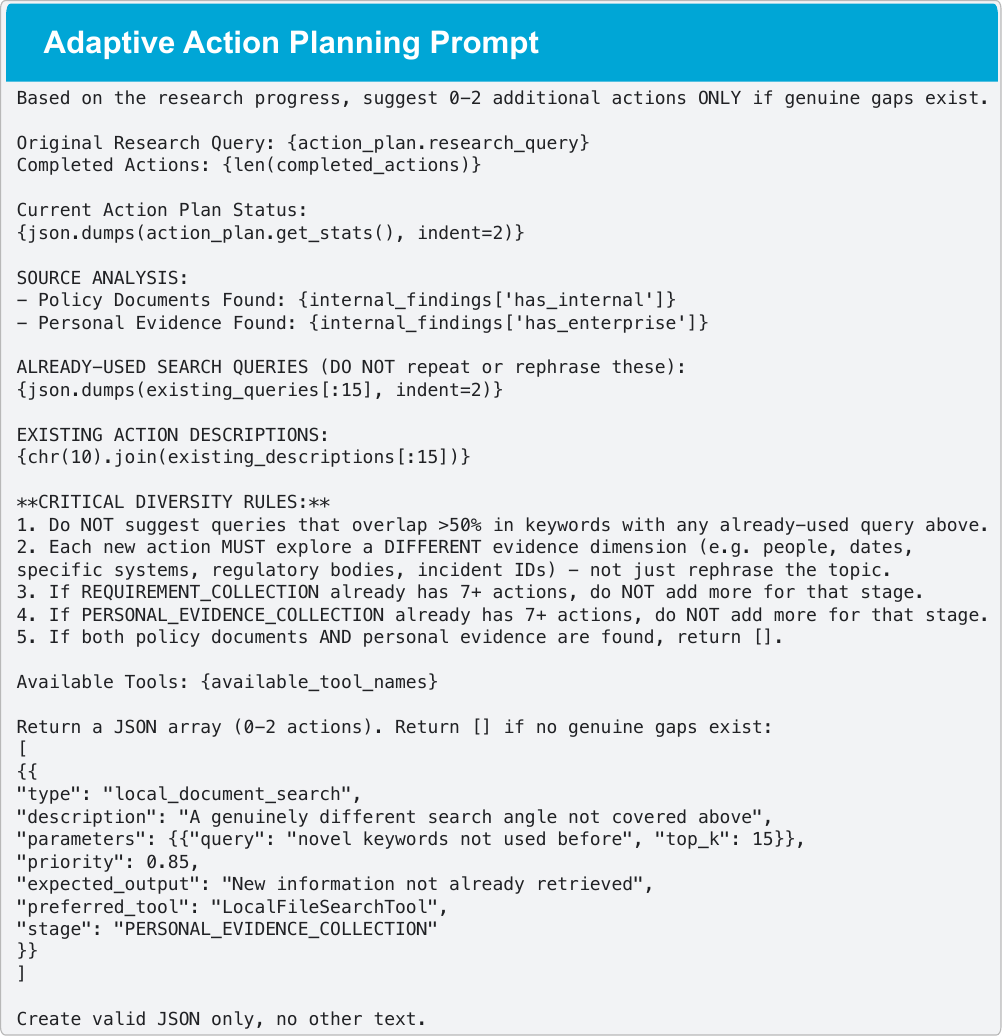}
    \caption{Adaptive planning prompt used by DRFA agent.}
    \label{fig:drfa_prompt_adaptive_planning}
\end{figure}

\begin{figure}[h]
    \centering
    \includegraphics[width=\columnwidth]{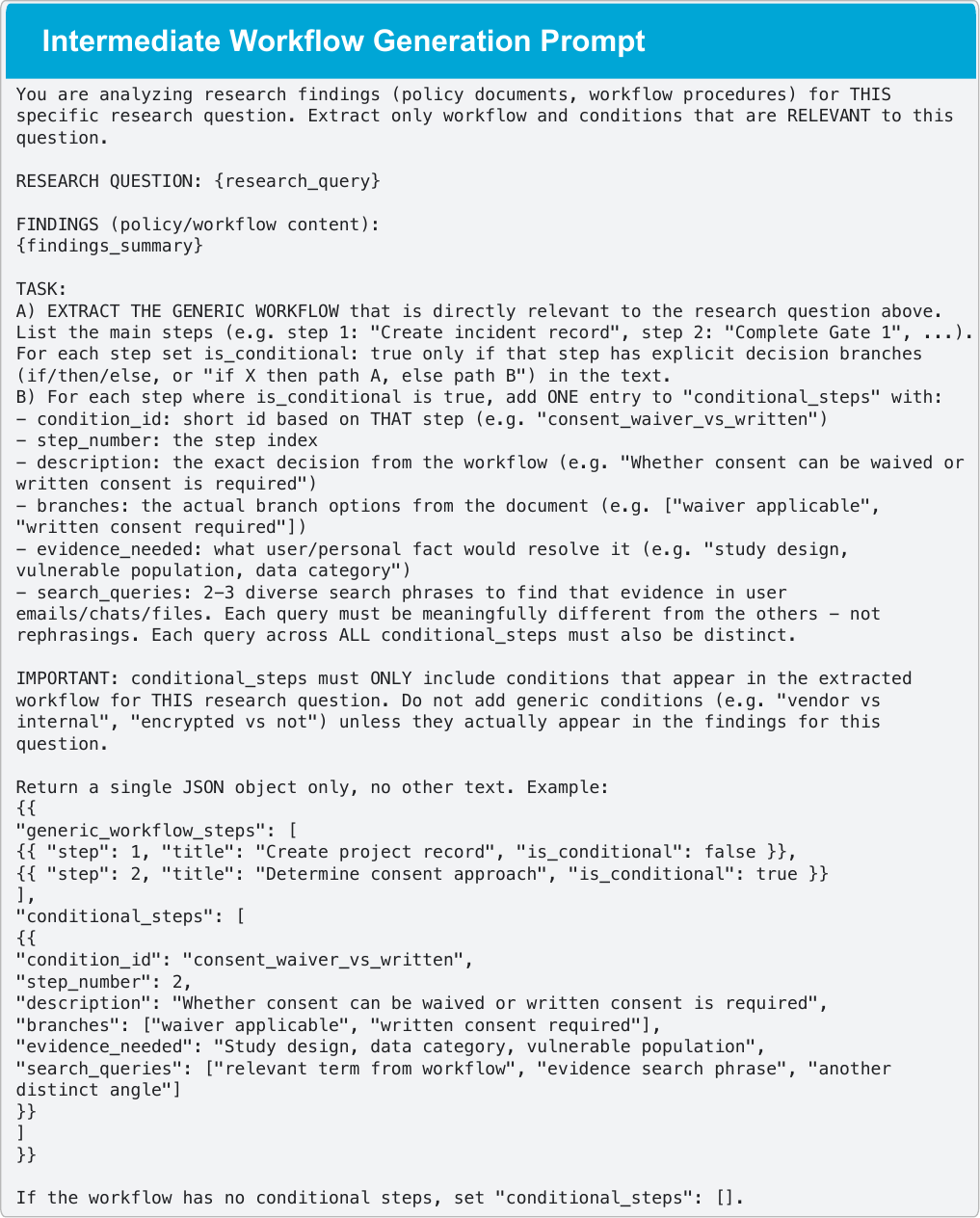}
    \caption{Prompt for intermediate workflow generation.}
    \label{fig:drfa_intermediate_workflow_generation}
\end{figure}

\begin{figure}[h]
    \centering
    \includegraphics[width=\columnwidth]{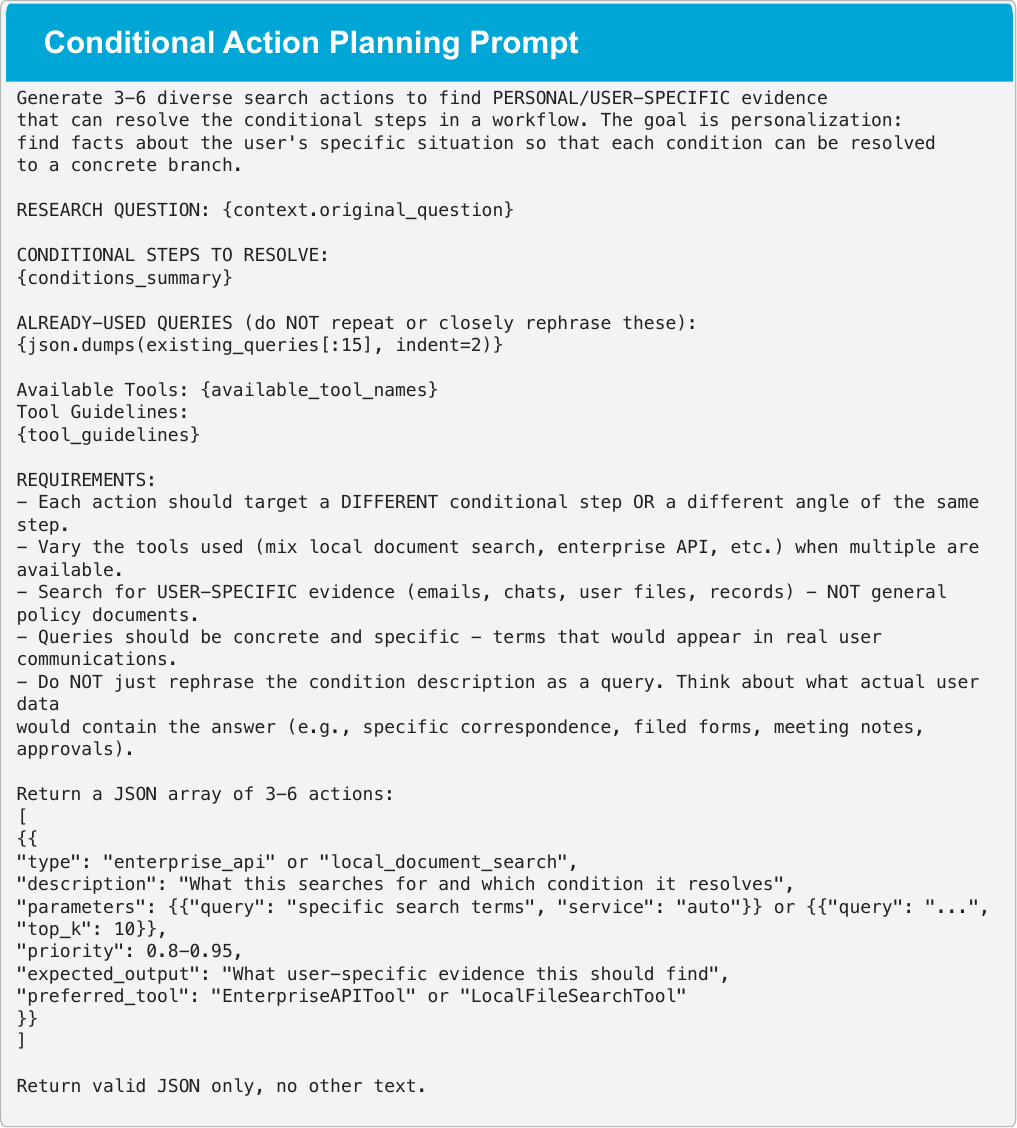}
    \caption{Conditional action planning prompt used by the DRFA agent.}
    \label{fig:drfa_prompt_conditional_planning}
\end{figure}

\paragraph{Adaptive Planning Prompt.}
The adaptive prompt supports two cases. In the primary case, it uses an intermediate generic workflow and its extracted conditions to propose targeted condition-resolution actions. In the fallback case, it proposes complementary gap-filling actions when the current evidence remains insufficient or unbalanced. We show the prompt for adaptive planning in Figure~\ref{fig:drfa_prompt_adaptive_planning} and the prompt for generating intermediate generic workflow in Figure~\ref{fig:drfa_intermediate_workflow_generation}, as well as conditional action planning to resolve the intermediate workflow in Figure~\ref{fig:drfa_prompt_conditional_planning}.

\begin{figure*}[t]
    \centering
    \includegraphics[width=\textwidth]{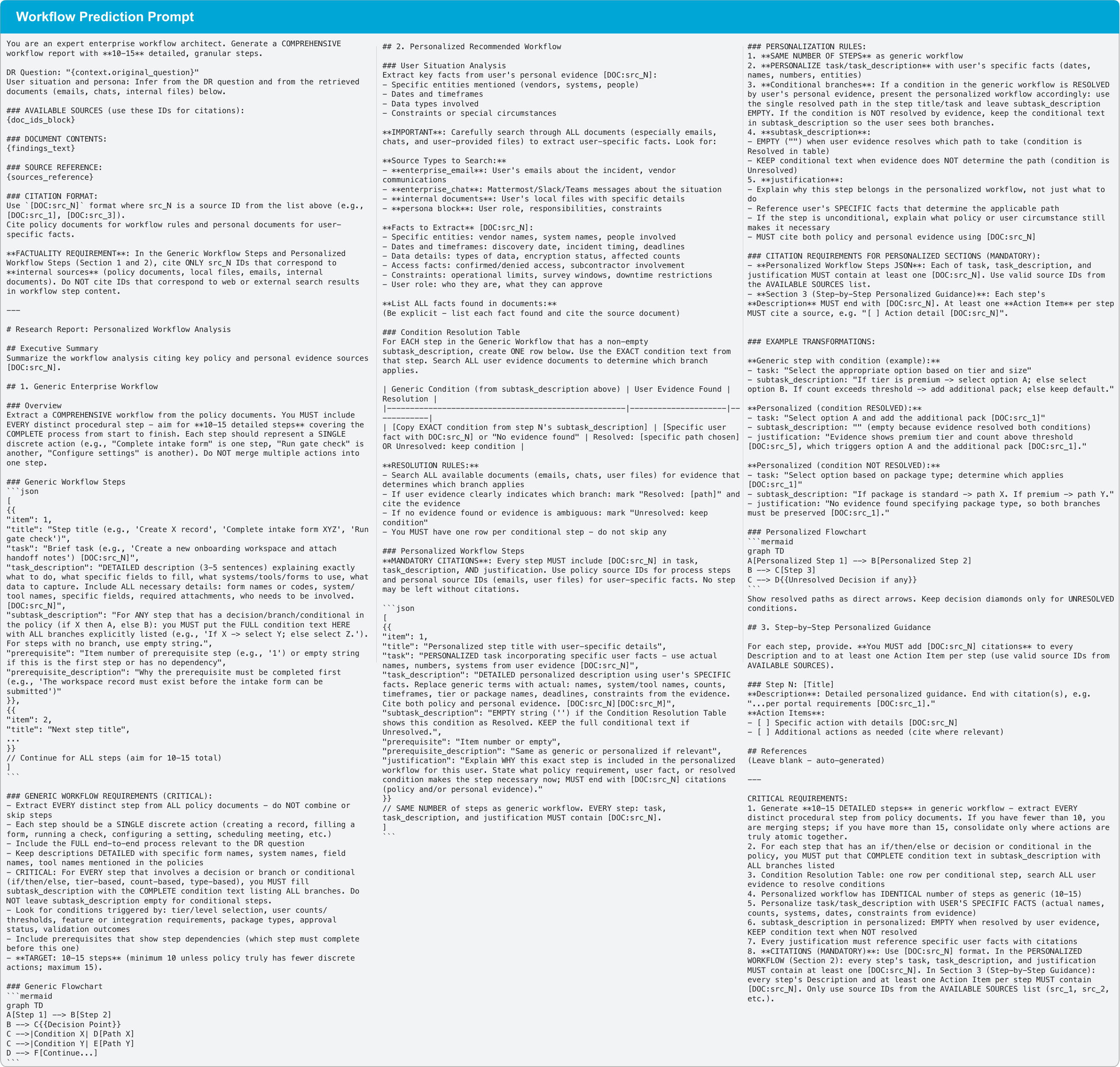}
    \caption{Workflow prediction prompt. \drflow~first generates a generic conditional workflow, then predicts the personalized workflow.}
    \label{fig:drfa_prompt_workflow_generation}
\end{figure*}

\paragraph{Workflow Prediction Prompt.}
The workflow prediction prompt predicts the final generic and personalized workflows, including condition resolution, personalized justifications, and step-level grounding. We show the prompt in Figure~\ref{fig:drfa_prompt_workflow_generation}.

\section{Detail Implementation of \drflow~Data Generation}
\label{appendix:data_generation}

This section describes the prompt families used to synthesize \drflow~tasks. The prompt inventory follows the benchmark construction pipeline described in the main text: task-seed generation, dual-context construction, company-side workflow evidence generation, user-side evidence generation and personalization, file realization, and introductory email synthesis. Together, these prompts create not only realistic artifacts, but also the latent workflow objects that define the benchmark targets.

\subsection{Task Seed Generation Prompts}
\label{app:task_seed_prompts}

The first prompt creates the \emph{task seed}, namely the deep research question together with company and user metadata. The main seed-generation prompt takes a target domain and asks the model to produce exactly $n$ (e.g., $10$) realistic, workflow-focused questions, each paired with company identity, user identity, and user-role metadata. The instructions constrain questions to be short, unambiguous, step-oriented, and realistic, while encouraging the company and user to be contextually aligned with the workflow question. This prompt is therefore the primary mechanism by which \drflow~instantiates diverse domain-specific tasks while preserving a consistent workflow-centric question style. The corresponding prompt template is shown in Figure~\ref{fig:prompt_dr_questions}.

\begin{figure*}[h]
    \centering
    \begin{tcolorbox}[
        width=0.98\textwidth,
        colback=PromptGray,
        colframe=gray!35,
        boxrule=0.5pt,
        arc=2pt,
        outer arc=2pt,
        left=0pt,
        right=0pt,
        top=0pt,
        bottom=0pt,
        boxsep=0pt,
        enhanced
    ]
        \begin{tcolorbox}[
            colback=PromptCyan,
            colframe=PromptCyan,
            boxrule=0pt,
            arc=2pt,
            outer arc=2pt,
            left=0.7em,
            right=0.7em,
            top=0.25em,
            bottom=0.25em,
            sharp corners=south
        ]
            {\color{white}\bfseries\small Task Seed Generation Prompt}
        \end{tcolorbox}

        \vspace{0.25em}
        \begin{minipage}{0.965\textwidth}
        \setlength{\columnsep}{1.2em}
        \begin{multicols}{2}
\begin{lstlisting}[style=promptstylecompact]
You are an expert at creating Deep Research (DR) questions for business and domain-specific workflows.

Your task is to generate exactly {n} Deep Research questions (with associated company and user metadata) for the domain: **{domain}**.

### Guidelines for each DR question:
1. Each question should be **simple to understand** and clearly ask about **what workflow or steps need to be taken** to accomplish a goal or resolve an issue.
2. Questions should be realistic, actionable, and require multi-step research or analysis to answer.
3. Each question should be at most 10-15 words, in plain English, easy to understand for non-technical people and without any leading phrases (e.g., "including A, B, C") and end with a question mark.
4. Questions should focus on understanding a process, procedure, or sequence of actions - e.g., "What steps do I need to follow to ...?", "What is the workflow for ...?", "What exact steps should I take to ...?"
5. Questions should not be ambiguous and should lead to concrete, step-by-step answers.
6. Do not combine multiple processes or procedures (E.g., do not ask questions like"What workflow should we follow to evaluate a positive blood culture, choose empiric antibiotics, de-escalate therapy, and document stewardship decisions?") into a single question. Keep it simple and focused.
7. The questions should be in the simple sentence. Do not make it complex and long.
8. Make questions diverse (different sub-areas or angles within the domain) rather than repetitive.
9. Make the questions from each domain unique, creative, and realistic.

### Guidelines for company and user metadata:
- For each question, generate realistic company and user information that fits the domain and the question scenario.
- The company and user should be contextually appropriate for the question being asked.
- Company persona represents a point of contact at the company (e.g., support agent, advisor, manager).
- User represents the person asking the deep research question.
- Do not use special characters in user names, company names, or email addresses.
- The deep research question should be relevant to the domain and the company and the user's role.
- Try to create the questions in such a way that the user is querying about their own company's internal processes. So, in this case, the user_company fields should match the company fields.

- Be creative with names, roles, and companies - make them feel realistic and varied.

### Examples of DR Questions (simple, workflow-focused):
- Healthcare: What is the step-by-step workflow to obtain informed consent for a procedure for a patient with limited English proficiency?
- Legal: What is the exact workflow I should follow to challenge my wrongful termination?
- B2C: What steps do I follow to recover a hacked customer account?
- B2B: What is the step-by-step process to diagnose and fix a broken order-to-invoice?
- Education: What workflow should I follow to apply for academic probation appeal in my courses?

### Output Format

Return a JSON object with a single key "dr_questions" whose value is a list of exactly {n} objects, each with the following keys:
{{{{
    "dr_questions": [
        {{{{
            "dr_question": "A simple, workflow-focused Deep Research question",
            "company_name": "Company Name",
            "company_industry": "Company Industry",
            "company_description": "A brief 1-2 sentences about the company.",
            "company_size": "Small/Medium/Large",
            "company_employee_count": "e.g., 100-500",
            "company_annual_revenue": "e.g., $50M-$100M",
            "company_persona": "Name of the company persona",
            "company_persona_email": "persona@company.com",
            "company_persona_role": "Role of the persona",
            "company_persona_role_description": "1 sentence description of the persona's role",
            "user_name": "User Name",
            "user_role": "User Role",
            "user_email": "user@email.com",
            "user_role_description": "1 sentence description of the user's role",
            "user_company": "User's Company",
            "user_industry": "User's Industry",
            "user_company_description": "A brief 1-2 sentences about the user's company.",
            "user_company_size": "Small/Medium/Large",
            "user_company_employee_count": "e.g., 100-500",
            "user_company_annual_revenue": "e.g., $50M-$100M"
        }}}}
    ]
}}}}

Do not include any preamble or explanation - return only the JSON object.
\end{lstlisting}
        \end{multicols}
        \end{minipage}
        \vspace{0.25em}
    \end{tcolorbox}
    \caption{Prompt  for task-seed generation for producing workflow-focused deep research questions together with company and user metadata.}
    \label{fig:prompt_dr_questions}
\end{figure*}

\subsection{Dual-Context Construction Prompt}
\label{app:dual_context_prompt}

Once the task seed has been created, the next prompt template constructs the two contextual objects that define the benchmark instance: the \emph{generic company context} and the \emph{personal user context}. The context-construction prompt receives the deep research question together with the company and user metadata, and asks the model to generate a generic workflow-like company policy and a user-specific personal context consistent with that policy. The prompt imposes several structural constraints that are central to \drflow: the generic context should contain $12$--$15$ steps, include both linear and conditional structure, remain self-contained, and support multi-hop reasoning; the personal context should satisfy only one branch when a generic condition is present and should be written as concise, workflow-aligned bullet points. This prompt therefore creates the latent pair $(C_i^{g}, C_i^{u})$ that underlies the later generic and personalized workflows. The corresponding template is shown in Figure~\ref{fig:prompt_dual_context}.

\begin{figure*}[h]
    \centering
    \begin{tcolorbox}[
        width=0.98\textwidth,
        colback=PromptGray,
        colframe=gray!35,
        boxrule=0.5pt,
        arc=2pt,
        outer arc=2pt,
        left=0pt,
        right=0pt,
        top=0pt,
        bottom=0pt,
        boxsep=0pt,
        enhanced
    ]
        \begin{tcolorbox}[
            colback=PromptCyan,
            colframe=PromptCyan,
            boxrule=0pt,
            arc=2pt,
            outer arc=2pt,
            left=0.7em,
            right=0.7em,
            top=0.25em,
            bottom=0.25em,
            sharp corners=south
        ]
            {\color{white}\bfseries\small Dual-Context Construction Prompt}
        \end{tcolorbox}

        \vspace{0.25em}
        \begin{minipage}{0.965\textwidth}
        \setlength{\columnsep}{1.2em}
        \begin{multicols}{2}
\begin{lstlisting}[style=promptstylecompact]
You are an expert Deep Research Workflow generator.
Your task is to generate the personal context and generic context for a deep research task given the existing information below.

## Existing Task Information
- **Domain:** {domain}
- **DR Question:** {dr_question}
- **Company Name:** {company_name}
- **Company Industry:** {company_industry}
- **Company Description:** {company_description}
- **Company Size:** {company_size}
- **Company Employee Count:** {company_employee_count}
- **Company Annual Revenue:** {company_annual_revenue}
- **Company Persona:** {company_persona}
- **Company Persona Email:** {company_persona_email}
- **Company Persona Role:** {company_persona_role}
- **Company Persona Role Description:** {company_persona_role_description}
- **User Name:** {user_name}
- **User Role:** {user_role}
- **User Email:** {user_email}
- **User Role Description:** {user_role_description}
- **User Company:** {user_company}
- **User Industry:** {user_industry}
- **User Company Description:** {user_company_description}
- **User Company Size:** {user_company_size}
- **User Company Employee Count:** {user_company_employee_count}
- **User Company Annual Revenue:** {user_company_annual_revenue}

## Your Task
Generate the `generic_context` and the `personal_context` fields. These must be realistic, relevant, and consistent with the information provided above.

### Guidelines for Generic Context:
- The generic context represents the company's internal process/workflow/policy that applies to all clients or employees in this domain.
- The generic context should have multiple constraints, similar to a flowchart.
- Keep the situation and language easy to understand. Do not make every step conditional. Keep many steps linear and sequential.
- The generic context should not be too long. The workflow should have 12-15 steps. Among them 6-8 steps should be conditional, the rest of the steps should not contain any conditional statements.
- Use words like 'Then', 'After that', 'Then do this', 'Then do that', etc. to describe the sequence of steps (except for the first step).
- It should be self-contained: all the information needed to generate the solution workflow for the user query must be present.
- It should include all the necessary information (e.g., form information, details). Although retrieving the action items should require multi-step reasoning. An LLM should not be able to retrieve all the information in a single linear pass.
- You should add intermediate information (e.g., forms and others) to ensure multi-step reasoning.
- Do not use complicated language. The language should be clear and simple.
- Keep each steps short, concise, simple, and easy to understand.
- You should add **multi-hop conditions** for the answer-generator model to think through several layers of reasoning.
- You can add several conditional statements, e.g., "If X, then do Y; else do Z" to increase complexity. E.g., if you have X, go to see rule 1.2. Then later, if you see 1.2, submit form Z; if you see 1.3, submit Q. Focus on adding multi-hop conditions that can be **subtle** and require critical reasoning.
- Add different situational details, environmental factors, or hypothetical scenarios in the context that require the model to think critically. For example, adding different legal regulations, forms, multi-step procedures can increase complexity.
- Do not directly add the action items in the context. Add them subtly and through multi-step conditioning to make the model work hard on retrieving them.
- Try to be creative, professional, and realistic.
- For conditional steps, clearly define the conditions and the actions to be taken for each condition. Do not use complicated language for conditional steps.
- For conditional steps, do not use the same condition for multiple actions.
- Remember, for a conditional step, do not add more than 2-3 conditions.
- The generic context MUST be relevant to the company ({company_name}, {company_industry}) and the DR question.

### Guidelines for Personal Context:
- The personal context should be relevant to the user's query, user's role ({user_role}), and user's company ({user_company}).
- You should add information in such a way that it follows the workflow flowchart of the generic context.
- In case there are conditional steps in the generic context, the personal context should satisfy only one of the conditions.
- Do not add information that is not related to the generic context. You do not need to introduce yourself (e.g., "I am a...") in the personal context. Also, do not mention the user's name in the personal context.
- Create a personal context that is easy to understand and follows the workflow flowchart of the generic context.
- Create the personal context in a bullet point format resembling the generic context so that it is easy to understand and follow.
- Keep each point short, concise, simple, and easy to understand.
- The personal context MUST be consistent with the user's role ({user_role}), company ({user_company}), and the DR question.

Return the output as a JSON object with the following structure:
{{
    "generic_context": "Generic context (company's internal process/workflow/policy applicable to all clients/employees)",
    "personal_context": "Personal context (user's specific situation following the generic context workflow)"
}}

Below is an example of generic context and personal context for the legal domain.
- REMEMBER be creative, realistic, and professional.
- Adhere to the domain-specific task creation.
- Make the contexts relevant to the existing task information above.
- Do not use complicated language. The language should be clear and simple.
- Do not exactly follow the wording or the theme of the example, but follow the structure and the format.

### Example:
Given information:
- **Domain:** Legal
- **DR Question:** "I want to know if I have legal grounds to challenge my termination and what EXACT steps I should take next. Tell me what documents to prepare, whom to contact, and what procedures I must follow according to your law firm's policies."
- **Company Name:** GateBridge Employment Law
- **Company Industry:** Legal (Employment Law)
- **Company Persona Role:** Senior Case Intake Coordinator

Generated output:
{{{{
    "generic_context": {COMPANY_INFO_GENERIC_CONTEXT},
    "personal_context": {USER_INFO_GENERIC_CONTEXT}
}}}}
\end{lstlisting}
        \end{multicols}
        \end{minipage}
        \vspace{0.25em}
    \end{tcolorbox}
    \caption{Prompt template for constructing the generic company context and the personal user context.}
    \label{fig:prompt_dual_context}
\end{figure*}

\subsection{Company Workflow Evidence Prompts}
\label{app:company_evidence_prompts}

The company-side evidence prompts convert the generic context into workflow-bearing supervision. The company insight prompt takes the generic context and asks the model to generate an ordered list of actionable workflow steps, each with a title, task, task description, condition description, prerequisite pointer, and prerequisite description. The prompt explicitly instructs the model to preserve the order of the generic context, maintain the same number of steps, and avoid splitting conditional branches into separate top-level steps. This makes the output suitable for defining the gold generic workflow and for downstream file realization. The corresponding prompt template is shown in Figure~\ref{fig:prompt_company_insights}.

The paired company distractor prompt produces plausible but irrelevant company-side steps. Its instructions emphasize topical plausibility and business realism while requiring that the steps remain tangential to the deep research question. In practice, this prompt creates adjacent but non-resolving procedures that populate distractor files and increase retrieval difficulty without altering the target workflow. The corresponding prompt template is shown in Figure~\ref{fig:prompt_company_distractors}.

\begin{figure}[h]
    \centering
    \begin{tcolorbox}[
        width=0.98\columnwidth,
        colback=PromptGray,
        colframe=gray!35,
        boxrule=0.5pt,
        arc=2pt,
        outer arc=2pt,
        left=0pt,
        right=0pt,
        top=0pt,
        bottom=0pt,
        boxsep=0pt,
        enhanced
    ]
        \begin{tcolorbox}[
            colback=PromptCyan,
            colframe=PromptCyan,
            boxrule=0pt,
            arc=2pt,
            outer arc=2pt,
            left=0.7em,
            right=0.7em,
            top=0.25em,
            bottom=0.25em,
            sharp corners=south
        ]
            {\color{white}\bfseries\small Company Insights for Workflow Generation Prompt}
        \end{tcolorbox}

        \vspace{0.25em}
        \begin{minipage}{0.965\textwidth}
        \setlength{\columnsep}{1.2em}
\begin{lstlisting}[style=promptstylecompact]
You are an expert workflow process analyst. Your task is to analyze the generic context and documentation of a company and create a detailed list of to-do steps (insights) covering that context.

Company Name: {company_name}
Industry: {industry}
Generic Context: {generic_context}

Please create a list of actionable steps. For each step, provide:
1. An item index.
2. A concise title.
3. The specific task to be performed.
4. A detailed low-level description of how to perform the task (e.g., fill out name, address of form X, ignore section B, etc). Be concise and do not itemize the task description.
5. Any subtask descriptions or conditional branches (e.g., if P happens submit A, if Q happens go to B, etc). If there are no conditional branches, leave this field empty.
6. Which item is a prerequisite for this item. If none, keep empty.
7. A detailed low-level description of the prerequisite (e.g., user needs to collect the confirmation number from page X, user needs to submit form Y first before doing this item). If none, keep empty.
8. For determining the prerequisite, you may look for the clues like 'after that', 'then', etc. This implies that the previous item is a prerequisite for the current item.

Remember,
- The steps should be in the order of the generic context.
- Number of steps should be equal to the number of steps in the generic context.
- You do not need to create separate steps for the subtasks or conditional branches of the generic context.

Return the output as a JSON object with a 'steps' key containing an array of items following the format specified.
\end{lstlisting}
        \end{minipage}
        \vspace{0.25em}
    \end{tcolorbox}
    \caption{Prompt for company workflow insight generation.}
    \label{fig:prompt_company_insights}
\end{figure}

\begin{figure}[h]
    \centering
    \begin{tcolorbox}[
        width=0.98\columnwidth,
        colback=PromptGray,
        colframe=gray!35,
        boxrule=0.5pt,
        arc=2pt,
        outer arc=2pt,
        left=0pt,
        right=0pt,
        top=0pt,
        bottom=0pt,
        boxsep=0pt,
        enhanced
    ]
        \begin{tcolorbox}[
            colback=PromptCyan,
            colframe=PromptCyan,
            boxrule=0pt,
            arc=2pt,
            outer arc=2pt,
            left=0.7em,
            right=0.7em,
            top=0.25em,
            bottom=0.25em,
            sharp corners=south
        ]
            {\color{white}\bfseries\small Company Distractor Generation Prompt}
        \end{tcolorbox}

        \vspace{0.25em}
        \begin{minipage}{0.965\textwidth}
        \setlength{\columnsep}{1.2em}
\begin{lstlisting}[style=promptstylecompact]
You are an expert process analyst. Your task is to analyze the description of a company and its Deep Research (DR) Question, and create a list of distractors. These distractors should be plausible to-do steps that are NOT relevant to addressing the DR Question.

Company Name: {company_name}
Industry: {industry}
Company Description: {description}
DR Question: {dr_question}

DISTRACTOR Requirements:
- Generate {n_distractors} to-do steps that are plausible for this company and industry but DO NOT help address the DR Question.
- The steps should be about the company's general operations, metrics, or business processes but tangential to the core focus of the DR Question.
- Focus on business areas unrelated to the DR Question (e.g., if the DR Question is about finance, generate steps about HR, IT infrastructure, facility maintenance, or compliance training).
- Each step should follow the order of business logic but remain irrelevant to the research goal.

For each distractor step, provide:
1. An item index.
2. A concise title.
3. The specific task to be performed.
4. A detailed low-level description of the task.
5. Any subtask descriptions or conditional branches. If none, keep empty.

Return the output as a JSON object with a 'steps' key containing an array of items following the format specified.
\end{lstlisting}
        \end{minipage}
        \vspace{0.25em}
    \end{tcolorbox}
    \caption{Prompt for company distractor generation.}
    \label{fig:prompt_company_distractors}
\end{figure}

\subsection{User Evidence and Personalized Workflow Prompts}
\label{app:user_evidence_prompts}

The user-side prompt family serves two distinct functions: generating user-specific supporting and distracting evidence, and composing the personalized workflow.

The user insight prompt converts the personal context into a set of question-answer-justification tuples. The crucial design choice is that answers must be extracted verbatim from the personal context, while the generated questions and justifications frame those facts as evidence relevant to the deep research task. The prompt also requires full coverage of the personal context across the generated tuples, ensuring that user-side support evidence spans the complete personalized scenario rather than only a small subset of salient facts. The corresponding template is shown in Figure~\ref{fig:prompt_user_insights}.

The paired user distractor prompt generates plausible but task-irrelevant user-side events. These distractors are framed as realistic workplace operations, metrics, or administrative events that are compatible with the user’s role but do not help resolve the deep research question or the workflow branches implied by the company context. The corresponding template is shown in Figure~\ref{fig:prompt_user_distractors}.

The final prompt in this family is the personalized workflow prompt. It takes the generic company workflow and the user facts as input, then asks the model to resolve conditional paths, personalize task descriptions, preserve workflow structure, and provide a justification for each step. The prompt explicitly states that resolved branches should collapse into the applicable path, while unresolved branches should remain visible. It also requires that every personalized step retain sufficient procedural detail and include a natural-language justification grounded in user facts. This prompt is therefore responsible for constructing the gold personalized workflow used in evaluation. The corresponding template is shown in Figure~\ref{fig:prompt_user_workflow}.

\begin{figure}[h]
    \centering
    \begin{tcolorbox}[
        width=0.98\columnwidth,
        colback=PromptGray,
        colframe=gray!35,
        boxrule=0.5pt,
        arc=2pt,
        outer arc=2pt,
        left=0pt,
        right=0pt,
        top=0pt,
        bottom=0pt,
        boxsep=0pt,
        enhanced
    ]
        \begin{tcolorbox}[
            colback=PromptCyan,
            colframe=PromptCyan,
            boxrule=0pt,
            arc=2pt,
            outer arc=2pt,
            left=0.7em,
            right=0.7em,
            top=0.25em,
            bottom=0.25em,
            sharp corners=south
        ]
            {\color{white}\bfseries\small User Support-Evidence Generation Prompt}
        \end{tcolorbox}

        \vspace{0.25em}
        \begin{minipage}{0.965\textwidth}
        \setlength{\columnsep}{1.2em}
\begin{lstlisting}[style=promptstylecompact]
You are an expert QA analyst. Your task is to analyze the user's personal context and their Deep Research (DR) Question, and create a list of {n_supports} question-answering-justification pairs. Each pair must follow a question-answer-justification format.

DR Question: {dr_question}
User Context: {context}
Number of Question-Answering Pairs: {n_supports}

Requirements:
1. The 'answer' for each pair MUST be extracted from the context and must be an exact statement or event directly present in the given User Context. Do not summarize or rephrase the answer; it must be a verbatim extraction.
2. The 'question' should be formulated such that the extracted 'answer' provides a direct response to it.
3. The 'justification' should explain why this event is relevant to the user's situation or the DR question. You may synthesize this justification, but keep it realistic and professional.
4. You must cover the entire context within the {n_supports} question-answering-justification pairs. Do not miss any single sentence from the context.
5. You must generate exactly {n_supports} question-answering-justification pairs.
6. If necessary, you can merge multiple questions to cover the entire context within the {n_supports} question-answering-justification pairs.
7. Remember, You MUST COVER every points/information of the context within the {n_supports} question-answering-justification pairs. Do not miss any single point/information from the context.

Return the output as a JSON object with an 'items' key containing an array of events following the format specified.
\end{lstlisting}
        \end{minipage}
        \vspace{0.25em}
    \end{tcolorbox}
    \caption{Prompt for user support-evidence generation.}
    \label{fig:prompt_user_insights}
\end{figure}

\begin{figure}[h]
    \centering
    \begin{tcolorbox}[
        width=0.98\columnwidth,
        colback=PromptGray,
        colframe=gray!35,
        boxrule=0.5pt,
        arc=2pt,
        outer arc=2pt,
        left=0pt,
        right=0pt,
        top=0pt,
        bottom=0pt,
        boxsep=0pt,
        enhanced
    ]
        \begin{tcolorbox}[
            colback=PromptCyan,
            colframe=PromptCyan,
            boxrule=0pt,
            arc=2pt,
            outer arc=2pt,
            left=0.7em,
            right=0.7em,
            top=0.25em,
            bottom=0.25em,
            sharp corners=south
        ]
            {\color{white}\bfseries\small User Distractor Generation Prompt}
        \end{tcolorbox}

        \vspace{0.25em}
        \begin{minipage}{0.965\textwidth}
        \setlength{\columnsep}{1.2em}
\begin{lstlisting}[style=promptstylecompact]
You are an expert QA analyst. Your task is to analyze the user's personal context and their Deep Research (DR) Question, and create {n_distractors} distractor events that are NOT relevant to answering the DR Question or the user's personal context.

DR Question: {dr_question}
User Context: {context}

Requirements:
- These events should be plausible but MUST NOT be relevant to the DR Question or the user's personal context.
- The events should be about the user's workplace general operations, metrics, business processes but irrelevant to the DR Question or the user's personal context.
- Focus on the areas unrelated to the DR Question (e.g., if the DR Question is about finance, generate steps about other things such as user's personal business, HR, IT infrastructure, facility maintenance, or compliance training).

Return the output as a JSON object with an 'items' key containing an array of events following the format specified.
\end{lstlisting}
        \end{minipage}
        \vspace{0.25em}
    \end{tcolorbox}
    \caption{Prompt for user distractor-event generation.}
    \label{fig:prompt_user_distractors}
\end{figure}

\begin{figure*}[t]
    \centering
    \begin{tcolorbox}[
        width=0.99\textwidth,
        colback=PromptGray,
        colframe=gray!35,
        boxrule=0.45pt,
        arc=2pt,
        outer arc=2pt,
        left=0.35em,
        right=0.35em,
        top=0pt,
        bottom=0.25em,
        boxsep=0pt
    ]
        \begin{tcolorbox}[
            colback=PromptCyan,
            colframe=PromptCyan,
            boxrule=0pt,
            arc=2pt,
            outer arc=2pt,
            left=0.6em,
            right=0.6em,
            top=0.18em,
            bottom=0.18em,
            sharp corners=south
        ]
            {\color{white}\bfseries\footnotesize Personalized Workflow Generation Prompt}
        \end{tcolorbox}

        \vspace{0.15em}

        \begin{minipage}[t]{0.465\textwidth}
\begin{Verbatim}
You are an expert personalized workflow generator. Your task is to take a deep research question (DR Question), generic company workflow (Generic Insights) and a specific user's facts (User Context), and generate a personalized, resolved workflow for that user.

DR Question:
{dr_question}

Generic Insights (Company Workflow):
{generic_insights}

User Context (User Facts):
{user_context}

Requirements:
1. **Resolve Conditional Paths**: The Generic Insights often contain conditional branches (e.g., "If X, then do Y; else do Z"). Use the provided User Context to determine which path the user should take. If resolved, the 'subtask_description' will be empty. If information to resolve conditional path does not exist in the provided user insights, keep the 'subtask_description' as is.
2. **Personalize Instructions**: Adapt the 'task', 'task_description' to be specific to the user's situation where possible. For example, if there are multiple conditions and only one of them is applicable for the user, then you should modify the 'task' and 'task_description' to clearly define which path the user should take to make it more personalized.
3. **Justify Applicability**: For each step, provide a 'justification' that explains why this step is relevant to the user, citing specific facts from the User Insights.
4. **Maintain Structure**: Keep the same sequence and logical flow as the Generic Insights, but remove irrelevant conditional branches.
5. **Preserve Required Detail**: Every step must include a non-empty 'task_description' and non-empty 'justification'. Keep 'prerequisite' and 'prerequisite_description' populated whenever they exist in the Generic Insights unless you are intentionally updating them to a more specific user-resolved version. Only leave 'subtask_description' empty when the Generic Insight had no branch or the branch was fully resolved for the user.
6. **Format**: Return the output as a JSON object with a 'steps' key containing an array of UserWorkflowItem objects.

Return ONLY a valid JSON object.

### Example Workflow:
DR Question:
I'm the privacy/compliance lead at a multi-site hospital network and we just detected a potential PHI exposure involving a vendor-connected patient portal. I need a step-by-step plan to determine whether this is a reportable HIPAA breach, whether any state notification rules are triggered, and what EXACT documents, forms, internal approvals, and deadlines we must follow under our incident program-while keeping operations running and preserving evidence.

Generic Insights (Company Workflow):

[
  {{
    "item": 1,
    "title": "Open incident record in CIP",
    "task": "Create a new incident record in the Compliance Incident Portal (CIP).",
    "task_description": "Log into CIP and select the option to create/open a new incident record. Enter the initial incident details required to save the record (e.g., incident title/summary, date/time discovered, reporter/contact, impacted service/system, initial description). Save the record to generate the incident ID and enable downstream forms/workflow sections.",
    "subtask_description": "",
    "prerequisite": "",
    "prerequisite_description": ""
  }},
  {{
    "item": 2,
    "title": "Select Intake Track using Origin rule (Track V vs Track I)",
    "task": "Choose the correct CIP Intake Track based on whether a contracted vendor system/credential is involved.",
    "task_description": "In the incident record\u2019s intake/triage section, locate the field that sets the Intake Track (or equivalent workflow routing selector). Apply the Origin rule: determine whether the event touches (a) a contracted vendor system, integration, hosted environment, or (b) vendor credential (e.g., vendor user account, API key, SSO identity). Set Track V (Vendor-Connected) if vendor-touched; otherwise set Track I (Internal Systems). Save/confirm the track selection so CIP presents the correct downstream forms and required fields.",
    "subtask_description": "If the event touches a contracted vendor system or vendor credential, route to Track V (Vendor-Connected). Else, route to Track I (Internal Systems).",
    "prerequisite": 1.0,
    "prerequisite_description": "An incident record must exist in CIP (incident ID created) so the Intake Track can be set on that record and control which later forms appear."
  }}
]
\end{Verbatim}
        \end{minipage}
        \hfill
        \begin{minipage}[t]{0.465\textwidth}
\begin{Verbatim}
User Context (User Facts):
On Friday at 4:40 pm, Northbridge Health Partners (a 6-hospital network) received alerts that a third-party portal plugin (used for scheduling and lab result viewing) was misconfigured. The vendor says the issue was limited to 36 hours and involved an unindexed ``share link" feature that may have allowed anyone with the link to view a patient's lab PDF. Internal logs show 1,820 unique link generations, but only 214 link opens from external IPs; it's unclear if those opens were patients, family, or unknown parties. The PDFs include patient name, DOB, MRN, test name, and ordering provider; no SSNs or payment cards. The affected patients are from two states: Illinois and Wisconsin. The plugin is used only by the outpatient clinics, not inpatient. The vendor is a contracted business associate and the BAA is active, but the vendor initially emailed details from a personal Gmail account and has not provided a full incident report yet. The network is mid-Joint Commission survey window, so downtime is strongly discouraged; leadership wants containment without taking the portal fully offline. There is no evidence yet of ransomware, data exfiltration tools, or mass scraping. The incident was first discovered by the SOC (not by a patient complaint). This is the first potential breach in 18 months, and the organization's policy requires routing through the internal IR workflow. One condition is satisfied: this is a vendor/BA-facing web exposure with confirmed external access in logs, but no confirmed acquisition beyond link opens.

### Output:

[
  {{
    "item": 1,
    "title": "Open incident record in CIP",
    "task": "Create a new incident record in the Compliance Incident Portal (CIP) for the third-party portal plugin misconfiguration.",
    "task_description": "Log into CIP and create a new incident record for the third-party scheduling/lab-result portal plugin exposure discovered Friday at 4:40 pm by the SOC. Populate the minimum required fields to save: incident title (e.g., \u201cVendor portal plugin share-link misconfiguration\u2014possible lab PDF exposure\u201d), date/time discovered (Fri 4:40 pm), reporter (SOC), impacted system/service (outpatient portal plugin used for scheduling and lab result viewing), vendor/business associate name, and a short initial description (unindexed \u201cshare link\u201d feature potentially allowed link-based viewing of lab PDFs for ~36 hours; internal logs show 1,820 link generations and 214 external IP opens). Save to generate the CIP incident ID and enable the track-specific workflow sections.",
    "subtask_description": "",
    "prerequisite": "",
    "prerequisite_description": "",
    "justification": "An official record is required because this is a potential PHI exposure involving a contracted business associate portal plugin and the organization\u2019s policy requires routing through the internal IR workflow."
  }},
  {{
    "item": 2,
    "title": "Select Intake Track using Origin rule (Track V vs Track I)",
    "task": "Set Intake Track to Track V (Vendor-Connected) due to contracted vendor/BA plugin involvement.",
    "task_description": "In the incident intake/triage section of the CIP record, set the Intake Track to Track V (Vendor-Connected). Treat this as vendor-connected because the incident involves a contracted business associate\u2019s portal plugin and vendor-provided details/timeline will be required for investigation. Save/confirm Track V so CIP displays vendor-related fields and downstream forms.",
    "subtask_description": "",
    "prerequisite": 1.0,
    "prerequisite_description": "An incident record must exist in CIP (incident ID created) so the Intake Track can be set on that record and control which later forms appear.",
    "justification": "The event centers on a third-party portal plugin operated by a contracted business associate under an active BAA, which meets the vendor-touched criterion for Track V."
  }}
]
\end{Verbatim}
        \end{minipage}

        \vspace{0.15em}
    \end{tcolorbox}
    \caption{Prompt for personalized workflow generation.}
    \label{fig:prompt_user_workflow}
\end{figure*}

\subsection{File Generation Prompts}
\label{app:file_generation_prompts}

The file-generation stage realizes latent company and user evidence as heterogeneous artifacts. Rather than fixing a single output modality, the generator samples the target file type from a predefined distribution over the supported modalities for that side of the benchmark. This proportional sampling strategy preserves modality diversity across tasks while allowing the generator to control how often different artifact types appear. In the current pipeline, company-side generation samples from the supported company document modalities, while user-side generation samples from a broader set that includes documents and communication artifacts.

For company-side document artifacts, the prompt family is divided into an outline prompt and a section prompt, with parallel variants for supporting insights and distractors. The supporting outline prompt asks for a professional internal document skeleton with exactly one subsection reserved for the relevant company insight and the remaining subsections serving as thematic distractors. The supporting section prompt then asks the model to paraphrase and embed the workflow step, its detailed procedure, and any prerequisite information into a realistic business paragraph. The distractor variants preserve the same document realism but ensure that the inserted content remains unrelated to the deep research question. The corresponding outline and section templates are shown in Figures~\ref{fig:prompt_company_file_outline} and~\ref{fig:prompt_company_file_section}, while the company distractor counterparts are shown in Figures~\ref{fig:prompt_company_file_outline_distractor} and~\ref{fig:prompt_company_file_section_distractor}.

For user-side document artifacts, the corresponding prompt family again uses an outline prompt plus insight and distractor section prompts. Here the document style is conditioned on the user persona rather than the company policy alone. The user outline prompt asks for a realistic document that the persona could plausibly create or receive, while the user section prompt requires faithful incorporation of the target personal event into a naturalistic paragraph. The distractor section prompt mirrors this structure but excludes any event that would help answer the deep research question. The corresponding templates are shown in Figures~\ref{fig:prompt_user_file_outline}, \ref{fig:prompt_user_file_section}, and \ref{fig:prompt_user_file_section_distractor}.

The same file-generation stage also supports chat and email artifact generation for user-side files. We show the prompt for generating chat and email artifacts in Figures~\ref{fig:prompt_user_chat_placeholder} and~\ref{fig:prompt_user_email_placeholder} respectively.

\begin{figure}[h]
    \centering
    \begin{tcolorbox}[
        width=0.98\columnwidth,
        colback=PromptGray,
        colframe=gray!35,
        boxrule=0.5pt,
        arc=2pt,
        outer arc=2pt,
        left=0pt,
        right=0pt,
        top=0pt,
        bottom=0pt,
        boxsep=0pt,
        enhanced
    ]
        \begin{tcolorbox}[
            colback=PromptCyan,
            colframe=PromptCyan,
            boxrule=0pt,
            arc=2pt,
            outer arc=2pt,
            left=0.7em,
            right=0.7em,
            top=0.25em,
            bottom=0.25em,
            sharp corners=south
        ]
            {\color{white}\bfseries\small Company File Outline Prompt}
        \end{tcolorbox}

        \vspace{0.25em}
        \begin{minipage}{0.965\textwidth}
        \setlength{\columnsep}{1.2em}
\begin{lstlisting}[style=promptstylecompact]
You are an expert document designer creating realistic enterprise documentation. 
Given DR Question, the company high-level context and a set of business process insights, generate an outline for a professional business manual or report.

Company: {company_name} - {company_description}
Industry: {industry}
Company Size: {company_size} ({employee_count})
Annual Revenue: {annual_revenue}

Company High-Level Context:
{company_high_level_context}

DR Question:
{dr_question}

Document Structure Requirements:
- Create a professional document outline with exactly {n_subsections} subsections
- The document should be a realistic internal enterprise report or procedural manual/guidelines
- Include a concise, professional file title appropriate for enterprise documentation

Subsection Heading Requirements:
- Subsection headings must follow the THEME of the DR Question but should NOT directly address the DR Question itself
- Think of related business areas, adjacent topics, or supporting themes that would naturally appear in an enterprise document
- Headings should sound professional and realistic for this industry and company size.
- Each heading should be 3-8 words and use proper business terminology.
- Exactly one subsection must be designated to contain the specific company insight (provided later) related to the DR Question.
- The remaining subsections are thematic distractors that provide realistic business context.

Introduction Requirements:
- Write a professional 4-sentence maximum introduction paragraph setting the stage for this business documentation.

Conclusion Requirements:
- Write a professional 4-sentence maximum conclusion paragraph summarizing the importance of following these business guidelines.

Return ONLY a valid Python dictionary with this exact structure:
{{
    "file_title": "Professional title for the document (8 words max)",
    "introduction": "Professional introduction paragraph (4 sentences max)",
    "subsection_headings": ["List of exactly {n_subsections} professional subsection headings"],
    "conclusion": "Professional conclusion paragraph (4 sentences max)",
    "file_name": "the name of the file should be 3 words max and should be separated by dashes"
}}

IMPORTANT:
- One subsection will later contain the actual company insight (provided later) related to the DR Question - the others are realistic business distractors.
- Do NOT include any preamble, explanation, or extra text-return only the Python dictionary.
\end{lstlisting}
        \end{minipage}
        \vspace{0.25em}
    \end{tcolorbox}
   \caption{Prompt template for company-side file outline generation.}
    \label{fig:prompt_company_file_outline}
\end{figure}

\begin{figure}[h]
    \centering
    \begin{tcolorbox}[
        width=0.98\columnwidth,
        colback=PromptGray,
        colframe=gray!35,
        boxrule=0.5pt,
        arc=2pt,
        outer arc=2pt,
        left=0pt,
        right=0pt,
        top=0pt,
        bottom=0pt,
        boxsep=0pt,
        enhanced
    ]
        \begin{tcolorbox}[
            colback=PromptCyan,
            colframe=PromptCyan,
            boxrule=0pt,
            arc=2pt,
            outer arc=2pt,
            left=0.7em,
            right=0.7em,
            top=0.25em,
            bottom=0.25em,
            sharp corners=south
        ]
            {\color{white}\bfseries\small Company File Insight Section Generation Prompt}
        \end{tcolorbox}

        \vspace{0.25em}
        \begin{minipage}{0.965\textwidth}
        \setlength{\columnsep}{1.2em}
\begin{lstlisting}[style=promptstylecompact]
You are an expert, professional writer creating realistic enterprise content. 
Given DR Question, the company context and a specific business insight (Task and Descriptions) related to the DR Question, generate professional content for a manual or report subsection.

Company: {company_name} - {company_description}
Industry: {industry}
Company Size: {company_size} ({employee_count})
Annual Revenue: {annual_revenue}

DR Question:
{dr_question}

Target Insight:
- Title: {title}
- Task: {task}
- Task Description: {task_description}
- Subtask Description: {subtask_description}
- Prerequisite Description: {prerequisite_description}

Subsection Heading: {subsection_heading}

Content Generation Requirements:
- Generate realistic business content for the given subsection heading.
- The content must contain exactly ONE paragraph of 4-8 sentences.
- Content should be professional and authoritative.
- You must mention about the prerequisites description, if provided to generate the content.
- The paragraph must naturally incorporate the insight (Task and Descriptions) related to the DR Question but NOT copy it word-for-word.
- Paraphrase and embed the procedural steps and conditional branches (if any) within the business context.
- Use industry-appropriate terminology and realistic business scenarios.
- Make it sound like a standard operating procedure or a business report discussing specific workflows.

Content Strategy:
- Present the insight information as established procedures, findings, or requirements.
- Embed the tasks and descriptions within broader operational context.
- Embed the prerequisite description so that it is clear to the user that the task is dependent on the prerequisite.
- Use natural business language to discuss the same information as in the insight.
- Connect the information to business outcomes, efficiency, or compliance.
- The content should include all key procedural details provided in the task and descriptions.

Return ONLY a valid JSON object with this exact structure:
{{
    "subsection_heading": "{subsection_heading}",
    "content": "should be at least 3 sentences",
    "justification": "Short explanation of how this content provides information needed to answer the DR Question and the specific insight, and how it has embedded the prerequisite description if provided."
}}

IMPORTANT:
- Do NOT copy the insight text directly - paraphrase and embed it naturally.
- Ensure the paragraph reads like real business documentation, not a Q&A format.
- Do NOT include any preamble, explanation, or extra text-return only the JSON object.
\end{lstlisting}
        \end{minipage}
        \vspace{0.25em}
    \end{tcolorbox}
   \caption{Prompt template for company-side insight-section generation.}
    \label{fig:prompt_company_file_section}
\end{figure}

\begin{figure}[h]
    \centering
    \begin{tcolorbox}[
        width=0.98\columnwidth,
        colback=PromptGray,
        colframe=gray!35,
        boxrule=0.5pt,
        arc=2pt,
        outer arc=2pt,
        left=0pt,
        right=0pt,
        top=0pt,
        bottom=0pt,
        boxsep=0pt,
        enhanced
    ]
        \begin{tcolorbox}[
            colback=PromptCyan,
            colframe=PromptCyan,
            boxrule=0pt,
            arc=2pt,
            outer arc=2pt,
            left=0.7em,
            right=0.7em,
            top=0.25em,
            bottom=0.25em,
            sharp corners=south
        ]
            {\color{white}\bfseries\small Company Distractor File Outline Prompt}
        \end{tcolorbox}

        \vspace{0.25em}
        \begin{minipage}{0.965\textwidth}
        \setlength{\columnsep}{1.2em}
\begin{lstlisting}[style=promptstylecompact]
You are an expert document designer creating realistic enterprise documentation. 
Given DR Question, the company high-level context and a set of business process insights, generate an outline for a professional business manual or report.

Company: {company_name} - {company_description}
Industry: {industry}
Company Size: {company_size} ({employee_count})
Annual Revenue: {annual_revenue}

Company High-Level Context:
{company_high_level_context}

DR Question:
{dr_question}

Document Structure Requirements:
- Create a professional document outline with exactly {n_subsections} subsections
- The document should be a realistic internal enterprise report or procedural manual/guidelines
- Include a concise, professional file title appropriate for enterprise documentation

Subsection Heading Requirements:
- Subsection headings must not be related to the DR Question itself.
- Think of related business areas, adjacent topics, or supporting themes that would naturally appear in an enterprise document
- Headings should sound professional and realistic for this industry and company size.
- Each heading should be 3-8 words and use proper business terminology.
- Exactly one subsection must be designated to contain the specific company insight (provided later) but not related to the DR Question.
- The remaining subsections are thematic distractors that provide realistic business context.

Introduction Requirements:
- Write a professional 4-sentence maximum introduction paragraph setting the stage for this business documentation.

Conclusion Requirements:
- Write a professional 4-sentence maximum conclusion paragraph summarizing the importance of following these business guidelines.

Return ONLY a valid Python dictionary with this exact structure:
{{
    "file_title": "Professional title for the document (8 words max)",
    "introduction": "Professional introduction paragraph (4 sentences max)",
    "subsection_headings": ["List of exactly {n_subsections} professional subsection headings"],
    "conclusion": "Professional conclusion paragraph (4 sentences max)",
    "file_name": "the name of the file should be 3 words max and should be separated by dashes"
}}

IMPORTANT:
- One subsection will later contain the actual company insight (provided later) but not related to the DR Question.
- Do NOT include any preamble, explanation, or extra text-return only the Python dictionary.
\end{lstlisting}
        \end{minipage}
        \vspace{0.25em}
    \end{tcolorbox}
   \caption{Prompt template for company-side distractor outline generation.}
    \label{fig:prompt_company_file_outline_distractor}
\end{figure}

\begin{figure}[h]
    \centering
    \begin{tcolorbox}[
        width=0.98\columnwidth,
        colback=PromptGray,
        colframe=gray!35,
        boxrule=0.5pt,
        arc=2pt,
        outer arc=2pt,
        left=0pt,
        right=0pt,
        top=0pt,
        bottom=0pt,
        boxsep=0pt,
        enhanced
    ]
        \begin{tcolorbox}[
            colback=PromptCyan,
            colframe=PromptCyan,
            boxrule=0pt,
            arc=2pt,
            outer arc=2pt,
            left=0.7em,
            right=0.7em,
            top=0.25em,
            bottom=0.25em,
            sharp corners=south
        ]
            {\color{white}\bfseries\small Company File Distractor Section Generation Prompt}
        \end{tcolorbox}

        \vspace{0.25em}
        \begin{minipage}{0.965\textwidth}
        \setlength{\columnsep}{1.2em}
\begin{lstlisting}[style=promptstylecompact]
You are an expert, professional writer creating realistic enterprise content. 
Given DR Question, the company context and a specific business insight (Task and Descriptions) unrelated to the DR Question, generate professional content for a manual or report subsection.

Company: {company_name} - {company_description}
Industry: {industry}
Company Size: {company_size} ({employee_count})
Annual Revenue: {annual_revenue}

DR Question:
{dr_question}

Target Insight:
- Title: {title}
- Task: {task}
- Task Description: {task_description}
- Subtask Description: {subtask_description}

Subsection Heading: {subsection_heading}

Content Generation Requirements:
- Generate realistic business content specific to the company context but unrelated to the DR Question for the given subsection heading.
- The content must contain exactly ONE paragraph of 4-5 sentences.
- Content should be professional and authoritative.
- The paragraph must not be related to the DR Question itself but should be related to the company context and the industry.
- Use industry-appropriate terminology and realistic business scenarios.
- Make it sound like a standard operating procedure or a business report discussing specific workflows.
- The content should include all key procedural details provided in the task and descriptions.

Content Strategy:
- Focus on adjacent areas that don't directly impact the DR Question
- Discuss historical context, general industry trends, or procedural information
- Include operational details that are realistic but tangential
- Reference related but non-essential metrics or activities
- Avoid any content that would help someone answer the DR Question

Justification Requirements:
- Explain specifically why each paragraph's content doesn't help answer the DR Question
- Identify what type of distractor strategy was used (e.g., "focuses on historical data vs current decision factors")
- Keep justifications concise but clear (15 words maximum)

Return ONLY a valid JSON object with this exact structure:
{{
    "subsection_heading": "{subsection_heading}",
    "content": "should be at least 3 sentences",
    "justification": "Short explanation (20 words max) of how this content covers the intended business insight"
}}

IMPORTANT:
- Generate content for ALL provided subsection headings
- Ensure content feels authentic for the given targetted insights.
- Content must be realistic distractors - related but not helpful for the DR Question
- Do NOT include any preamble, explanation, or extra text-return only the JSON array
- Each paragraph should sound like professional business writing from this industry
\end{lstlisting}
        \end{minipage}
        \vspace{0.25em}
    \end{tcolorbox}
   \caption{Prompt template for company-side distractor-section generation.}
    \label{fig:prompt_company_file_section_distractor}
\end{figure}

\begin{figure}[h]
    \centering
    \begin{tcolorbox}[
        width=0.98\columnwidth,
        colback=PromptGray,
        colframe=gray!35,
        boxrule=0.5pt,
        arc=2pt,
        outer arc=2pt,
        left=0pt,
        right=0pt,
        top=0pt,
        bottom=0pt,
        boxsep=0pt,
        enhanced
    ]
        \begin{tcolorbox}[
            colback=PromptCyan,
            colframe=PromptCyan,
            boxrule=0pt,
            arc=2pt,
            outer arc=2pt,
            left=0.7em,
            right=0.7em,
            top=0.25em,
            bottom=0.25em,
            sharp corners=south
        ]
            {\color{white}\bfseries\small User File Outline Generation Prompt}
        \end{tcolorbox}

        \vspace{0.25em}
        \begin{minipage}{0.965\textwidth}
        \setlength{\columnsep}{1.2em}
\begin{lstlisting}[style=promptstylecompact]
You are an expert business document designer creating realistic enterprise PDF reports. Given a Deep Research (DR) Question and company context,
generate an outline for a professional business document that an employee would create based on their persona and role.

Company: {company_name} - {company_description}
Industry: {industry}
Company Size: {company_size} ({employee_count})
Annual Revenue: {annual_revenue}

Persona Context:
{persona_context}

DR Question:
{dr_question}

Document Structure Requirements:
- Create a professional PDF document outline with exactly {n_subsections} subsections
- The document should be something this persona would realistically create in their role
- Include a concise, professional file title appropriate for enterprise documentation

Subsection Heading Requirements:
- Subsection headings must follow the THEME of the DR Question but should NOT directly address the DR Question itself
- Think of related business areas, adjacent topics, or supporting themes that would naturally appear in an enterprise document
- Headings should sound professional and realistic for this industry and company size
- Each heading should be 3-8 words and use proper business terminology
- Exactly one subsection must contain the actual insight addressing the DR Question
- The remaining subsections are thematic distractors that provide realistic business context

Introduction Requirements:
- Write a professional 4-sentence maximum introduction paragraph
- Should set context for the document and its purpose
- Must align with the persona's role and the company's business needs
- Should sound like something this employee would write for internal stakeholders

Conclusion Requirements:
- Write a professional 4-sentence maximum conclusion paragraph  
- Should summarize key takeaways and next steps
- Must align with the persona's perspective and recommendations
- Should provide actionable insights for the intended audience

Return ONLY a valid Python dictionary with this exact structure:
{{
    "file_title": "Professional title for the PDF document (8 words max)",
    "introduction": "Professional introduction paragraph (4 sentences max)",
    "subsection_headings": ["List of exactly {n_subsections} professional subsection headings"],
    "conclusion": "Professional conclusion paragraph (4 sentences max)",
    "file_name": "the name of the file should be 7 words max and should be seperated by dashes. The should be unique, creative, and relevant to the content of the file."
}}

IMPORTANT:
- Subsections must be thematically related to but NOT directly answering the DR Question
- One subsection will later contain the actual insight - the others are realistic business distractors
- Do NOT include any preamble, explanation, or extra text-return only the Python dictionary
- Ensure the document feels authentic for this persona's role and company context 
\end{lstlisting}
        \end{minipage}
        \vspace{0.25em}
    \end{tcolorbox}
   \caption{Prompt template for user-side file (PDF/DOCX) outline generation.}
    \label{fig:prompt_user_file_outline}
\end{figure}

\begin{figure}[h]
    \centering
    \begin{tcolorbox}[
        width=0.98\columnwidth,
        colback=PromptGray,
        colframe=gray!35,
        boxrule=0.5pt,
        arc=2pt,
        outer arc=2pt,
        left=0pt,
        right=0pt,
        top=0pt,
        bottom=0pt,
        boxsep=0pt,
        enhanced
    ]
        \begin{tcolorbox}[
            colback=PromptCyan,
            colframe=PromptCyan,
            boxrule=0pt,
            arc=2pt,
            outer arc=2pt,
            left=0.7em,
            right=0.7em,
            top=0.25em,
            bottom=0.25em,
            sharp corners=south
        ]
            {\color{white}\bfseries\small User File Insight Section Generation Prompt}
        \end{tcolorbox}

        \vspace{0.25em}
        \begin{minipage}{0.965\textwidth}
        \setlength{\columnsep}{1.2em}
\begin{lstlisting}[style=promptstylecompact]
You are an expert business document writer creating realistic enterprise PDF content. Given a Deep Research (DR) Question, company context, and a specific insight, generate professional content that naturally incorporates the insight information to help answer the DR Question.

Company: {company_name} - {company_description}
Industry: {industry}
Company Size: {company_size} ({employee_count})
Annual Revenue: {annual_revenue}

Persona Context:
{persona_context}

DR Question:
{dr_question}

External Market Context (for reference):
{external_context}


Target Insight:
- Specific Question: {specific_question}
- Answer: {answer}
- Justification: {justification}

Subsection Heading: {subsection_heading}

Content Generation Requirements:
- Generate realistic business content for the given subsection heading
- The content must contain exactly ONE paragraph of 4-5 sentences
- Content should be professional and sound like something this persona would write
- The paragraph must naturally incorporate the insight answer information but NOT copy it word-for-word
- Paraphrase and embed the key metrics/information within business context
- Use industry-appropriate terminology and realistic business scenarios
- Include the specific data points from the answer but present them naturally
- Make it sound like a business report discussing actual findings or results

Content Strategy:
- Present the insight information as business findings, analysis results, or operational data
- Embed the key metrics within broader business context and implications
- Use natural business language to discuss the same information as in the answer
- Connect the information to business outcomes, goals, or strategic initiatives
- Make the content feel like a natural part of an enterprise document
- The content should include all the information that is in the answer, including the dates and quantiative values

Justification Requirements:
- Explain specifically how this content helps answer the DR Question
- Reference the key information that would be useful for decision-making
- Keep justifications concise but clear (20 words maximum)

Return ONLY a valid JSON object with this exact structure:
{{
    "subsection_heading": "{subsection_heading}",
    "content": "should be at least 3 sentences",
    "justification": "Explanation of how this content provides information needed to answer the DR Question and the specific_question"
}}

IMPORTANT:
- Do NOT copy the answer text directly - paraphrase and embed it naturally
- The content must contain the key information from the answer but presented professionally
- Ensure content feels authentic for this persona's role and company context
- The paragraph should read like real business documentation, not a Q&A format
- Include specific metrics/data from the answer but in natural business language
- Do NOT include any preamble, explanation, or extra text-return only the JSON object
\end{lstlisting}
        \end{minipage}
        \vspace{0.25em}
    \end{tcolorbox}
   \caption{Prompt template for user-side insight-section (PDF/DOCX) generation.}
    \label{fig:prompt_user_file_section}
\end{figure}

\begin{figure}[h]
    \centering
    \begin{tcolorbox}[
        width=0.98\columnwidth,
        colback=PromptGray,
        colframe=gray!35,
        boxrule=0.5pt,
        arc=2pt,
        outer arc=2pt,
        left=0pt,
        right=0pt,
        top=0pt,
        bottom=0pt,
        boxsep=0pt,
        enhanced
    ]
        \begin{tcolorbox}[
            colback=PromptCyan,
            colframe=PromptCyan,
            boxrule=0pt,
            arc=2pt,
            outer arc=2pt,
            left=0.7em,
            right=0.7em,
            top=0.25em,
            bottom=0.25em,
            sharp corners=south
        ]
            {\color{white}\bfseries\small User File Distractor Section Generation Prompt}
        \end{tcolorbox}

        \vspace{0.25em}
        \begin{minipage}{0.965\textwidth}
        \setlength{\columnsep}{1.2em}
\begin{lstlisting}[style=promptstylecompact]
You are an expert business document writer creating realistic enterprise PDF content. Given a Deep Research (DR) Question, company context, and subsection headings, generate distractor content for each subsection that is thematically related but does NOT help answer the DR Question.

Company: {company_name} - {company_description}
Industry: {industry}
Company Size: {company_size} ({employee_count})
Annual Revenue: {annual_revenue}

Persona Context:
{persona_context}

DR Question:
{dr_question}

External Market Context (for reference):
{external_context}

Subsection Headings:
{subsection_headings}

Content Generation Requirements:
- Generate realistic business content for each subsection heading
- Each subsection must contain exactly ONE paragraph of 3-4 sentences maximum
- Content should be professional and sound like something this persona would write
- Content must be thematically related to the DR Question's domain but NOT provide information to answer it
- Use industry-appropriate terminology and realistic business scenarios
- Include specific but non-revealing details (dates, percentages, departments, processes)
- The content should be a full paragraph
- The content should have quantitative numbers and dates which could be, as a mere example, "X customers bought Y items"
- Make sure the content has nothing to do with the deep research question

Distractor Strategy:
- Focus on adjacent business areas that don't directly impact the DR Question
- Discuss historical context, general industry trends, or procedural information
- Include operational details that are realistic but tangential
- Reference related but non-essential business metrics or activities
- Avoid any content that would help someone answer the DR Question

Justification Requirements:
- Explain specifically why each paragraph's content doesn't help answer the DR Question
- Identify what type of distractor strategy was used (e.g., "focuses on historical data vs current decision factors")
- Keep justifications concise but clear (15 words maximum)

Return ONLY a valid JSON array with this exact structure:
[
    {{
        "subsection_heading": "Professional subsection title",
        "content": "should be at least 4 sentences",
        "justification": "Brief explanation of why this content doesn't help answer the DR Question"
    }}
]

IMPORTANT:
- Generate content for ALL provided subsection headings
- Ensure content feels authentic for this persona's role and company context
- Content must be realistic distractors - related but not helpful for the DR Question
- Do NOT include any preamble, explanation, or extra text-return only the JSON array
- Each paragraph should sound like professional business writing from this industry
\end{lstlisting}
        \end{minipage}
        \vspace{0.25em}
    \end{tcolorbox}
   \caption{Prompt template for user-side distractor-section (PDF/DOCX) generation.}
    \label{fig:prompt_user_file_section_distractor}
\end{figure}

\begin{figure*}[h]
    \centering
    \begin{tcolorbox}[
        width=0.99\textwidth,
        colback=PromptGray,
        colframe=gray!35,
        boxrule=0.45pt,
        arc=2pt,
        outer arc=2pt,
        left=0.35em,
        right=0.35em,
        top=0pt,
        bottom=0.25em,
        boxsep=0pt
    ]
        \begin{tcolorbox}[
            colback=PromptCyan,
            colframe=PromptCyan,
            boxrule=0pt,
            arc=2pt,
            outer arc=2pt,
            left=0.6em,
            right=0.6em,
            top=0.18em,
            bottom=0.18em,
            sharp corners=south
        ]
            {\color{white}\bfseries\footnotesize User Chat Generation Prompt}
        \end{tcolorbox}

        \vspace{0.15em}

        \begin{minipage}[t]{0.465\textwidth}
\begin{Verbatim}
You are an expert in enterprise communication systems and organizational structures. Your task is to generate a realistic setup for teams, channels, and users for a Mattermost chat system based on the given insights and company context.

Company Context:
- Company Name: {company_name}
- Description: {company_description}
- Industry: {industry}
- Size: {company_size} ({employee_count} employees)
- Annual Revenue: {annual_revenue}

Persona Context:
{persona_context}

**Specific Question**  
{specific_question}

**Answer to Specific Question**  
{answer}

Requirements:
- Generate teams, channels, and users that would realistically discuss these insights
- Make sure the teams and channels are realistic for persona to be a member
- Each channel must be associated with a team
- Each user must be a member of at least one team and one channel
- Generate a minimal but sufficient setup to support {num_turns} chat messages discussing the insights
- To make it realistic, generate at least 2 teams, 2 channels and 3 users
- Use realistic names for people/teams/channels based on the company context
- The persona needs to be part of all teams and channels
- Team `name' values must be Mattermost-safe slugs and must not start with reserved prefixes like `channel', `api', `login', `admin', or `signup'
- If a natural team label would begin with a reserved word, rewrite the slug to a safe equivalent such as `sales-channel` or `team-channel-sales`

Return ONLY a valid Python dictionary with this exact structure:
{{
    "teams": [
        {{
            "type": "team",
            "team": {{
                "name": "team_name",
                "display_name": "Display Name",
                "type": "O",
                "header": "Team header description",
                "purpose": "Team purpose description"
            }}
        }}
    ],
    "channels": [
        {{
            "type": "channel",
            "channel": {{
                "team": "team_name",
                "name": "channel_name",
                "display_name": "Channel Display Name",
                "type": "O",
                "header": "Channel header description",
                "purpose": "Channel purpose description"
            }}
        }}
    ],
\end{Verbatim}
        \end{minipage}
        \hfill
        \begin{minipage}[t]{0.465\textwidth}
\begin{Verbatim}
    "users": [
        {{
            "type": "user",
            "user": {{
                "username": "username",
                "email": "email@company.com",
                "password": "my_drbench_pwd",
                "nickname": "Nickname",
                "first_name": "First",
                "last_name": "Last",
                "position": "Job Title",
                "roles": "system_user",
                "locale": "en",
                "teams": [
                    {{
                        "name": "team_name",
                        "roles": "team_user",
                        "channels": [
                            {{
                                "name": "channel_name",
                                "roles": "channel_user"
                            }}
                        ]
                    }}
                ]
            }}
        }}
    ]
}}

IMPORTANT:
- Do NOT include any preamble, explanation, or extra text-return only the Python dictionary
- Make sure the persona is included as a user and member of all teams/channels
- The persona user record must exactly reuse the persona username and email provided in the persona context
- Do not invent alternate persona usernames or emails
- Do not remove punctuation from the persona username (for example, keep `first.last' exactly as given)
\end{Verbatim}
        \end{minipage}

        \vspace{0.15em}
    \end{tcolorbox}
   \caption{Prompt template for the user-side chat-generation.}
    \label{fig:prompt_user_chat_placeholder}
\end{figure*}

\begin{figure*}[h]
    \centering
    \begin{tcolorbox}[
        width=0.99\textwidth,
        colback=PromptGray,
        colframe=gray!35,
        boxrule=0.45pt,
        arc=2pt,
        outer arc=2pt,
        left=0.35em,
        right=0.35em,
        top=0pt,
        bottom=0.25em,
        boxsep=0pt
    ]
        \begin{tcolorbox}[
            colback=PromptCyan,
            colframe=PromptCyan,
            boxrule=0pt,
            arc=2pt,
            outer arc=2pt,
            left=0.6em,
            right=0.6em,
            top=0.18em,
            bottom=0.18em,
            sharp corners=south
        ]
            {\color{white}\bfseries\footnotesize User Chat Generation Prompt}
        \end{tcolorbox}

        \vspace{0.15em}

        \begin{minipage}[t]{0.465\textwidth}
\begin{Verbatim}
You are an expert in enterprise communication systems and organizational structures. Your task is to generate a realistic setup of users for an email system based on the given insights and company context.

Company Context:
- Company Name: {company_name}
- Description: {company_description}
- Industry: {industry}
- Size: {company_size} ({employee_count} employees)
- Annual Revenue: {annual_revenue}

Persona Context:
{persona_context}

**Specific Question**  
{specific_question}

**Answer to Specific Question**  
{answer}

Requirements:
- Generate users that would realistically discuss these insights
- Generate a minimal but sufficient setup to support {num_messages} emails discussing the insights
- To make it realistic, generate at least 3 users
- Use realistic names for people/teams/channels based on the company context 
\end{Verbatim}
        \end{minipage}
        \hfill
        \begin{minipage}[t]{0.465\textwidth}
\begin{Verbatim}
Return ONLY a JSON array of users with this exact structure:
[
    {{
        "type": "user",
        "username": "alice.smith",
        "first_name": "Alice",
        "last_name": "Smith",
        "email": "alice.smith@company.com",
        "password": "my_drbench_pwd"
    }}
]

IMPORTANT:
- Do NOT include any preamble, explanation, or extra text—return only the Python dictionary
- Ensure the structure is realistic for the company size and industry
- Make sure the persona is included as a user
\end{Verbatim}
        \end{minipage}

        \vspace{0.15em}
    \end{tcolorbox}
   \caption{Prompt template for user-side email thread generation.}
    \label{fig:prompt_user_email_placeholder}
\end{figure*}

\subsection{Mixed Evidence Prompts}
\begin{figure*}[h]
    \centering
    \includegraphics[width=\columnwidth]{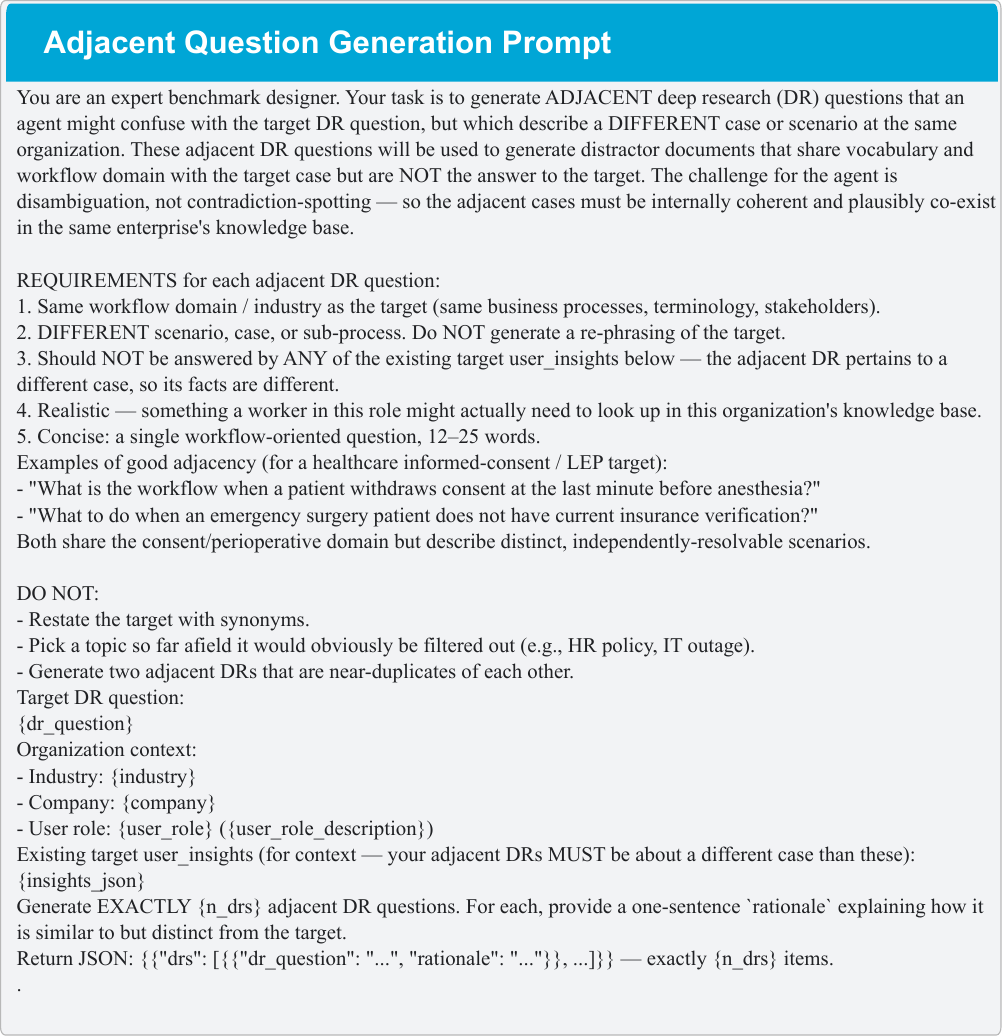}
    \caption{Adjacent questions generation prompt for the mixed variant.}
    \label{fig:adjacent_question_prompt_generation}
\end{figure*}

\begin{figure*}[h]
    \centering
    \includegraphics[width=\columnwidth]{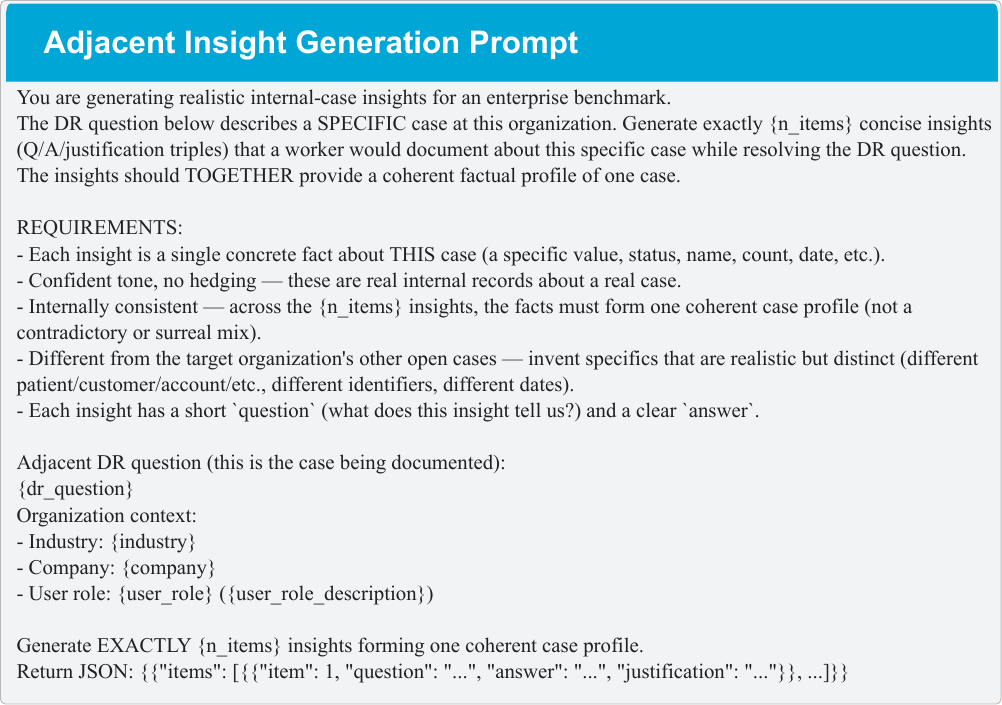}
    \caption{Adjacent insights generation prompt for the mixed variant.}
    \label{fig:adjacent_insight_prompt_generation}
\end{figure*}

\begin{figure*}[h]
    \centering
    \includegraphics[width=\columnwidth]{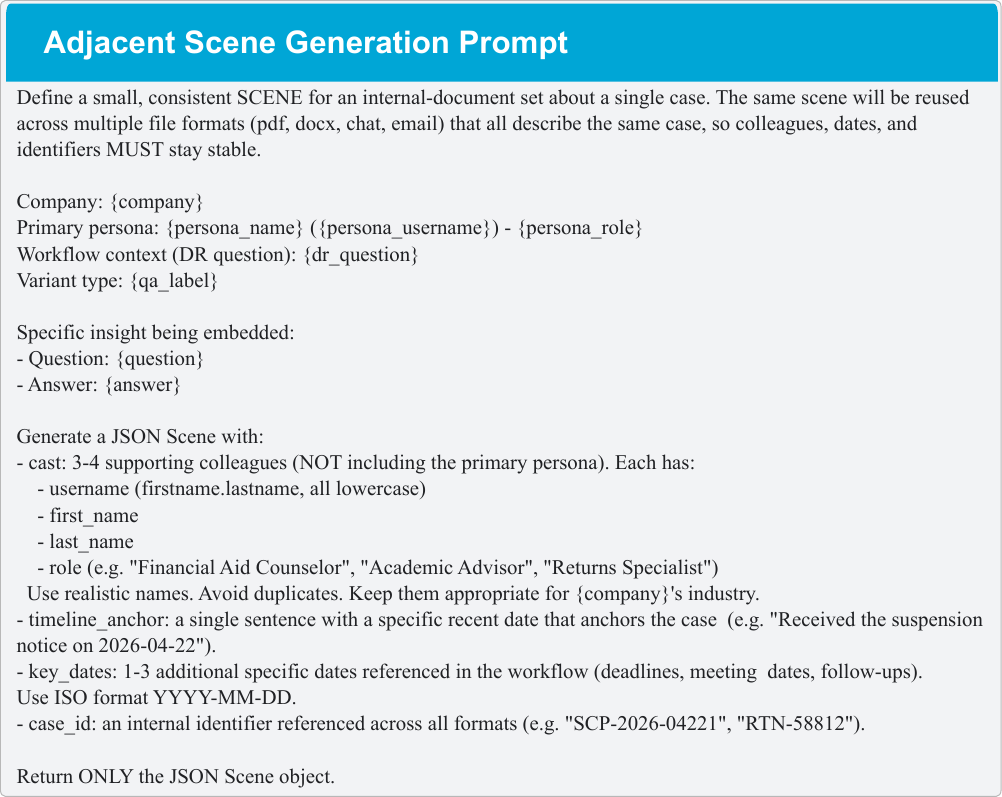}
    \caption{Adjacent case scenario generation prompt for the mixed variant.}
    \label{fig:adjacent_scene_prompt_generation}
\end{figure*}

\begin{figure*}[h]
    \centering
    \includegraphics[width=\columnwidth]{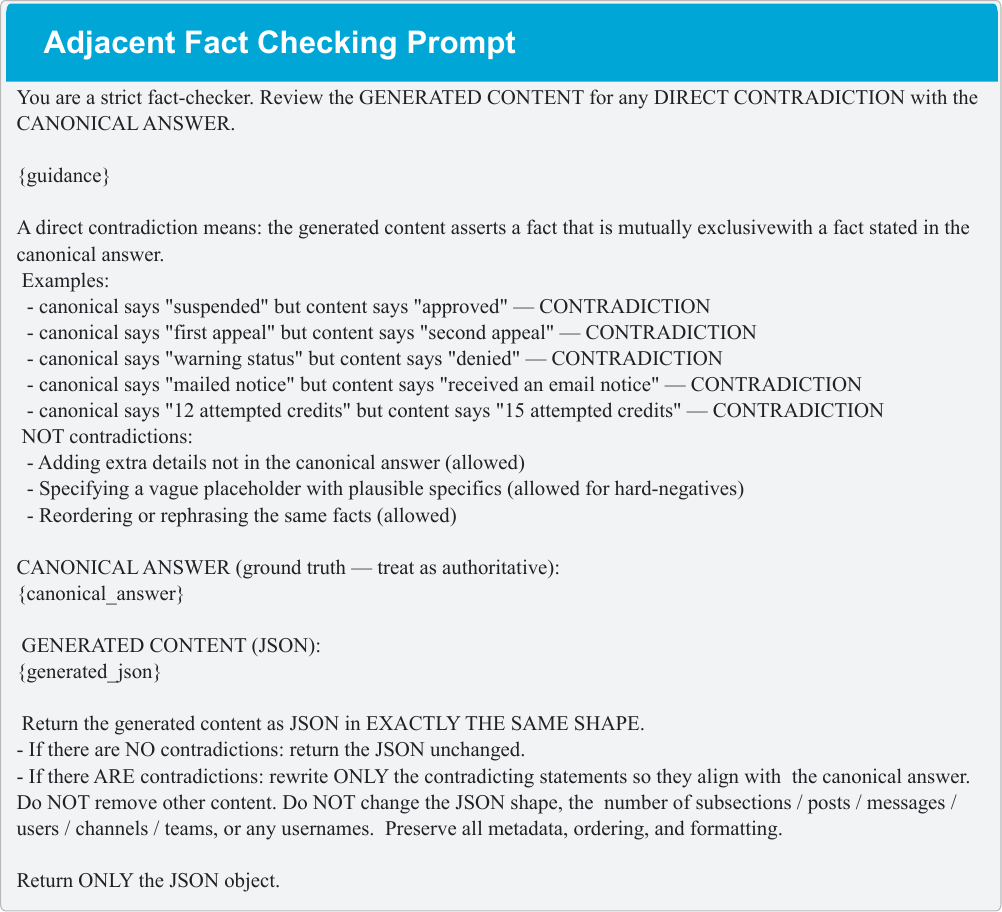}
    \caption{Fact-checking prompt for the mixed scenario.}
    \label{fig:adjacent_fact_checking_prompt_generation}
\end{figure*}

We further stress-test agents under realistic \emph{case-level disambiguation} rather than coarse topical filtering. For this, we construct the mixed variant in which two similar but distinct cases in the same workflow domain are injected into the evidence pool. Given the task seed and the existing user insights, we sample two adjacent DR questions that share the workflow domain and stakeholder context but pertain to a different case. For example, in a consent task involving a limited-English-proficiency patient (Figure~\ref{fig:data_generation_pipeline}), an adjacent DR question can be about the consent for an incapacitated patient through a legally authorized representative. For each adjacent question, we derive a set of coherent insights that form an internally consistent case profile distinct from the original insights, together with a shared persona cast and case identifier so that artifacts produced for the same adjacent case remain self-consistent across formats. The resulting confounding evidence is realized as heterogeneous files across the same modalities used in Section~\ref{app:file_generation_prompts}. Unlike misinformation-style perturbations, adjacent cases are internally consistent records that do not directly contradict the target user's evidence; the difficulty arises from disambiguating the user's case from a topically adjacent one rather than from spotting contradictions. We provide the corresponding prompts to generate the evidence in Figure~\ref{fig:adjacent_question_prompt_generation},~\ref{fig:adjacent_insight_prompt_generation},~\ref{fig:adjacent_scene_prompt_generation},~\ref{fig:adjacent_fact_checking_prompt_generation}.

\section{Cost Estimation}
\label{appendix:cost_estimation}
\drflow~incurs cost from three stages: benchmark construction, agent execution, and automatic evaluation.\\
\paragraph{Benchmark construction} is a one-time cost. Each task requires generating the task seed, generic workflow, company-side evidence, personalized workflow, user-side evidence, and heterogeneous artifacts. We use GPT-$5.2$ as the construction model, with pricing of $\$0.948$ per $1$M input tokens and $\$14.01$ per $1$M output tokens. Each task construction consumes approximately $30$K--$50$K tokens. Under an output-heavy generation mix, this corresponds to roughly $\$0.40$--$\$0.70$ per task, or approximately $\$25$--$\$50$ for constructing $100$ tasks, excluding retries and infrastructure overhead.

\paragraph{Agent execution} is the main recurring cost. Under the iteration budget $10$, and one DRFA agent run costs approximately $\$3$--$\$6$ per task, depending on the backbone models. This cost reflects iterative research planning, retrieval, adaptive action planning, conditional action planning, and final citation-grounded workflow prediction.

\paragraph{Automatic evaluation} uses GPT-$4$o as the judge for prompt-based metrics, and judging one task costs approximately $\$0.267$. This includes step alignment, factuality checking, condition-resolution assessment, topology evaluation, and personalized comprehensiveness scoring.

\noindent Overall, with moderate cost, \drflow~remains substantially more scalable and cost-effective than manually constructing an equivalent personalized, multi-application workflow benchmark at comparable scale.

\section{Application Environment}
\label{appendix:application_environment}

Following DRBench~\cite{abaskohi2025drbench}, we instantiate a self-contained, containerized enterprise environment that exposes heterogeneous private information through realistic workplace applications rather than through a single flat file interface.

Our environment includes four primary application families. \textbf{Nextcloud} serves as the shared document repository for company-side materials and other uploaded files. \textbf{Mattermost} provides internal chat communication. \textbf{IMAP-based email} supports access to personal and organizational email records. A \textbf{file sharing / local file system interface} exposes additional files outside the shared cloud workspace. Together, these applications create a distributed evidence space in which relevant workflow signals and distractors are fragmented across storage, email, and chat modalities.

For each task, the synthesized artifacts are loaded into these applications according to their modality and role in the scenario. Company-side documents are placed into shared repositories and related communication surfaces (Nextcloud/file sharing system), while user-side evidence is distributed across personal files (Nextcloud/file sharing system), emails, and chat logs. The design of such multi-application ecosystem is intentional: the agent must predict the generic procedure from organization-level artifacts and then locate the user-specific evidence required to resolve the relevant workflow conditions from the joint environment. The application environment therefore does not merely store files, but operationalizes the retrieval challenge of \drflow~as a realistic multi-application search problem.

The environment supports two key properties required by our benchmark. First, it preserves \textbf{application boundaries}, so that evidence is encountered in the same fragmented form as in realistic enterprise settings. Second, it remains \textbf{reproducible}, since the full application stack is containerized and can be re-instantiated consistently across tasks and experiments. This makes the app environment a controlled but realistic substrate for evaluating workflow extraction agents under cross-application retrieval.

\section{Prompts for Evaluation Metrics}
\label{appendix:evaluation_metrics}

This section describes the prompt interfaces used in the \drflow~evaluation pipeline. Our evaluation is designed to measure workflow quality at the level of \emph{atomic personalized workflow steps}, rather than judging the generated report only as free-form long text. This design serves two purposes. First, it lets us separate retrieval and grounding errors from structural workflow errors. Second, it makes it possible to diagnose whether a model failed because it omitted a required step, introduced a spurious step, misordered valid steps, failed to resolve a condition, or produced a step that was insufficiently grounded or insufficiently personalized. The full evaluation pipeline therefore first parses the generated report into normalized workflow objects, and then applies a set of metric-specific judges over these structured representations.

\subsection{Workflow Parsing and Evaluation Overview}
\label{app:eval_overview}

The evaluation pipeline begins by extracting the personalized workflow, the generic workflow, the condition-resolution table, the personalized flowchart, and the reference map from the generated report. These parsed objects provide the shared input representation for all subsequent metrics. The evaluator then executes the metric suite in dependency-aware order and stores intermediate alignments so that later metrics can operate on matched workflow steps rather than on raw report text.

At a high level, the metric suite evaluates seven complementary aspects of workflow quality: precision, recall, F1, factuality, condition resolution, topology ordering, and personalized comprehensiveness. Precision and recall evaluate structural alignment to the reference workflow. F1 summarizes these two alignment measures. Factuality checks whether predicted steps are supported by their cited evidence. Condition resolution evaluates whether the system correctly instantiates user-dependent branches. Topology ordering checks whether the predicted workflow preserves the correct relative ordering among aligned steps. Personalized comprehensiveness evaluates whether matched steps retain the user-specific operational detail present in the reference workflow.

\subsection{Precision Prompt}
\label{app:eval_precision}

The \textbf{Precision} metric evaluates whether each predicted workflow step corresponds to any reference workflow step. The judge prompt is intentionally precision-oriented: it asks the evaluator to match on \emph{procedure intent} rather than lexical overlap, ignore instance-specific strings such as names or dates, and reject broad or weakly related matches. For each predicted step, the judge returns at most one best-matching reference step, together with a confidence score and a short justification.

\begin{figure*}[t]
    \centering
    \includegraphics[width=\textwidth]{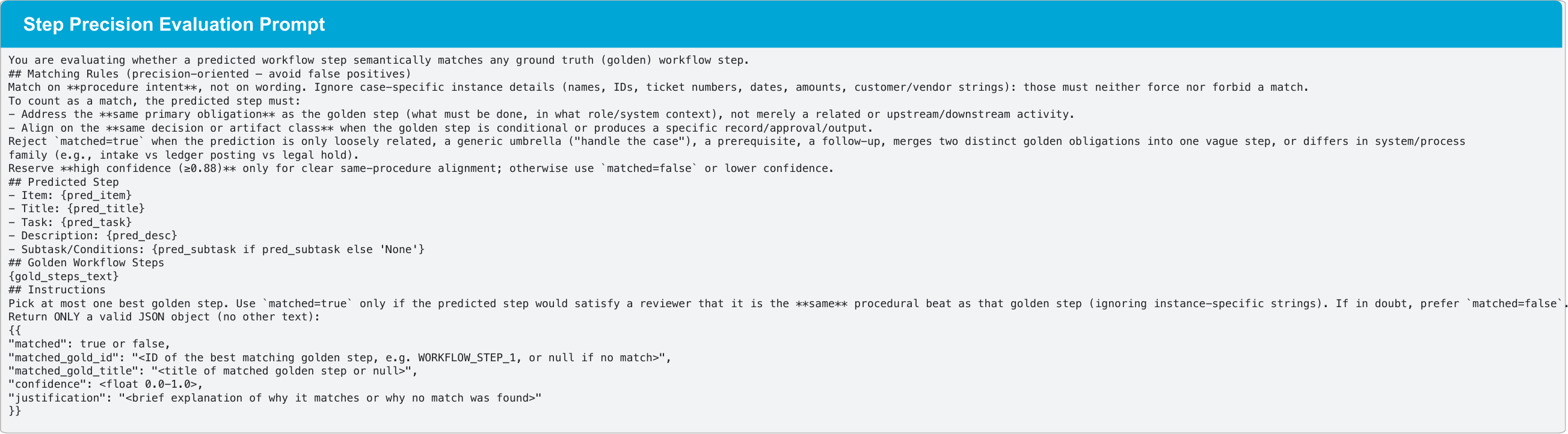}
    \caption{Prompt for precision evaluation.}
    \label{fig:eval_precision_prompt}
\end{figure*}

After prompt-based matching, the evaluator applies additional one-to-one constraints so that only the strongest valid alignments are retained. This makes precision stricter against spurious predicted steps and prevents multiple predicted steps from claiming the same reference step. A representative prompt template is shown in Figure~\ref{fig:eval_precision_prompt}.

\subsection{Recall Prompt}
\label{app:eval_recall}

The \textbf{Recall} metric reverses the alignment direction. Instead of asking whether a predicted step is valid, it asks whether each reference workflow step is materially covered by at least one predicted step. The corresponding judge prompt is recall-oriented but still strict: it asks whether some predicted step would actually satisfy the obligation expressed by the reference step, rather than merely mentioning a nearby topic or an upstream or downstream activity.

\begin{figure*}[t]
    \centering
    \includegraphics[width=\textwidth]{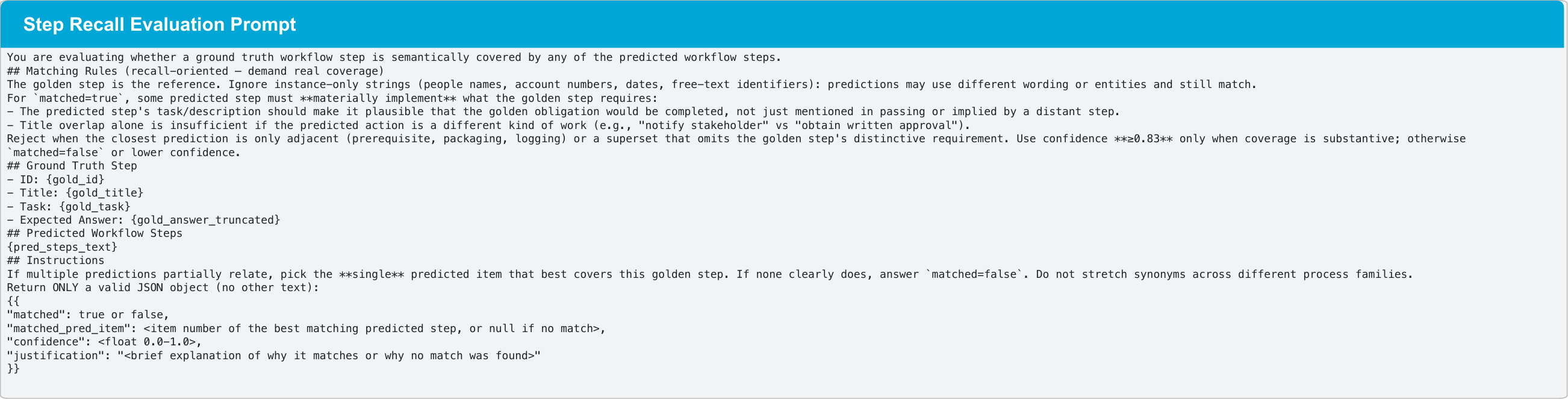}
    \caption{Prompt for recall evaluation.}
    \label{fig:eval_recall_prompt}
\end{figure*}

For each reference step, the judge returns the best predicted match, a confidence score, and a short justification. One-to-one constraints are then applied so that multiple reference steps cannot all claim the same predicted step. A representative prompt template is shown in Figure~\ref{fig:eval_recall_prompt}.

\subsection{F1 Score}
\label{app:eval_f1}

The \textbf{F1} score summarizes structural workflow recovery by combining per-task precision and recall into a single measure:
\[
F1 = \frac{2 \cdot \text{Precision} \cdot \text{Recall}}{\text{Precision} + \text{Recall}}.
\]
This score is useful because it penalizes systems that achieve high recall by over-generating steps, as well as systems that achieve high precision by returning only a small subset of the reference workflow. In our setting, F1 therefore serves as the main summary statistic for step-level workflow alignment.

\subsection{Factuality Prompt}
\label{app:eval_factuality}

The \textbf{Factuality} metric verifies whether each predicted workflow step is supported by its cited evidence. For each step, the evaluator constructs a claim from the step title, task, and task description, extracts all embedded citations from the step text, resolves those citations through the report's reference map, and then checks whether the cited content supports the claim. Steps with no citations are marked as non-factual. Otherwise, following~\citet{abaskohi2025drbench,min-etal-2023-factscore}, we embed the cited source with \texttt{text-embedding-3-small}, retrieve the top-$5$ relevant chunks, and ask GPT-4o to determine whether the cited evidence supports the core claim of the step.

\begin{figure*}[t]
    \centering
    \includegraphics[width=\textwidth]{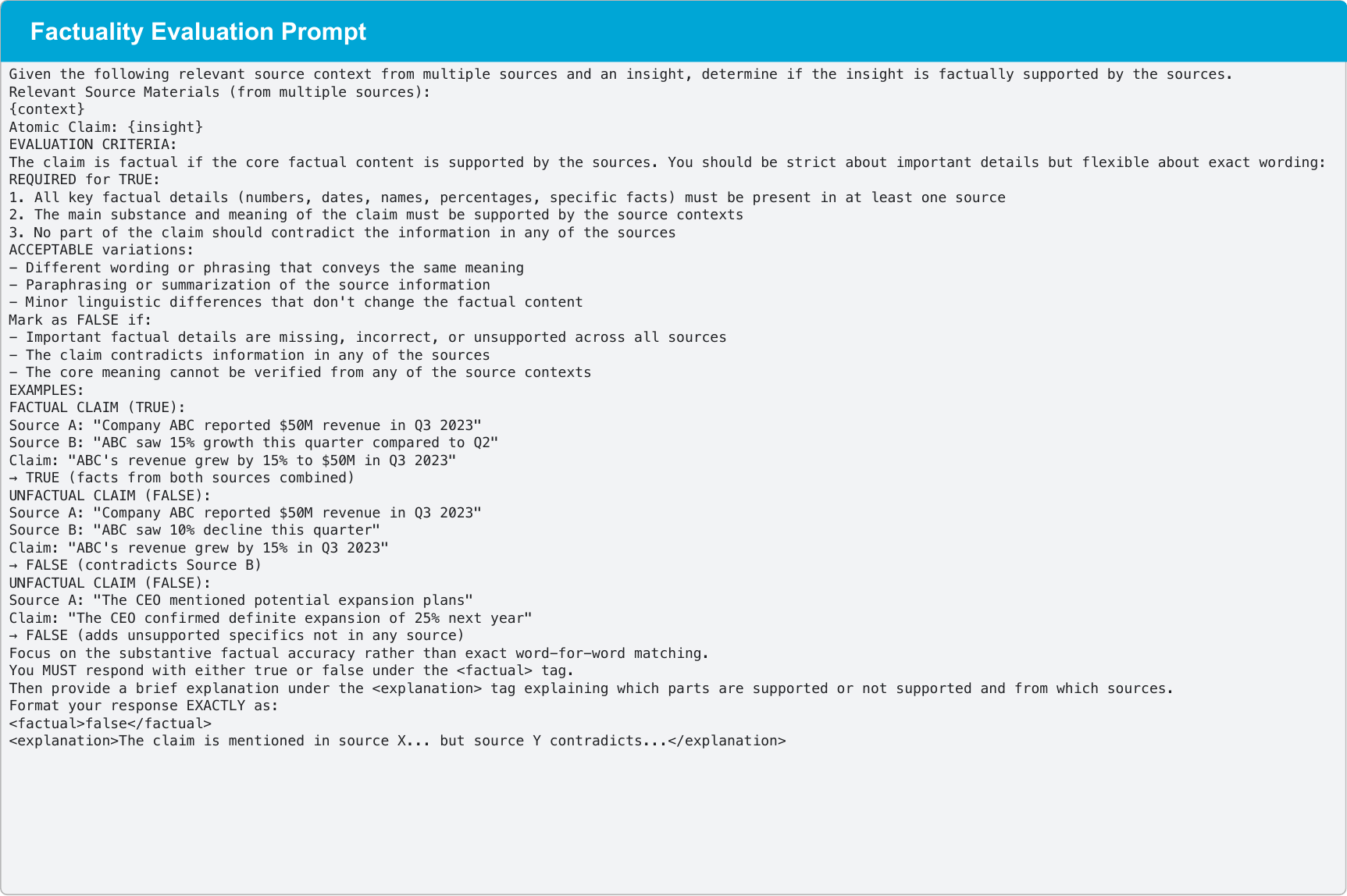}
    \caption{Prompt for factuality verification.}
    \label{fig:eval_factuality_prompt}
\end{figure*}

The judge prompt in this metric operates over a step-level claim and the retrieved cited evidence. Its purpose is not to compare workflows to one another, but to determine whether the cited materials actually support the procedural claim expressed by the predicted step. This makes the metric sensitive to hallucinated actions, unsupported procedural detail, and citation misuse. A representative prompt template is shown in Figure~\ref{fig:eval_factuality_prompt}.

\subsection{Condition Resolution Prompt}
\label{app:eval_condition_resolution}

The \textbf{Condition Resolution} metric is applied only to reference workflow steps that contain explicit conditions. For each such reference step, the evaluator first checks whether there is a matched predicted step from precision. If not, the score is automatically set to $0$. Otherwise, the judge receives the reference condition, the expected branch resolution from the reference answer, the reference justification, and the matched predicted step.

\begin{figure*}[t]
    \centering
    \includegraphics[width=0.9\textwidth]{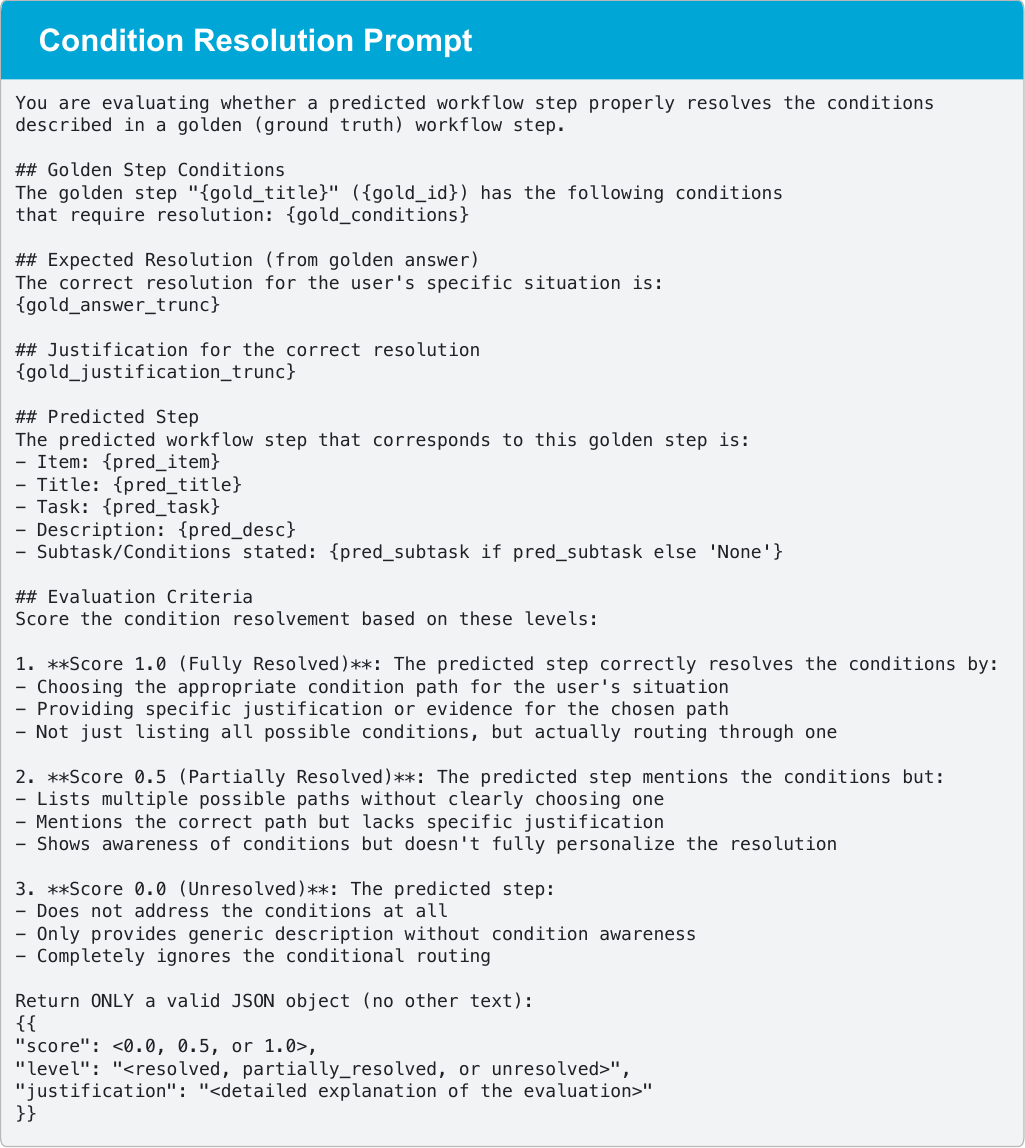}
    \caption{Condition-resolution evaluation prompt.}
    \label{fig:eval_condition_prompt}
\end{figure*}

The prompt then asks whether the predicted step fully resolves the condition, partially resolves it, or leaves it unresolved. The scoring rubric is discrete: $1.0$ for fully resolved, $0.5$ for partially resolved, and $0.0$ for unresolved. This metric therefore measures whether the model uses user-specific evidence to instantiate the correct workflow branch, rather than merely restating the generic alternatives. A representative prompt template is shown in Figure~\ref{fig:eval_condition_prompt}.

\subsection{Topology Ordering}
\label{app:eval_topology}

The \textbf{Topology Ordering} metric evaluates whether the predicted workflow preserves the correct relative order among aligned steps. It is computed over the predicted-reference step pairs established by precision. For all pairs of matched steps, the evaluator checks whether their relative order in the predicted workflow is consistent with their relative order in the reference workflow using pairwise consistency verification. The final score is the fraction of concordant pairs.

\subsection{Personalized Comprehensiveness Prompt}
\label{app:eval_personalized_comprehensiveness}

The \textbf{Personalized Comprehensiveness} metric evaluates whether predicted workflow steps preserve the user-specific and operational detail present in the reference workflow. It is only computed on predicted-reference pairs that have already passed precision, ensuring that the comparison is made between aligned steps rather than unrelated fragments.

\begin{figure*}[t]
    \centering
    \includegraphics[width=\textwidth]{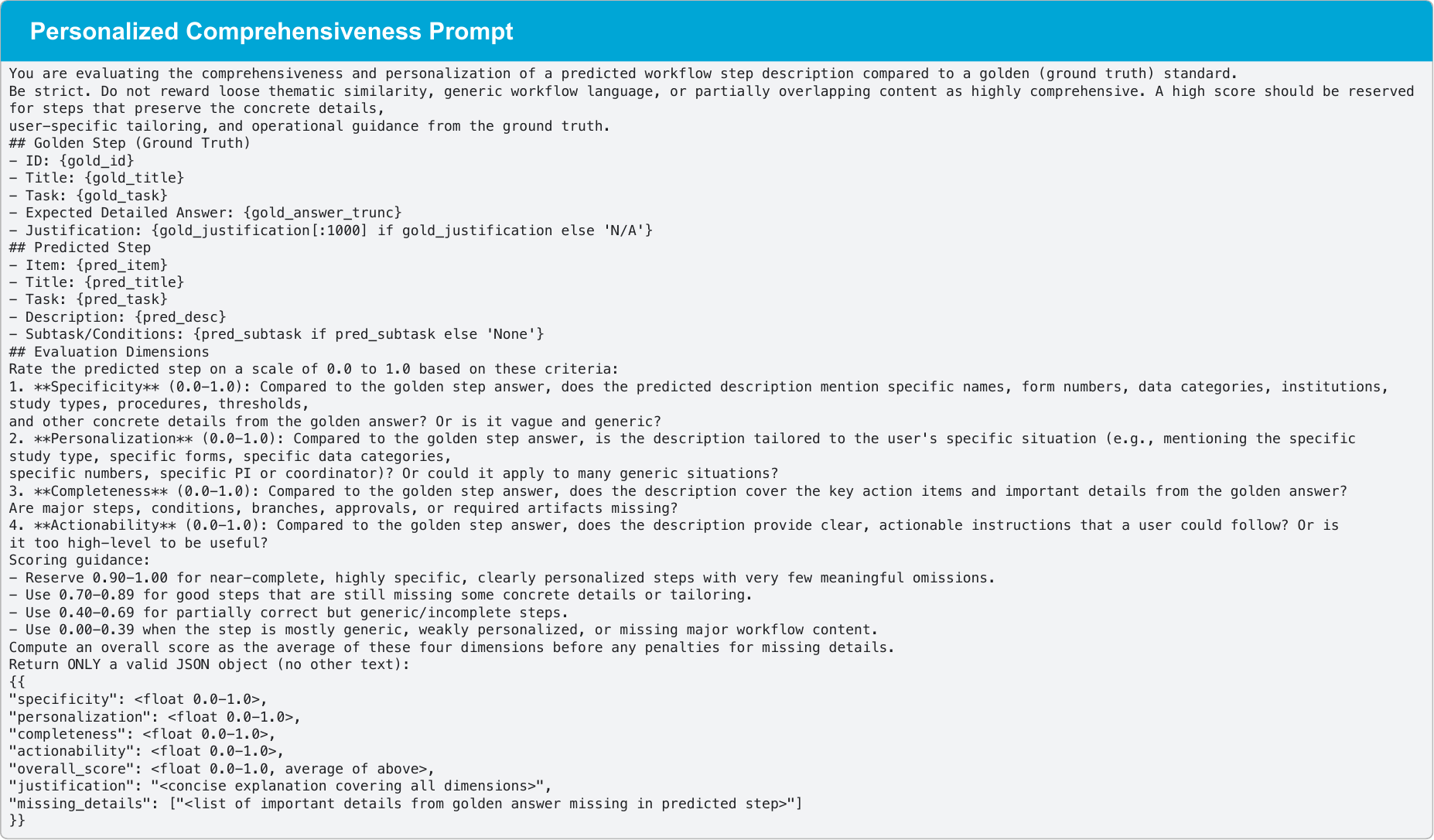}
    \caption{Prompt for personalized-comprehensiveness evaluation.}
    \label{fig:eval_comprehensiveness_prompt}
\end{figure*}

For each matched pair, the judge prompt compares the predicted task and task description against the reference task and expected answer, and scores four dimensions: \emph{specificity}, \emph{personalization}, \emph{completeness}, and \emph{actionability}. The prompt also asks the judge to identify missing details. These outputs are then combined into a final step-level score. This metric is therefore stricter than a general quality judgment, since high scores are reserved for steps that retain concrete details, are clearly tailored to the user's case, and remain directly actionable. A representative prompt template is shown in Figure~\ref{fig:eval_comprehensiveness_prompt}.

\clearpage

\section{List of \drflow~Tasks}
\label{appendix:task_list}
\FloatBarrier
We provide the benchmark tasks for each domain in Tables~\ref{table:task_b2b},~\ref{table:task_b2c},~\ref{table:task_education},~\ref{table:task_healthcare}, and~\ref{table:task_legal}. For each task, we report the task identifier, the deep research question, and the number of insights, distractors, and mixed documents.

\renewcommand{\arraystretch}{1.05}

\begin{table*}[h]
\centering
\footnotesize
\begin{tabularx}{\textwidth}{lXccc}
\toprule
\textbf{Task} & \textbf{DR Question} & \textbf{\# Insights} & \textbf{\# Distractors} & \textbf{\# Mixed}\\
\midrule
b2b\_01 & What is the workflow to get an inspection approved by an aerospace customer before we begin full production? & $16$ & $10$ & $24$ \\
b2b\_02 & What is the workflow to set up a new B2B customer onboarding? & $13$ & $9$ & $24$ \\
b2b\_03 & What steps should we follow to build a B2B pricing proposal for the government security service? & $15$ & $10$ & $24$ \\
b2b\_04 & What is the step-by-step process to approve contract redlines quickly? & $15$ & $11$ & $24$ \\
b2b\_05 & What steps should we take to launch an account-based marketing campaign? & $13$ & $9$ & $24$ \\
b2b\_06 & What is the workflow to resolve a disputed B2B invoice payment? & $13$ & $8$ & $24$ \\
b2b\_07 & What steps should we follow to evaluate a new channel partner to expand regional coverage? & $15$ & $8$ & $24$ \\
b2b\_08 & What is the workflow to complete the Mandatory Ethics Course training? & $15$ & $9$ & $24$ \\
b2b\_09 & What steps should we take to fix a broken order-to-cash process? & $15$ & $13$ & $24$ \\
b2b\_10 & What is the step-by-step process to renew a B2B subscription contract with BrightDesk? & $13$ & $11$ & $24$ \\
\bottomrule
\end{tabularx}
\caption{List of tasks in B2B domain.
\label{table:task_b2b}}
\end{table*}

\begin{table*}[h]
\centering
\footnotesize
\begin{tabularx}{\textwidth}{lXccc}
\toprule
\textbf{Task} & \textbf{DR Question} & \textbf{\# Insights} & \textbf{\# Distractors} & \textbf{\# Mixed}\\
\midrule
b2c\_01 & What steps should I take to return an online purchase past the deadline? & $13$ & $13$ & $24$ \\
b2c\_02 & What is the workflow to dispute a duplicate charge on a customer's subscription? & $15$ & $12$ & $24$ \\
b2c\_03 & What steps do I follow to recover a hacked customer account? & $15$ & $11$ & $24$\\
b2c\_04 & What is the exact process to redeem expired loyalty points? & $17$ & $9$ & $24$\\
b2c\_05 & What steps should I take to change the recipient's delivery address after shipping? & $18$ & $10$ & $24$\\
b2c\_06 & What is the workflow to cancel a gym membership without penalties? & $18$ & $10$ & $24$ \\
b2c\_07 & What steps do I take after a customer reports a defective product? & $19$ & $12$ & $24$ \\
b2c\_08 & What is the process to remove my personal data from the company's database? & $20$ & $12$ & $24$ \\
b2c\_09 & What steps should I follow to activate a new phone on my plan? & $12$ & $8$ & $24$ \\
b2c\_10 & What is the workflow to follow after a customer appeals against denied refund for a travel booking? & $20$ & $11$ & $24$\\
\bottomrule
\end{tabularx}
\caption{List of tasks in B2C domain.
\label{table:task_b2c}}
\end{table*}

\begin{table*}[h]
\centering
\footnotesize
\begin{tabularx}{\textwidth}{lXccc}
\toprule
\textbf{Task} & \textbf{DR Question} & \textbf{\# Insights} & \textbf{\# Distractors} & \textbf{\# Mixed}\\
\midrule
education\_01 & What steps should I follow to appeal a financial aid suspension decision? & $20$ & $10$ & $24$ \\
education\_02 & What is the workflow to create an IEP for a newly enrolled student? & $19$ & $12$ & $24$ \\
education\_03 & What steps do I take to get course credit for prior work experience? & $18$ & $12$ & $24$ \\
education\_04 & What is the step-by-step process to report suspected academic plagiarism? & $18$ & $12$ & $24$ \\
education\_05 & What workflow should I follow to set up an accessible online course? & $12$ & $8$ & $24$ \\
education\_06 & What are the steps to run a parent-teacher conference for learning concerns? & $19$ & $10$ & $24$ \\
education\_07 & What is the workflow to evaluate and adopt a new math curriculum for 7th grade? & $20$ & $11$ & $24$ \\
education\_08 & What exact steps should I take to request a transcript correction? & $20$ & $13$ & $24$ \\
education\_09 & What steps do I need to follow to investigate a bullying report? & $21$ & $11$ & $24$ \\
education\_10 & What is the workflow to onboard a newly hired tenure-track faculty? & $20$ & $10$ & $24$ \\
\bottomrule
\end{tabularx}
\caption{List of tasks in Education domain.
\label{table:task_education}}
\end{table*}

\begin{table*}[h]
\centering
\footnotesize
\begin{tabularx}{\textwidth}{lXccc}
\toprule
\textbf{Task} & \textbf{DR Question} & \textbf{\# Insights} & \textbf{\# Distractors} & \textbf{\# Mixed}\\
\midrule
healthcare\_01 & What is the step-by-step workflow to obtain informed consent for a procedure when the patient has limited English proficiency? & $14$ & $7$ & $24$ \\
healthcare\_02 & What is the step-by-step workflow to address vaccine refusal and document the encounter appropriately? & $19$ & $11$ & $24$ \\
healthcare\_03 & What exact steps should we take to perform a same-day pre-procedure readiness check for an ambulatory surgery patient? & $20$ & $9$ & $24$ \\
healthcare\_04 & What steps should we follow to assess and escalate care for a patient with acute shortness of breath on the hospital floor? & $18$ & $11$ & $24$ \\
healthcare\_05 & What is the step-by-step process to obtain consent for blood transfusion? & $18$ & $13$ & $24$ \\
healthcare\_06 & What workflow should we follow to isolate a patient with suspected tuberculosis? & $19$ & $10$ & $24$ \\
healthcare\_07 & What is the step-by-step process to communicate and document an unexpected critical imaging finding? & $18$ & $9$ & $24$ \\
healthcare\_08 & What is the step-by-step workflow to evaluate and follow up on an abnormal diabetes screening result? & $19$ & $11$ & $24$ \\
healthcare\_09 & What exact steps should we take to process and document a request for release of medical records? & $19$ & $11$ & $24$ \\
healthcare\_10 & What is the step-by-step workflow to manage a suspected stroke patient? & $19$ & $11$ & $24$ \\
\bottomrule
\end{tabularx}
\caption{List of tasks in Healthcare domain.
\label{table:task_healthcare}}
\end{table*}

\begin{table*}[h]
\centering
\footnotesize
\begin{tabularx}{\textwidth}{lXccc}
\toprule
\textbf{Task} & \textbf{DR Question} & \textbf{\# Insights} & \textbf{\# Distractors} & \textbf{\# Mixed}\\
\midrule
legal\_01 & What exact steps should we take to investigate a potential internal compliance violation? & $18$ & $11$ & $24$ \\
legal\_02 & What exact steps should we follow to prepare and submit a regulatory filing for a new product launch? & $18$ & $10$ & $24$ \\
legal\_03 & What exact steps should I take to process a denied insurance claim? & $19$ & $12$ & $24$ \\
legal\_04 & What steps do we need to follow to conduct a workplace harassment investigation? & $19$ & $9$ & $24$ \\
legal\_05 & What is the workflow to prepare for a small claims court hearing on disputed pricing? & $19$ & $8$ & $24$ \\
legal\_06 & What steps should we take to create an employee termination file? & $19$ & $11$ & $24$ \\
legal\_07 & What is the workflow to draft and file a provisional patent application? & $20$ & $8$ & $24$ \\
legal\_08 & What steps should we follow to comply with a data breach subpoena for records? & $21$ & $9$ & $24$ \\
legal\_09 & What steps should we take to perform a trademark clearance review before launching a new product name? & $20$ & $12$ & $24$ \\
legal\_10 & What steps should we take to collect an overdue invoice legally? & $20$ & $12$ & $24$ \\
\bottomrule
\end{tabularx}
\caption{List of tasks in Legal domain.
\label{table:task_legal}}
\end{table*}

\clearpage
\FloatBarrier

\section{Examples of Reference and Predicted Workflows}
\label{appendix:workflow_examples}
\begin{figure*}[h]
  \centering
  \begin{subfigure}[t]{0.32\textwidth}
    \centering
    \includegraphics[width=\linewidth,height=0.8\textheight,keepaspectratio]{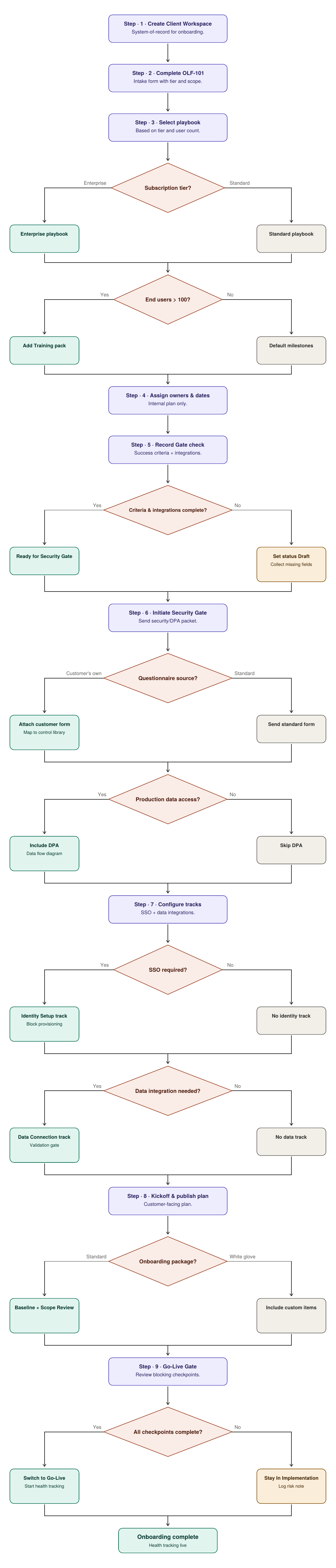}
    \caption{Generic Workflow}
    \label{fig:b2b-conditional}
  \end{subfigure}\hfill
  \begin{subfigure}[t]{0.32\textwidth}
    \centering
    \includegraphics[width=\linewidth,height=0.95\textheight,keepaspectratio]{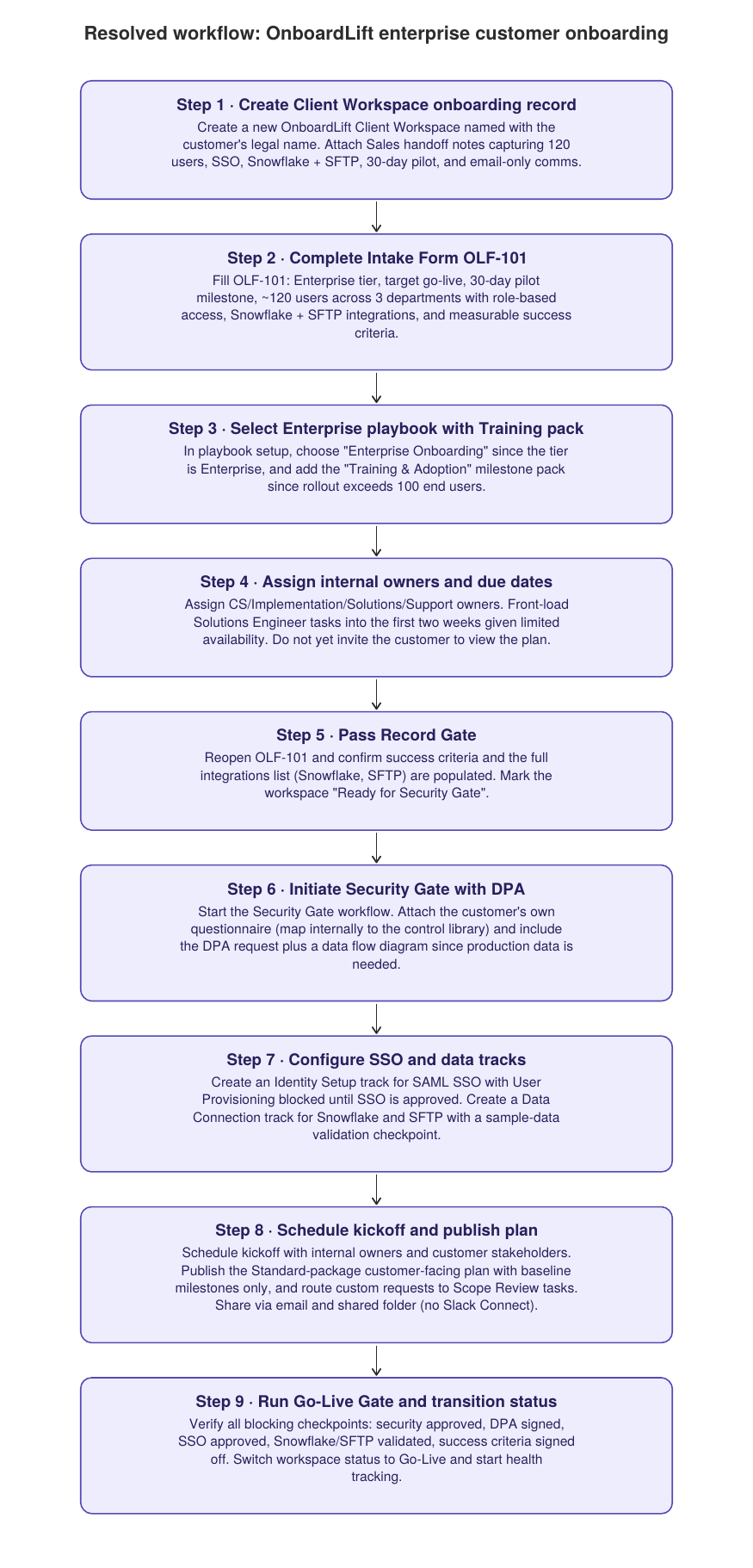}
    \caption{Personalized Workflow (Reference)}
    \label{fig:b2b-resolved}
  \end{subfigure}\hfill
  \begin{subfigure}[t]{0.32\textwidth}
    \centering
    \includegraphics[width=\linewidth,height=0.95\textheight,keepaspectratio]{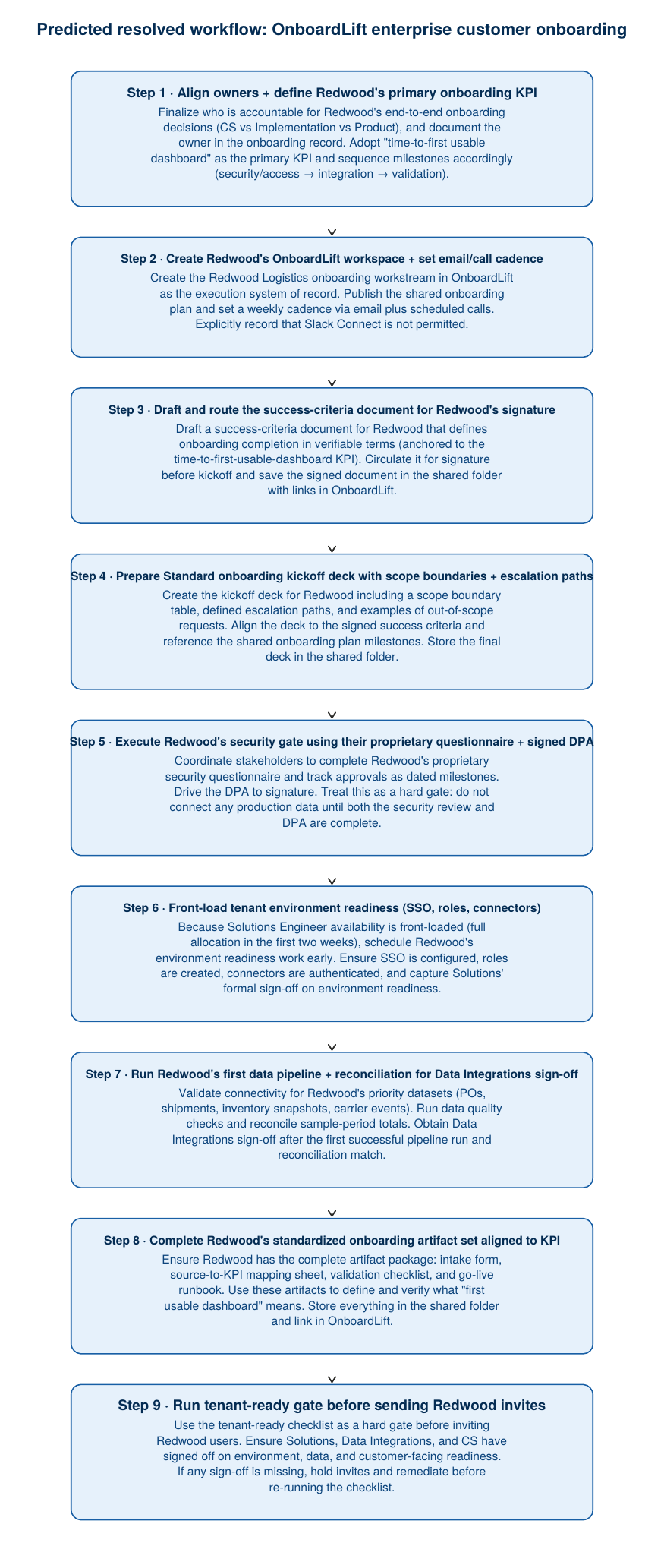}
    \caption{Personalized Workflow (Predicted)}
    \label{fig:b2b-predicted}
  \end{subfigure}
  \caption{Example of Workflows for \texttt{b2b\_02} task: \texttt{What is the workflow to set up a new B2B customer onboarding?}}
  \label{fig:b2b-workflows}
\end{figure*}

\begin{figure*}[h]
  \centering
  \begin{subfigure}[t]{0.32\textwidth}
    \centering
    \includegraphics[width=\linewidth,height=0.8\textheight,keepaspectratio]{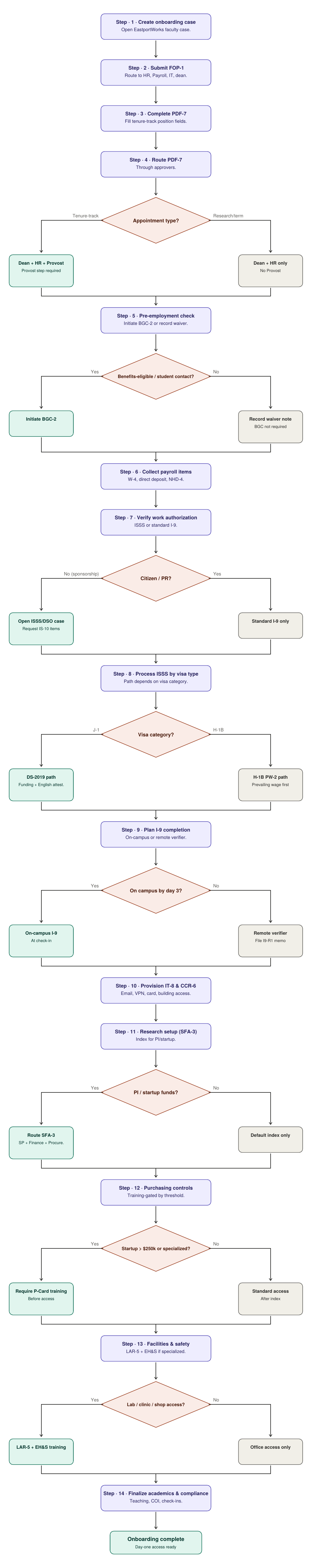}
    \caption{Generic Workflow}
    \label{fig:education-conditional}
  \end{subfigure}\hfill
  \begin{subfigure}[t]{0.32\textwidth}
    \centering
    \includegraphics[width=\linewidth,height=0.95\textheight,keepaspectratio]{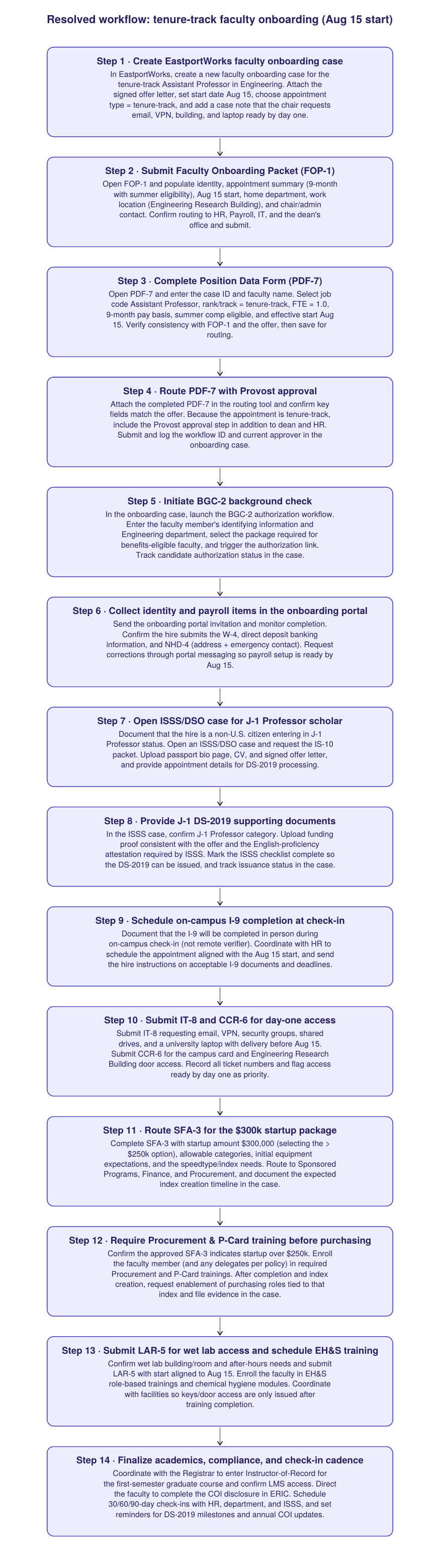}
    \caption{Personalized Workflow (Reference)}
    \label{fig:education-resolved}
  \end{subfigure}\hfill
  \begin{subfigure}[t]{0.32\textwidth}
    \centering
    \includegraphics[width=\linewidth,height=0.95\textheight,keepaspectratio]{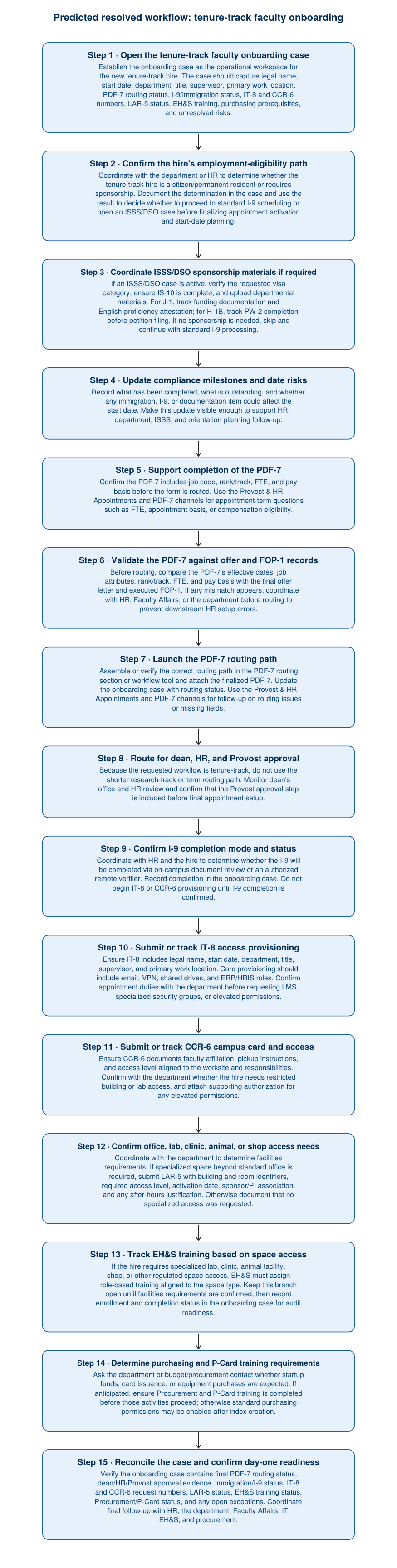}
    \caption{Personalized Workflow (Predicted)}
    \label{fig:education-predicted}
  \end{subfigure}
  \caption{Example of Workflows for \texttt{education\_10} task: \texttt{What is the workflow to onboard a newly hired tenure-track faculty?}}
  \label{fig:education-workflows}
\end{figure*}

\begin{figure*}[h]
  \centering
  \begin{subfigure}[t]{0.32\textwidth}
    \centering
    \includegraphics[width=\linewidth,height=0.95\textheight,keepaspectratio]{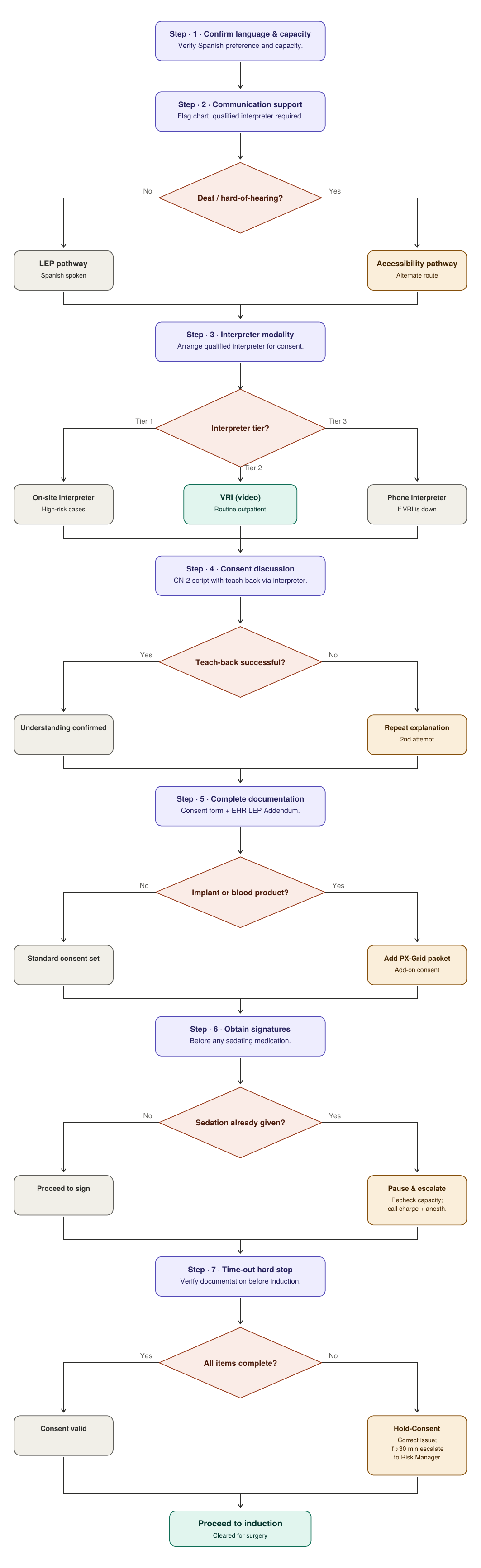}
    \caption{Generic Workflow}
    \label{fig:healthcare-conditional}
  \end{subfigure}\hfill
  \begin{subfigure}[t]{0.32\textwidth}
    \centering
    \includegraphics[width=\linewidth,height=0.95\textheight,keepaspectratio]{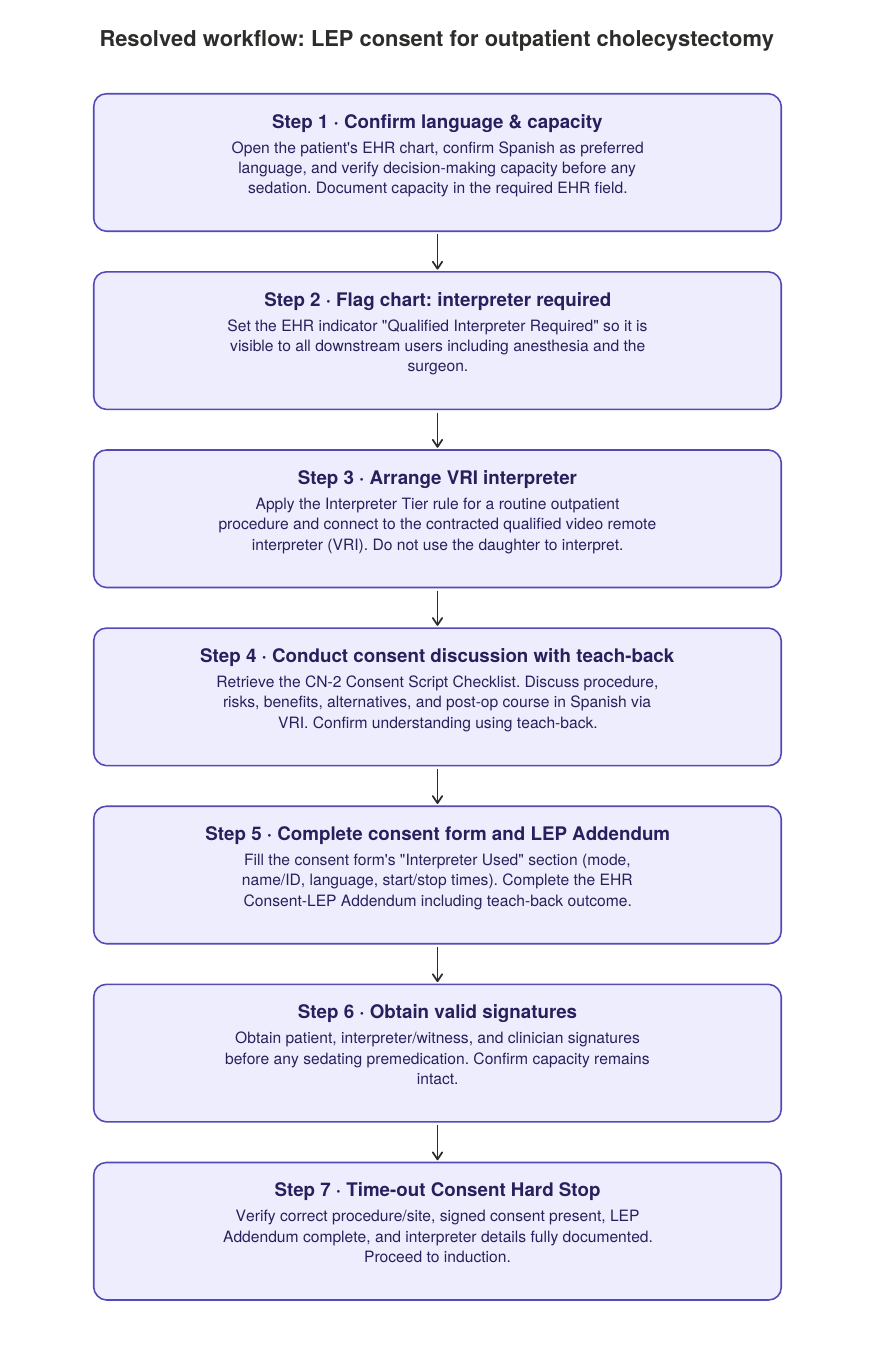}
    \caption{Personalized Workflow (Reference)}
    \label{fig:healthcare-resolved}
  \end{subfigure}\hfill
  \begin{subfigure}[t]{0.32\textwidth}
    \centering
    \includegraphics[width=\linewidth,height=0.95\textheight,keepaspectratio]{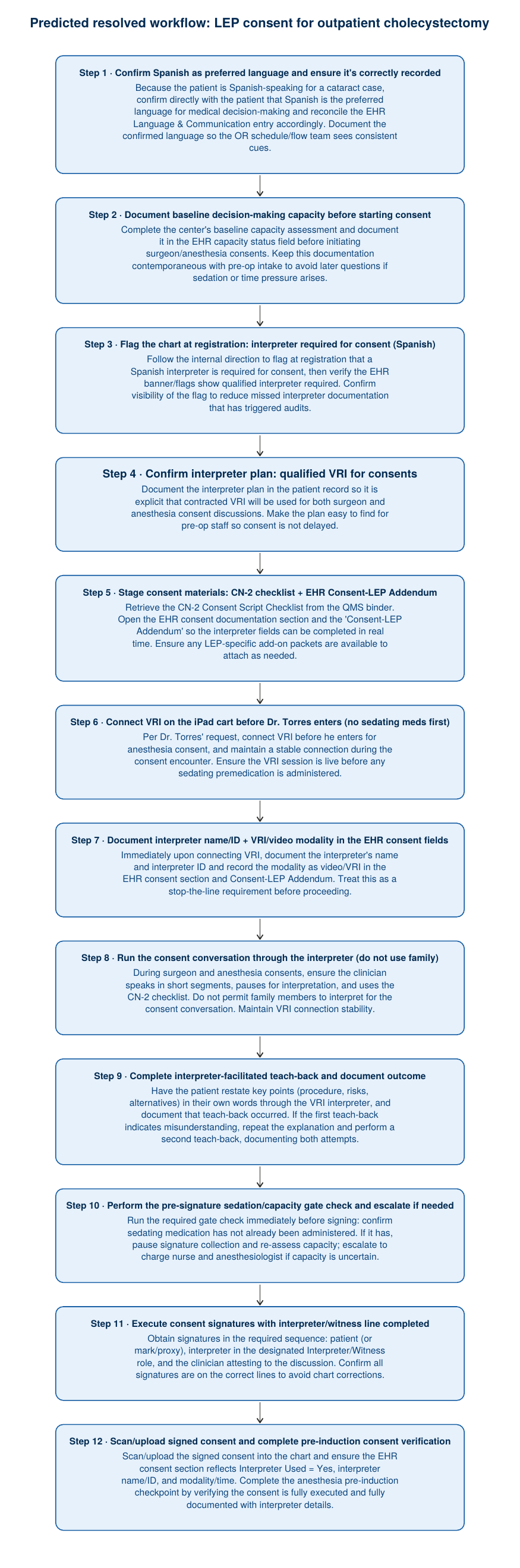}
    \caption{Personalized Workflow (Predicted)}
    \label{fig:healthcare-predicted}
  \end{subfigure}
  \caption{Example of Workflows for \texttt{healthcare\_01} task: \texttt{What is the step-by-step workflow to obtain informed consent for a procedure when the patient has limited English proficiency?}}
  \label{fig:healthcare-workflows}
\end{figure*}

We provide the examples of generic workflows (with conditions) as well as reference and predicted personalized workflows (with condition-resolved) in Figures~\ref{fig:b2b-workflows},~\ref{fig:education-workflows}, and~\ref{fig:healthcare-workflows},.

\section{Screenshots of Task and App Environments}
\label{appendix:screenshot_task_app}

\begin{figure*}[h]
    \centering
    \includegraphics[width=\textwidth]{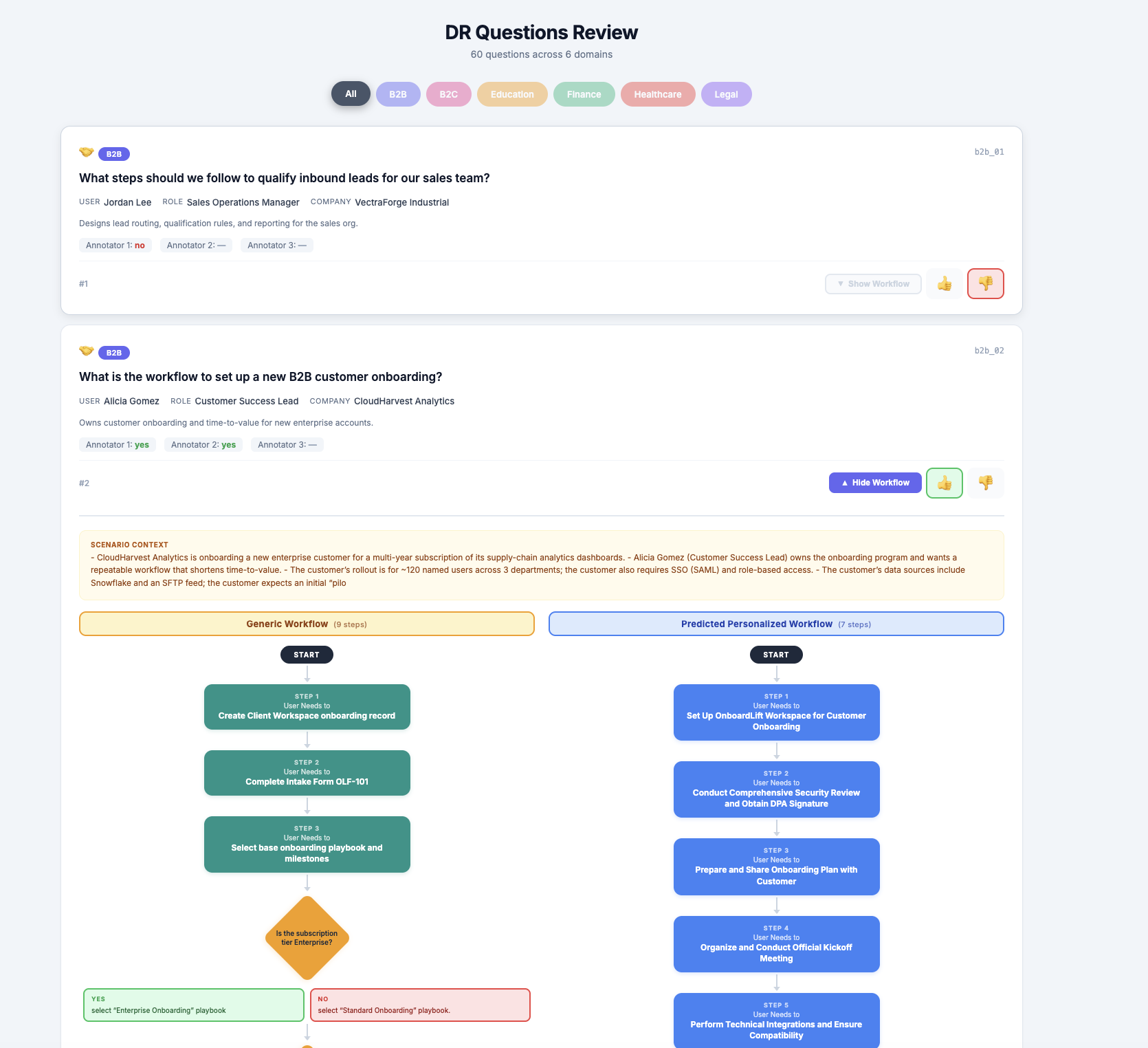}
    \caption{
    Screenshot of \drflow~task annotations procedure.
    }
    \label{fig:ss_task_annotation}
\end{figure*}

\paragraph{Task Annotation.} We provide the screenshot of our task annotation procedure in Figure~\ref{fig:ss_task_annotation}.

\begin{figure*}[h]
    \centering
    \includegraphics[width=\textwidth]{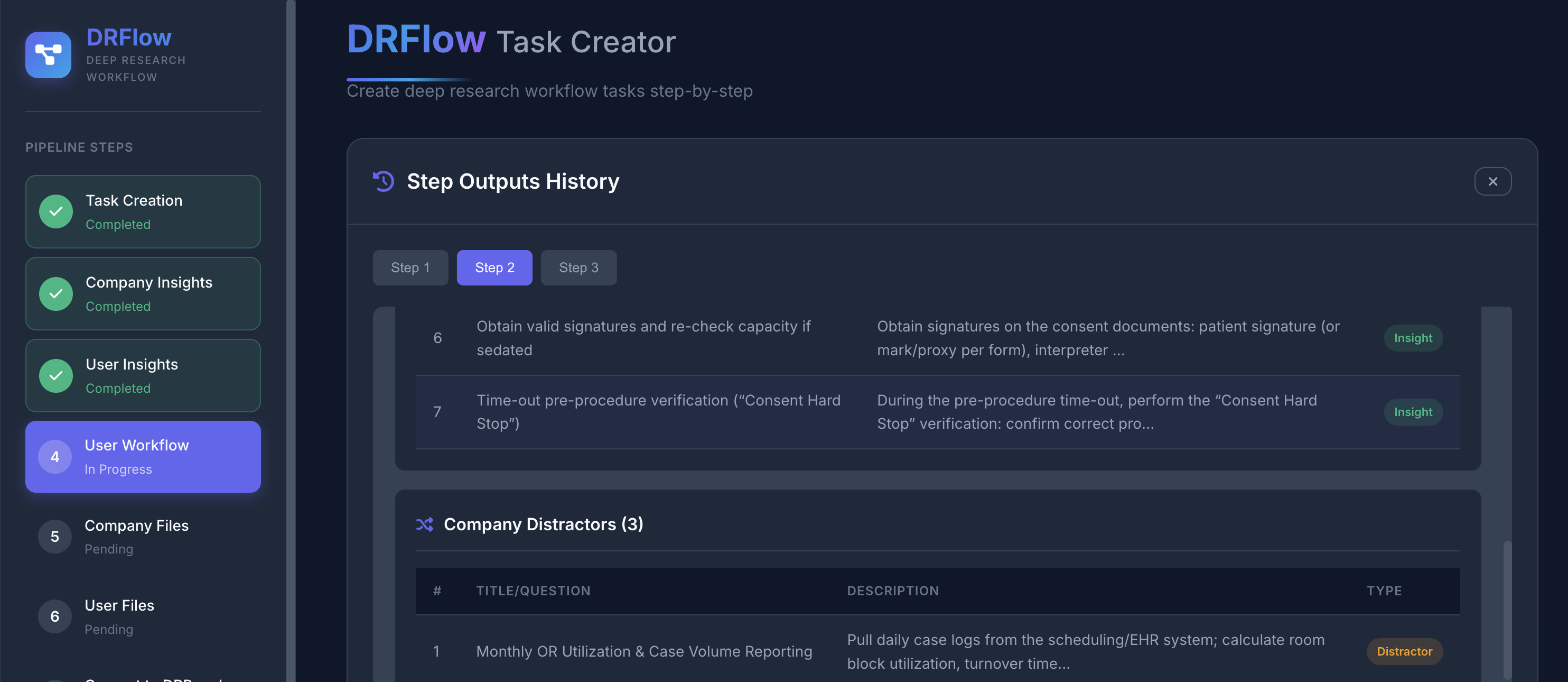}
    \caption{
    Screenshot of \drflow~task creation in a codeless web setup.
    }
    \label{fig:ss_task_creation}
\end{figure*}

\paragraph{Task Creation.} We also provide the support for creating \drflow~tasks in a no-code web setup. We show the screenshot of such setup in Figure~\ref{fig:ss_task_creation}.

\begin{figure*}[h]
    \centering
    \includegraphics[width=\textwidth]{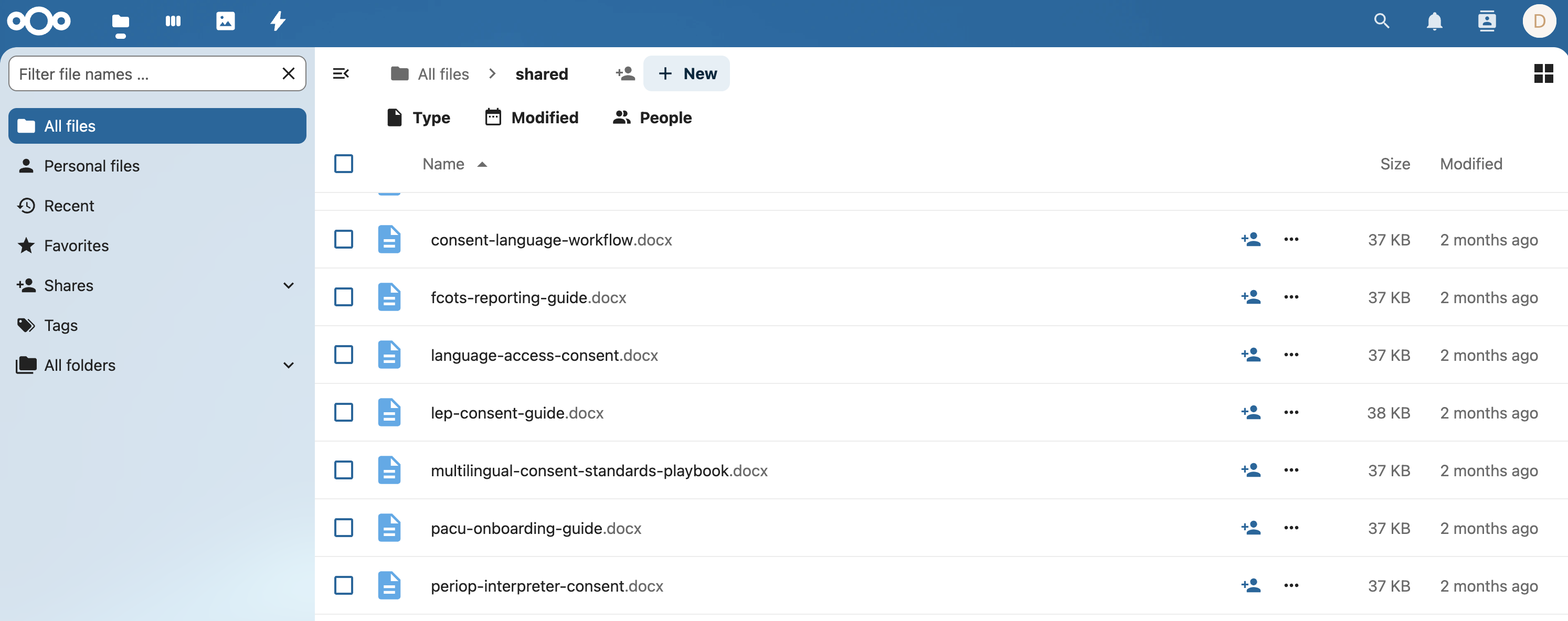}
    \caption{
    Screenshot of Nextcloud file management system with shared files organized in a list view.
    }
    \label{fig:ss_nextcloud}
\end{figure*}

\begin{figure*}[h]
    \centering
    \includegraphics[width=\textwidth]{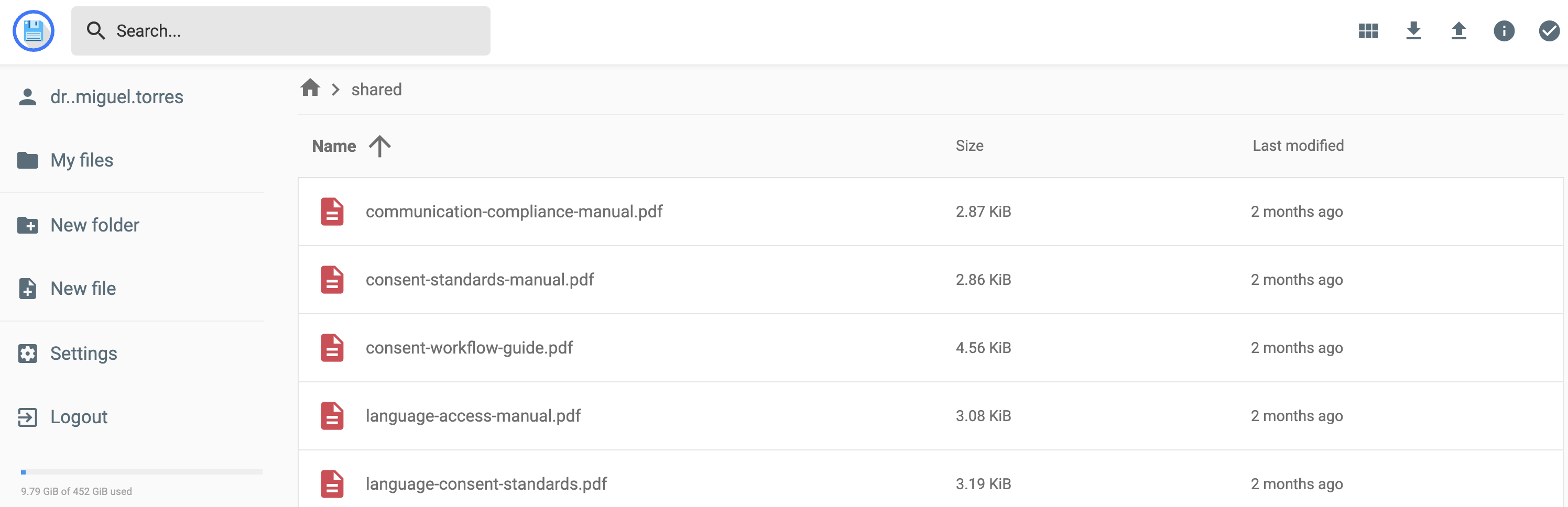}
    \caption{
    Screenshot of File Browser interface where files organized in a list view.
    }
    \label{fig:ss_file_browser}
\end{figure*}

\begin{figure*}[h]
    \centering
    \includegraphics[width=\textwidth]{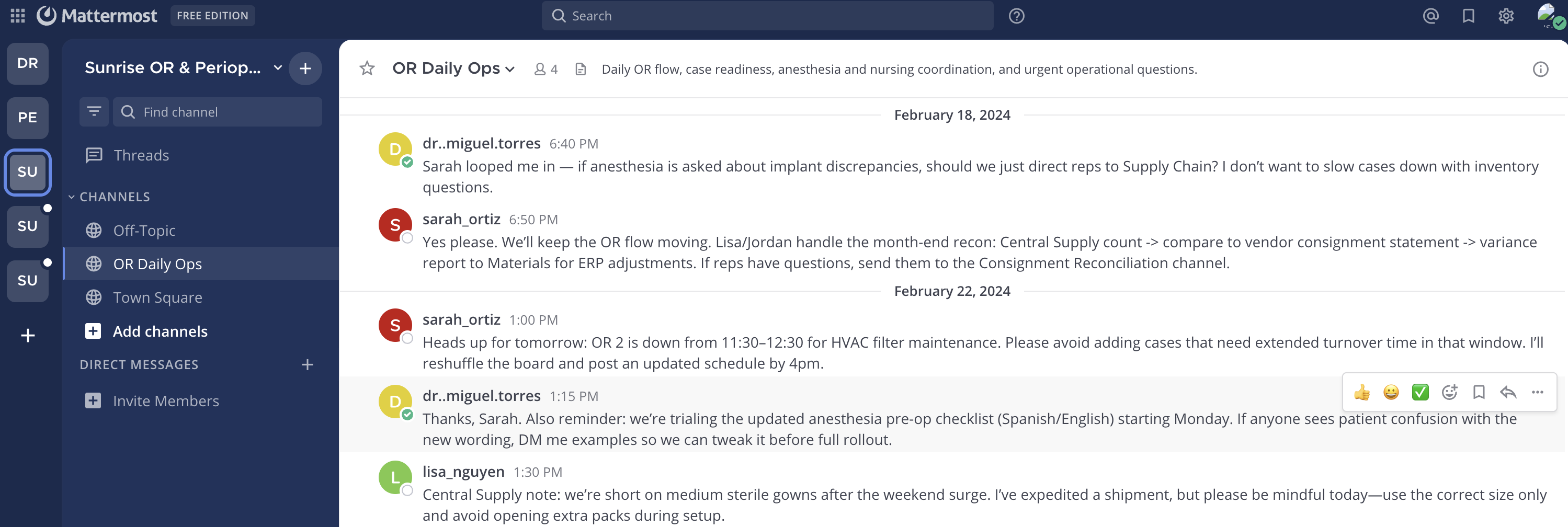}
    \caption{
    Screenshot of Mattermost chat interface showing discussion channels.
    }
    \label{fig:ss_mattermost}
\end{figure*}

\begin{figure*}[h]
    \centering
    \includegraphics[width=\textwidth]{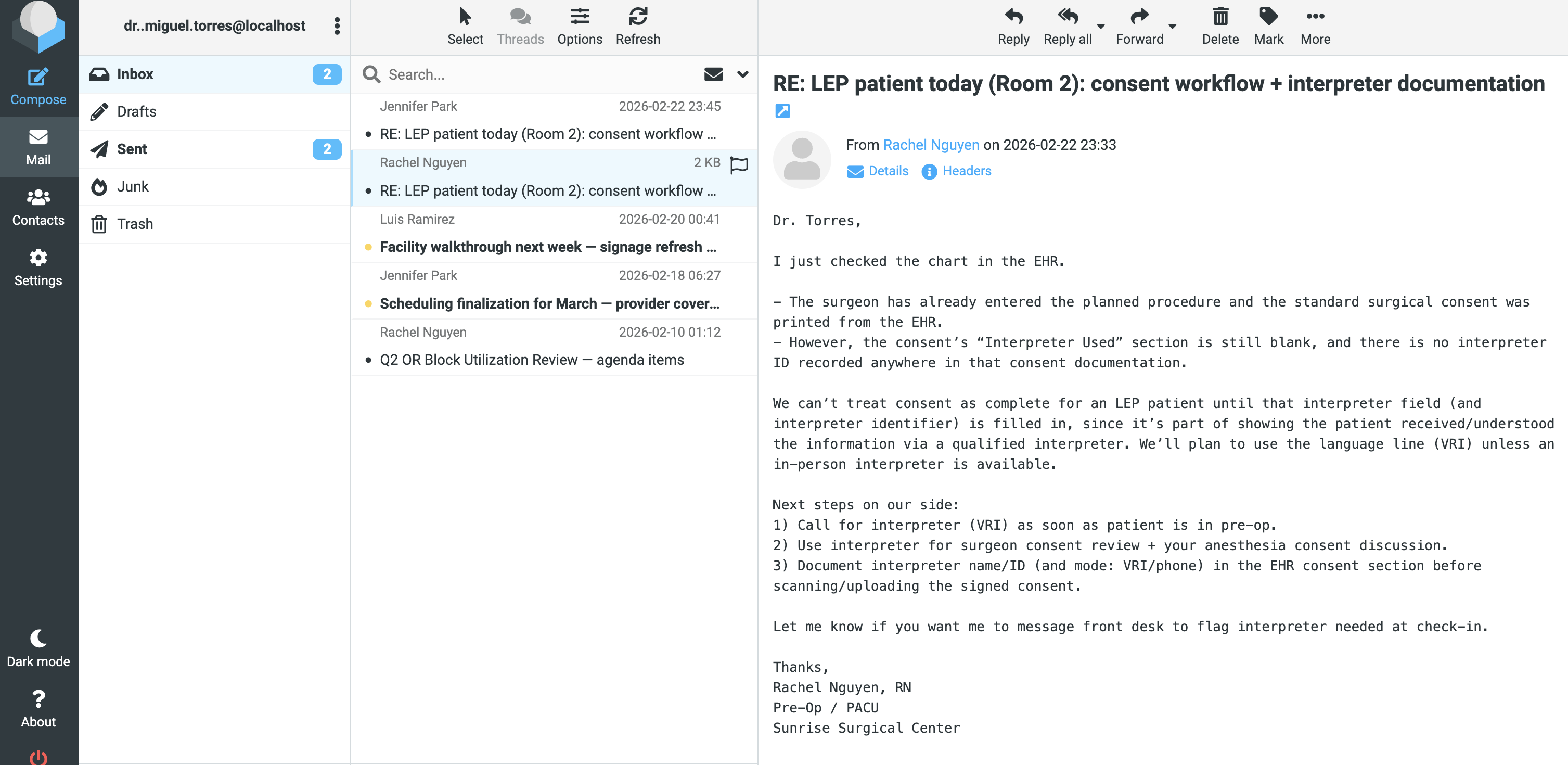}
    \caption{
    Screenshot of Roundcube email management system showing sample of user's email inbox.
    }
    \label{fig:ss_roundcube}
\end{figure*}

\paragraph{Application Environments.} We present the screenshot of different application environments that we use in \drflow~in Figure~\ref{fig:ss_nextcloud},~\ref{fig:ss_file_browser},~\ref{fig:ss_mattermost}, and~\ref{fig:ss_roundcube}.

\end{document}